\PassOptionsToPackage{caption=false,font=footnotesize}{subfig}
\documentclass[10pt,journal]{IEEEtran}
\usepackage{amsmath,amsfonts}
\usepackage{array}
\usepackage[caption=false,font=normalsize,labelfont=sf,textfont=sf]{subfig}
\usepackage{textcomp}
\usepackage{stfloats}
\usepackage{url}
\usepackage{tabularx,booktabs}
\newcolumntype{Y}{>{\centering\arraybackslash}X}
\usepackage{verbatim}
\usepackage{graphicx}
\usepackage{cite}  
\usepackage[utf8]{inputenc} 
\usepackage[T1]{fontenc}    
\usepackage{hyperref}       
\usepackage{nicefrac}       
\usepackage{microtype}      
\usepackage{xcolor}         
\usepackage{amssymb}
\usepackage[inline]{enumitem}
\usepackage{adjustbox}
\usepackage[linesnumbered,ruled,vlined]{algorithm2e}

\usepackage{multirow}
\usepackage{makecell}

\usepackage{newtxtext,newtxmath}  

\usepackage{todonotes}

\definecolor{questioncolor}{RGB}{200,0,0}

\definecolor{highlightcolor}{RGB}{0,128,128}

\usepackage{textcomp}
\usepackage[normalem]{ulem}
\definecolor{delcolor}{RGB}{119,119,17}       

\definecolor{bluecolor}{RGB}{0,0,255}

\definecolor{rob}{RGB}{150,150,150}
\newcommand{\rob}[1]{\textcolor{rob}{#1}}

\definecolor{reg}{RGB}{32,119,180}
\newcommand{\reg}[1]{\textcolor{reg}{#1}}

\definecolor{gen}{RGB}{116,196,118}
\newcommand{\gen}[1]{\textcolor{gen}{#1}}

\definecolor{com}{RGB}{255,87,1}

\newcommand{\CSI}{\mathbf{H}}
\newcommand{\csi}{\mathbf{h}}
\newcommand{\noise}{\mathbf{E}}

\newcommand{\ecsi}{h_{m,k}^t}
\newcommand{\ecsin}{\tilde{h}_{m,k}^t}

\newcommand{\ethet}{\theta_{m,k}^t}
\newcommand{\ethetn}{\tilde{\theta}_{m,k}^t}
\newcommand{\ethetnoise}{\Delta_{m,k}^t}

\newcommand{\ebell}{\text{bell}_{m,k}^t}
\newcommand{\startburst}{s_{m,k}}
\newcommand{\eepsilon}{\epsilon_{m,k}^t}
\newcommand{\aburst}{A_{\text{burst}}}
\newcommand{\pburst}{P_{\text{burst}}}
\newcommand{\lburst}{L_{\text{burst}}}

\newcommand{\enoise}{e_{m,k}^t}

\newcommand{\predCSI}{\mathbf{\hat{H}}}
\newcommand{\predcsi}{\mathbf{\hat{h}}}
\newcommand{\futureCSI}{\CSI}
\newcommand{\pastCSI}{\CSI}
\newcommand{\noisyCSI}{\mathbf{\tilde{H}}}

\newcommand{\speed}{v_{\mathrm{ue}}}
\newcommand{\delay}{\sigma_\tau}
\newcommand{\cm}{\mathrm{CM}}
\newcommand{\noisetype}{\mathrm{NT}}
\newcommand{\noisedegree}{\mathrm{ND}}

\newcommand{\Scenarios}{\mathbf{S}}

\newcommand{\CC}{\mathbb{C}}
\newcommand{\RR}{\mathbb{R}}
\newcommand{\EE}{\mathbb{E}}

\newcommand{\Model}{\textit{CSI-4CAST}}
\newcommand{\regular}{\textit{regular}}
\newcommand{\Regular}{\textit{Regular}}

\newcommand{\Robustness}{\textit{Robustness}}
\newcommand{\generalization}{\textit{generalization}}
\newcommand{\Generalization}{\textit{Generalization}}
\newcommand{\dataset}{\textit{CSI-RRG}}

\newcommand{\hfds}{\href{https://huggingface.co/CSI-4CAST}{dataset (Hugging Face)}}
\newcommand{\ghrepo}{\href{https://github.com/AI4OPT/CSI-4CAST}{evaluation protocols (GitHub)}}

\definecolor{darkgreen}{RGB}{0,100,0}
\definecolor{darkred}{RGB}{139,0,0}

\newcommand{\Yes}{\textcolor{darkgreen}{\textbf{\checkmark}}}
\newcommand{\No}{\textcolor{darkred}{\textbf{\ensuremath{\times}}}}

\definecolor{responsecolor}{RGB}{150,105,85}  
\newcommand{\response}[1]{\textcolor{black}{#1}}

\begin{document}

\title{\Model{}: A Hybrid Deep Learning Model for CSI Prediction with Comprehensive Robustness and Generalization Testing}
\author{Sikai Cheng*, Reza Zandehshahvar*, Haoruo Zhao*,
\thanks{*S. Cheng, R. Zandehshahvar, H. Zhao, and P. Van Hentenryck are with the H. Milton Stewart School of Industrial and Systems Engineering, NSF AI Institute for Advances in Optimization (AI4OPT), Georgia Institute of Technology, Atlanta, GA, USA. Corresponding author: Sikai Cheng (email: sikaicheng@gatech.edu).}
\and
$\text{Daniel A. Garcia-Ulloa}^\dagger$, $\text{Alejandro Villena-Rodriguez}^\dagger$, \\ $\text{Carles Navarro Manch\'on}^\dagger$, Pascal Van Hentenryck*
\thanks{$\dagger$Daniel A. Garcia-Ulloa, Alejandro Villena-Rodriguez, and Carles Navarro Manch\'on are with Keysight Technologies in Atlanta, GA; M\'alaga, Spain; and Barcelona, Spain, respectively.}
}



\maketitle

\begin{abstract}
Channel state information (CSI) prediction is a promising strategy for ensuring reliable and efficient operation of massive multiple-input multiple-output (mMIMO) systems by providing timely downlink (DL) CSI. While deep learning–based methods have advanced beyond conventional model-driven and statistical approaches, they remain limited in robustness to practical non-Gaussian noise, generalization across diverse channel conditions, and computational efficiency. This paper introduces \Model{}, a hybrid deep learning architecture that integrates \underline{4} key components, i.e., \underline{C}onvolutional neural network residuals, \underline{A}daptive correction layers, \underline{S}huffleNet blocks, and \underline{T}ransformers, to efficiently capture both local and long-range dependencies in CSI prediction. To enable rigorous evaluation, this work further presents a comprehensive benchmark, \dataset{} for \underline{R}egular, \underline{R}obustness and \underline{G}eneralization testing, which includes more than 300,000 samples across 3,060 realistic scenarios for both TDD and FDD systems. The dataset spans multiple channel models, a wide range of delay spreads and user velocities, and diverse noise types and intensity degrees. Experimental results show that \Model{} achieves superior prediction accuracy with substantially lower computational cost, outperforming baselines in 81.5\% of TDD scenarios and 44.4\% of FDD scenarios—the best performance among all evaluated models—while reducing FLOPs by $5\times$ and $3\times$ compared to LLM4CP, the strongest baseline. In addition, evaluation over \dataset{} provides valuable insights into how different channel factors affect the performance and generalization capability of deep learning models. Both the \hfds{} \footnote{Dataset: \url{https://huggingface.co/CSI-4CAST}} and \ghrepo{} \footnote{Code: \url{https://github.com/AI4OPT/CSI-4CAST}} are publicly released to establish a standardized benchmark and to encourage further research on robust and efficient CSI prediction.
\end{abstract}


\section{Introduction}
\label{sec:introduction}

\IEEEPARstart{M}{assive} multiple-input multiple-output (mMIMO) system has been widely adopted in fifth-generation (5G) wireless communication networks \cite{chataut2020massive}. By deploying large antenna arrays at base stations (BS) and employing advanced antenna configuration techniques such as directional beamforming, efficient precoding, and adaptive power allocation \cite{marzetta2016fundamentals,bjornson2017massive}, mMIMO significantly enhances spectral and energy efficiency, improves coverage, and reduces multiuser interference \cite{ngo2013energy,bjornson2018massive,bjornson2015optimal}. These advantages enable high user density and diverse services such as calling, video streaming, and internet browsing simultaneously, even within the same frequency band. However, the benefits of mMIMO rely on accurate real-time downlink (DL) channel state information (CSI) acquisition at BS \cite{huang2009limited,jose2011pilot}. In practice, CSI acquisition remains challenging due to the well-known aging effect \cite{nguyen-2005,truong-2013}, which arises from inevitable delays in wireless systems, including transmission, estimation \cite{gao-2020-channel-estimation}, and feedback delays, particularly in frequency division duplexing (FDD) mode \cite{li-2016}. Rapid channel variations caused by user mobility, multipath propagation, and channel noise further exacerbate this problem. Consequently, the acquired CSI often diverges from the true channel conditions, degrading BS operations such as precoding and power allocation.

To address this issue, CSI prediction has emerged as a promising strategy to mitigate the aging effect and provide timely DL CSI. In CSI prediction, the BS attempts to predict the DL CSI at a future time instant based on (noisy) observations of the uplink (UL) CSI at a previous instant. By forecasting future CSI from past observations, CSI prediction alleviates the impact of aging and reduces CSI acquisition overhead. In time division duplexing (TDD) systems, where UL and DL transmissions occur sequentially over the same frequency band, prediction constitutes an intra-band task. In contrast, in FDD systems, UL and DL transmissions occupy separate frequency bands, making prediction an inter-band task.

\response{Conventional CSI prediction has traditionally relied on explicit parametric and statistical channel models, including autoregressive (AR) models \cite{eyceoz-1998, truong-2013}, polynomial approximation predictors \cite{shen-2003-short-range-wireless-channel-prediction}, Taylor expansion-based models \cite{peng-2017-channel-prediction}, Kalman filtering frameworks \cite{kashyap-2017-performance,kim-2021}, and Prony-based predictors that exploit the angular-delay structure of mMIMO channels \cite{yin-2020}. These approaches are appealing due to their structured formulation, interpretability, and, in some cases, low computational complexity. For instance, AR and polynomial predictors are lightweight at inference time, while more structured approaches explicitly leverage physical channel properties. In particular, Prony-based angular-delay (PAD) prediction exploits angle-delay-Doppler structure and can achieve asymptotically vanishing prediction error as the number of base station antennas and bandwidth increase \cite{yin2020dealing,li2022multi}, and Wiener-type prediction admits theoretical guarantees on MIMO spectral efficiency \cite{loschenbrand2023spectral}.  However, the performance of model-based predictors is inherently tied to the accuracy of their underlying assumptions. In practice, these assumptions may not hold due to channel non-stationarity, model mismatch, and non-Gaussian noise. As a result, such methods often exhibit degraded performance in complex environments, scale poorly with increasing system dimensionality, and tend to deteriorate rapidly over longer prediction horizons \cite{kim2025machinelearningfuturewireless, kim-2021, stenhammar2023comparisonneuralnetworkswireless, gardner-1988}.}

\response{Recent studies position CSI prediction methods along several complementary axes to better understand their strengths and limitations. From a complexity-performance perspective, learning-based predictors can achieve strong predictive accuracy but typically require substantial offline effort for data collection and training \cite{diaz2025csi,kim2020massive}. Once trained, however, they enable efficient inference—particularly when computational resources such as GPUs are available at the BS—without the need for repeated parameter estimation at runtime \cite{diaz2025csi,kim2020massive}. In contrast, in resource-constrained scenarios, lightweight model-based approaches remain attractive due to their small footprint and predictable computational cost.  Another key dimension is the trade-off between interpretability and generalization. Model-based methods explicitly encode physical channel structure and often admit theoretical guarantees, whereas learning-based approaches are better suited to capturing nonlinear temporal and spatial dependencies in high-dimensional and non-stationary environments. This complementarity has motivated the development of hybrid CSI predictors that integrate model-driven structure with data-driven components to improve robustness while retaining interpretability \cite{adeogun2025towards,sun2025hybrid,ginige2025efficient}. For example, \cite{sun2025hybrid} introduces a data-driven state-space framework that preserves the interpretability of recursive filtering while enabling label-free training under standardized mmWave channel models, demonstrating strong performance under challenging channel conditions.}

\response{More recently, deep learning models have emerged as powerful tools for high-dimensional time-series forecasting and have been increasingly applied to CSI prediction in recent years \cite{jiang2025aicsiprediction5gadvanced, romoni2025survey}. Their primary advantage lies in their ability to capture complex temporal and spatial dependencies beyond the reach of conventional approaches, while still enabling efficient inference after training. A wide range of architectures has been explored in this context. Recurrent models such as LSTM-GRU combinations have been proposed to improve training stability and mitigate vanishing gradients \cite{helmy-2023-lstm-gru}. CNN-based approaches model the CSI matrix as a complex-valued image to preserve spatial and phase information \cite{zhang-2021}. Generative models, including GAN-based frameworks, have been used to reconstruct future CSI from historical observations \cite{safari-2020}. Transformer-based architectures leverage attention mechanisms to enable parallel multi-step prediction and reduce error accumulation \cite{jiang-2022}. Graph-based models, such as STEMGNN, capture joint temporal and frequency-domain correlations \cite{mourya-2024}. More recently, LLM-inspired approaches have framed CSI prediction as a sequence modeling problem analogous to next-token prediction, using GPT-2 \cite{liu-2024} or BERT-based architectures \cite{zhao2024mining} to improve generalization and handle missing data.} Despite these advances, deep learning-based CSI prediction methods still face three core challenges:
\begin{itemize}
    \item \textbf{Robustness to noise:} In practice, real-world wireless channels are affected by various noise sources that deviate from the standard additive white Gaussian noise (AWGN) assumption. These include phase noise stemming from imperfect local oscillators \cite{pitarokoilis-2015, zhilin2025effectrealisticoscillatorphase}, burst noise marked by short, high-amplitude spikes due to electromagnetic interference \cite{li-2004-performance-evaluation, zhou-2021-uplink-channel-estimation}, and packet drop noise resulting from system-level issues such as scheduling delays or network congestion \cite{morato2024packet}. Despite their practical significance, the robustness of CSI prediction methods under such noise conditions had not been properly explored.
    \item \textbf{Limited generalization:} Deep learning models often struggle to generalize to unseen scenarios or maintain performance under distribution shifts. Most models perform well only when training and evaluation data distributions are closely aligned. However, some previous works consider narrow in-distribution testing and adopt impractical assumptions, such as training separate models for different user velocities (e.g., \cite{mourya-2024}). In contrast, real-world wireless systems feature continuously varying channel conditions and user mobility, demanding models that can generalize reliably across diverse environments.
    \item \textbf{High computational cost:} Leading approaches like LLM4CP \cite{liu-2024} and CSI-BERT \cite{zhao2024mining} impose high computational costs due to their reliance on large pre-trained language models. These models demand substantial hardware resources, often exceeding what is feasible for deployment at each BS, particularly due to their high memory and throughput requirements. Moreover, recent findings suggest that such complex architectures may not be necessary for time-series forecasting, as simpler designs can deliver comparable or even superior performance \cite{tan2024are}.
\end{itemize}

\begin{table*}[t]
  \centering
  \caption{\response{\textbf{Architectural design choices in the previous learning-based CSI prediction models.}}}
  \label{tab:arch_design_comparison}
  \scriptsize
  \setlength{\tabcolsep}{4pt}
  \renewcommand{\arraystretch}{1.12}
  \resizebox{\linewidth}{!}{
  \begin{tabular}{p{3cm} p{5cm} c c c}
  \toprule
  \textbf{Model} & \textbf{Main design focus} & \textbf{Freq.--delay branch}$^{a}$ & \textbf{Duplex-aware arch.}$^{b}$ & \textbf{Efficiency-oriented lightweight design}$^{c}$ \\
  \midrule
  
  LSTM-GRU \cite{helmy-2023-lstm-gru}
  & Hybrid LSTM--GRU architecture for stronger temporal modeling and improved training stability
  & \No & \No & \No \\
  \midrule
  
  Transformer predictor \cite{jiang-2022}
  & Transformer-first design for parallel multi-step prediction and long-range dependency modeling
  & \No & \No & \No \\
  \midrule

  Deep UL2DL \cite{safari-2020}
  & CNN+GAN framework for UL$\rightarrow$DL mapping in FDD, formulated the prediction as a generative adversarial reconstruction task
  & \No & \No \, (FDD-only task) & \No \\
  \midrule
  
  STEMGNN \cite{mourya-2024}
  & Graph/spectral-temporal fusion
  & \No & \No & \No \\
  \midrule
  
  LLM4CP \cite{liu-2024}
  & CSI-specific preprocessing, patching, and CSI attention to adapt CSI for a pretrained GPT-2 backbone with task-specific fine-tuning
  & \Yes & \No & \No \, (heavy backbone) \\
  \midrule
  
  CSI-BERT \cite{zhao2024mining}
  & BERT-inspired CSI predictor based on masked-token learning, with emphasis on handling missing data
  & \No & \No & \No \, (heavy backbone)\\
  \midrule
  
  \Model{}
  & CSI-specific integration
  & \Yes & \Yes & \Yes \\
  
  \bottomrule
  \end{tabular}
  }
  
  \vspace{2pt}
  \begin{minipage}{0.98\linewidth}
  \footnotesize
  $^{a}$ Explicit construction of both frequency- and delay-domain CSI representations via a transform (e.g., IDFT).\\
  $^{b}$ Different architectural processing for TDD and FDD within the same model.\\
  $^{c}$ A lightweight local extractor/backbone is an explicit architectural objective, not merely a side effect.
  \end{minipage}
  \end{table*}

This paper addresses the challenges mentioned above in CSI prediction through the following core contributions:
\begin{itemize}
  \item \textbf{\Model{}}: \response{This paper proposes a novel deep learning architecture for CSI prediction that improves the accuracy-efficiency trade-off through a CSI-specific integration of complementary modules. \Model{} integrates \underline{4} key neural components—\underline{C}NN-based residual blocks, duplex-aware \underline{A}CL modules, \underline{S}huffleNet-based lightweight feature extractors, and \underline{T}ransformer encoders—together with an explicit IDFT-based frequency--delay branch. The overall design is guided by the structural properties of CSI and the underlying duplexing conditions. Specifically, the architecture jointly captures (i) frequency--delay coupling, (ii) duplex-aware axis-wise correction, and (iii) efficient local-global feature modeling within a unified framework. Importantly, the novelty of \Model{} does not stem from introducing a new individual module, but from a principled integration of existing components tailored to the characteristics of CSI prediction and practical computational constraints. Table~\ref{tab:arch_design_comparison} provides a structured comparison of this design against representative prior models.} Experimental results demonstrate that \Model{} achieves the lowest NMSE in \response{81.5\%} of test scenarios under TDD, while reducing computational cost by approximately 5$\times$ in FLOPs compared to LLM4CP, the second-best model. In the more challenging FDD setting, \Model{} leads in \response{44.4\%} of test scenarios—ranking highest among all models—and achieves over a $3\times$ reduction in FLOPs relative to LLM4CP.

  \item \textbf{Comprehensive evaluation suite:} This paper presents a large-scale and realistic benchmark designed for the training and evaluation of CSI prediction models. The dataset includes over 300,000 samples across 3,060 distinct scenarios for both TDD and FDD, encompassing multiple standardized channel models, a range of delay spreads, varying user mobility speeds, and a wide spectrum of SNR conditions. Additionally, it incorporates several types of non-Gaussian noise—such as burst, phase, and packet-drop noise—at multiple intensity levels, reflecting perturbations commonly encountered in practice yet often overlooked in previous studies. Collectively, these elements constitute the most comprehensive framework to date for rigorously stress-testing CSI prediction methods under diverse and realistic conditions.

  \item \textbf{Reproducibility and impact:} The \dataset{}, short for \Regular{}, \Robustness{}, and \Generalization{}, together with its evaluation protocols, is publicly released to provide a standardized benchmark for CSI prediction. It covers TDD/FDD operation, noise stress testing, and cross-scenario generalization. The \dataset{} benchmark is designed to lower the barrier to rigorous comparison and foster progress toward more robust and efficient CSI prediction.
\end{itemize}

Throughout this paper, the following notational conventions are adopted. The bold capital notation $\CSI$ represents the CSI matrix or tensor, while the bold lowercase notation $\csi$ denotes the CSI vector (e.g., along the antenna or subcarrier dimension), and the non-bold lowercase notation $h$ indicates the element-wise CSI. For any matrix, vector, or element of CSI, ${(\cdot)}^t$ and ${(\cdot)}^\mathcal{T}$ denote the CSI at a specific time index $t$ and over a time sequence $\mathcal{T}$, respectively. The notation $\mathbf{\tilde{(\cdot)}}$ represents the noisy observation of the CSI, while $\mathbf{\hat{(\cdot)}}$ denotes its predicted value. For a general complex matrix, $(\cdot)^{\dagger}$ denotes the Hermitian transpose, and $\|\cdot\|_{F}$ represents the Frobenius norm.

\section{Problem Definition}
\label{sec:problem-definition}

This study investigates the task of DL CSI prediction in a MIMO system, employing Orthogonal Frequency Division Multiplexing (OFDM) for signal transmission. In this system, the BS is equipped with a dual-polarized uniform planar array (UPA) composed of $M_h$ rows and $M_v$ columns of antenna elements. The User Equipment (UE) has a single omnidirectional receive antenna. OFDM splits the transmission bandwidth into $N_{\mathrm{sc}}$ orthogonal subcarriers, allowing for efficient frequency-domain processing.

According to established models~\cite{kalachikov-2018, li-2019, ma-2018}, the CSI at time $t$, denoted as $\CSI^t$, is represented as a complex-valued tensor:
\begin{equation}
  \CSI^t \in \CC^{N_{\mathrm{tr}} \times N_{\mathrm{re}} \times N_{\mathrm{sc}}}.
  \label{eq: definition of CSI}
\end{equation}
It captures the channel coefficients between each transmit-receive antenna pair (spatial dimension) and across all subcarriers (frequency dimension) at a specific time instant $t$. Each matrix element is complex-valued: its magnitude reflects path gain (attenuation) and its phase reflects the propagation-induced phase shift, including hardware/oscillator offsets. The transmitter employs $N_{\mathrm{tr}} = 2 \times M_h \times M_v$ antennas, while the receiver is equipped with $N_{\mathrm{re}} = 1$.A single snapshot $\CSI^t$ implicitly encodes delay and angular structure across subcarriers and antennas, and a sequence of CSI further captures the channel's temporal evolution (e.g., Doppler and path dynamics).

\begin{figure}[!t]
  \centering
  \includegraphics[width=0.85\linewidth]{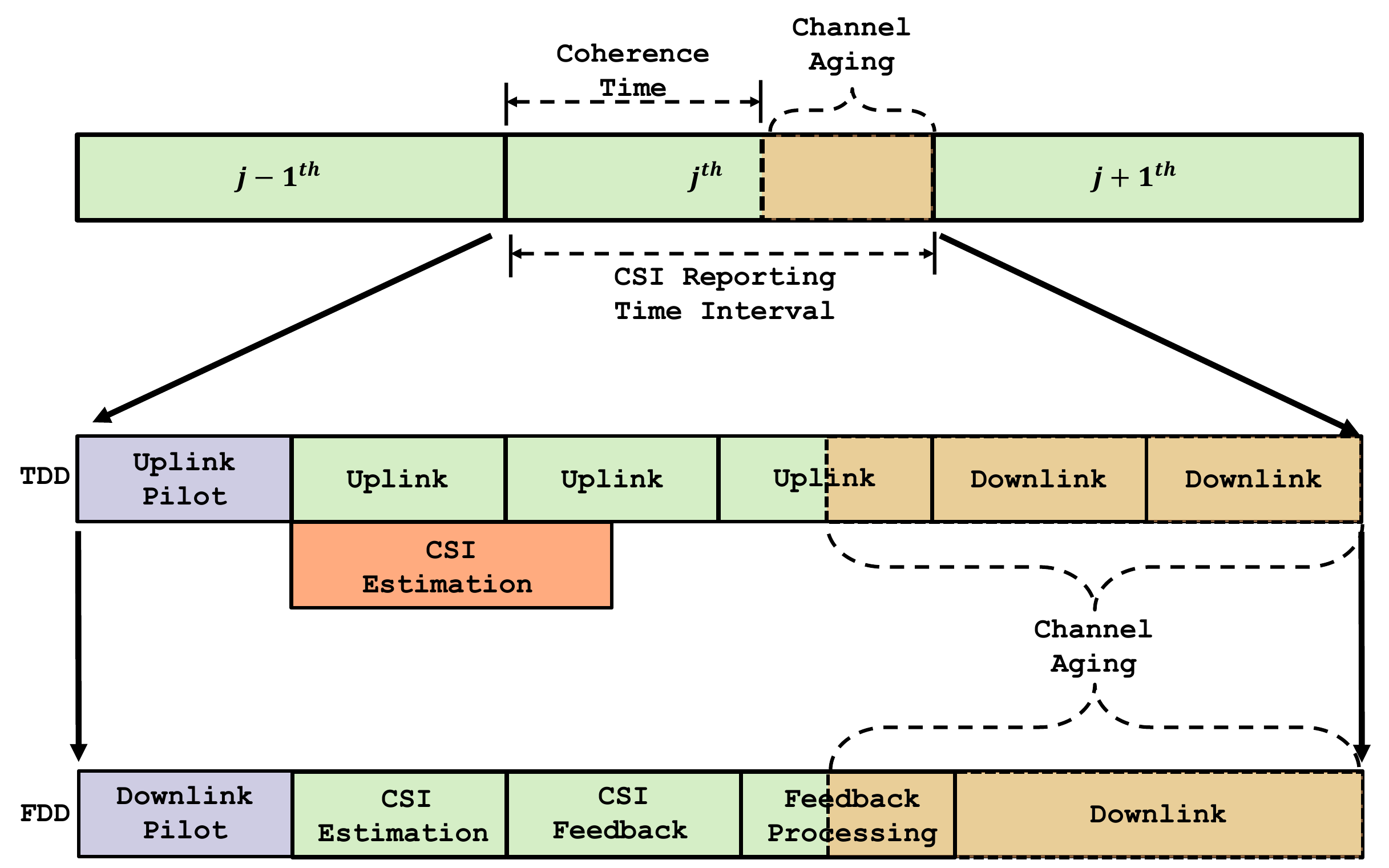}
  \caption{\textbf{The channel aging problem.}}
  \label{fig:channel_aging}
\end{figure}

Accurate CSI conveys essential information for BS functions such as precoding, scheduling, and power control. However, channel aging often prevents the BS from accessing the up-to-date DL CSI. Fig.~\ref{fig:channel_aging} illustrates the detailed transmission sequence in TDD and FDD. In the top part of the figure, the time axis is divided into successive CSI reporting intervals ${j-1}, j, j+1$. Ideally, for each interval, a pilot signal—known reference signals shared between the BS and UE—is transmitted at the beginning. CSI is then estimated by comparing the received pilot with the shared reference, and the resulting CSI is assumed stationary, guiding the BS's operations. However, in modern communication systems characterized by mobility and rapidly changing multipath environments, the coherence time—i.e., the time interval over which channel conditions can be considered constant—shortens. This, combined with inevitable estimation, processing, scheduling delays, and feedback delays (in FDD systems), leads to the channel aging problem: by the time DL transmission occurs, the reported CSI is already outdated and no longer reflects current DL conditions. The bottom part of the figure illustrates detailed time-interval breakdowns for TDD and FDD scenarios, highlighting the occurrence of channel aging. According to 3GPP procedures \cite{3gpp-38331-r17-v1720}, in TDD systems, where UL and DL share the same frequency band and are separated in time, the BS can—after reciprocity calibration—use UL pilot-based estimates for precoding in the next DL slot. However, a scheduled UL transmission block typically follows the UL pilot transmission, and the delay before the next scheduled DL transmission may render these estimates stale. In FDD systems, where UL and DL operate on different frequencies and reciprocity does not apply, DL CSI is estimated at the UE based on the DL pilots. The BS transmits DL pilots; the UE estimates the DL channel, compresses or quantizes it, and sends it back on the UL band. The BS decodes this feedback and applies it at the subsequent DL scheduling boundary. Despite the concurrent operation of UL and DL in FDD—eliminating the need to wait for UL transmissions to complete—latency from feedback and processing may still exceed the coherence time.

\begin{figure}[!t]
  \centering
  \subfloat[TDD: intra-band UL/DL CSI prediction]{
    \includegraphics[width=0.9\linewidth]{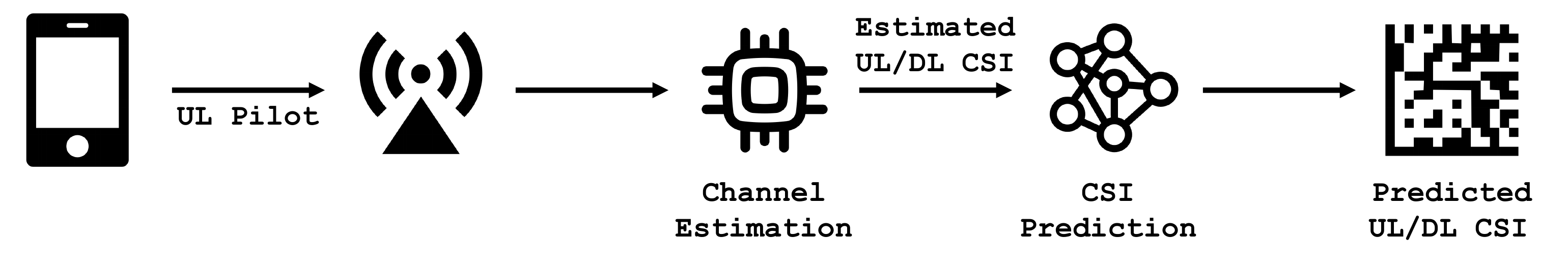}
    \label{fig:dl_csi_prediction_schema1}
  }\
  \subfloat[FDD: inter-band CSI prediction from UL to DL]{
    \includegraphics[width=0.9\linewidth]{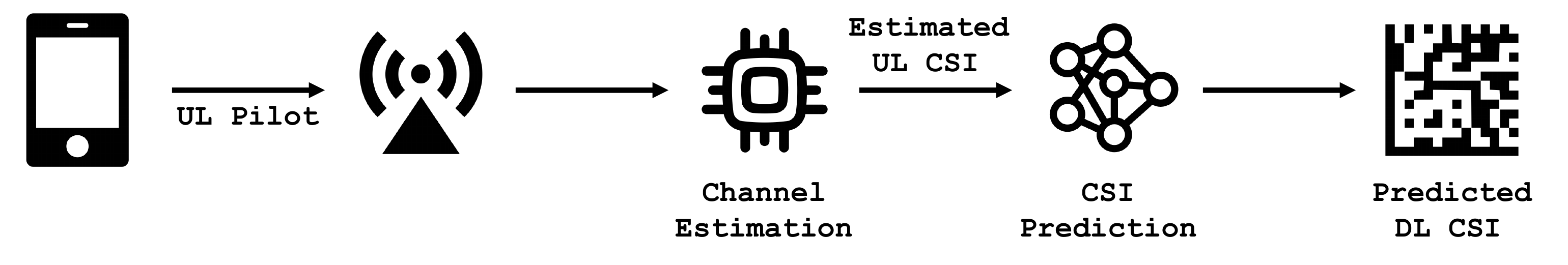}
    \label{fig:dl_csi_prediction_schema3}
  }
  \caption{\textbf{An illustration of the DL CSI acquisition schemas.}}
  \label{fig:dl_csi_prediction_schema}
\end{figure}

\begin{figure}[!t]
  \centering
  \subfloat[TDD]{
    \includegraphics[width=0.45\linewidth]{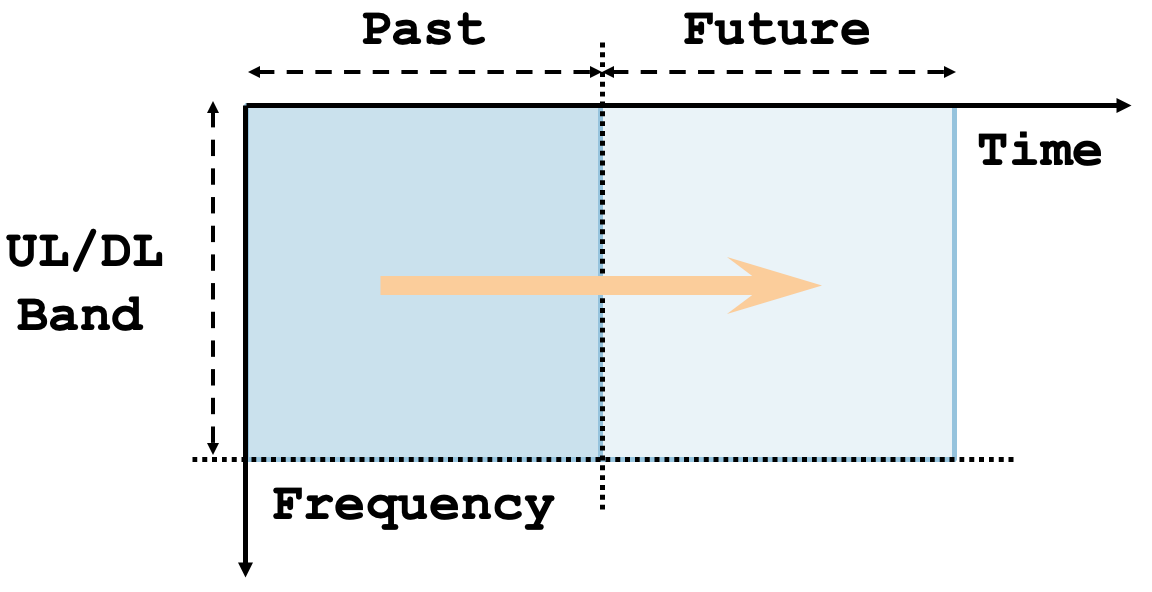}
    \label{fig:tdd_csi_prediction}
  }
  \subfloat[FDD]{
    \includegraphics[width=0.45\linewidth]{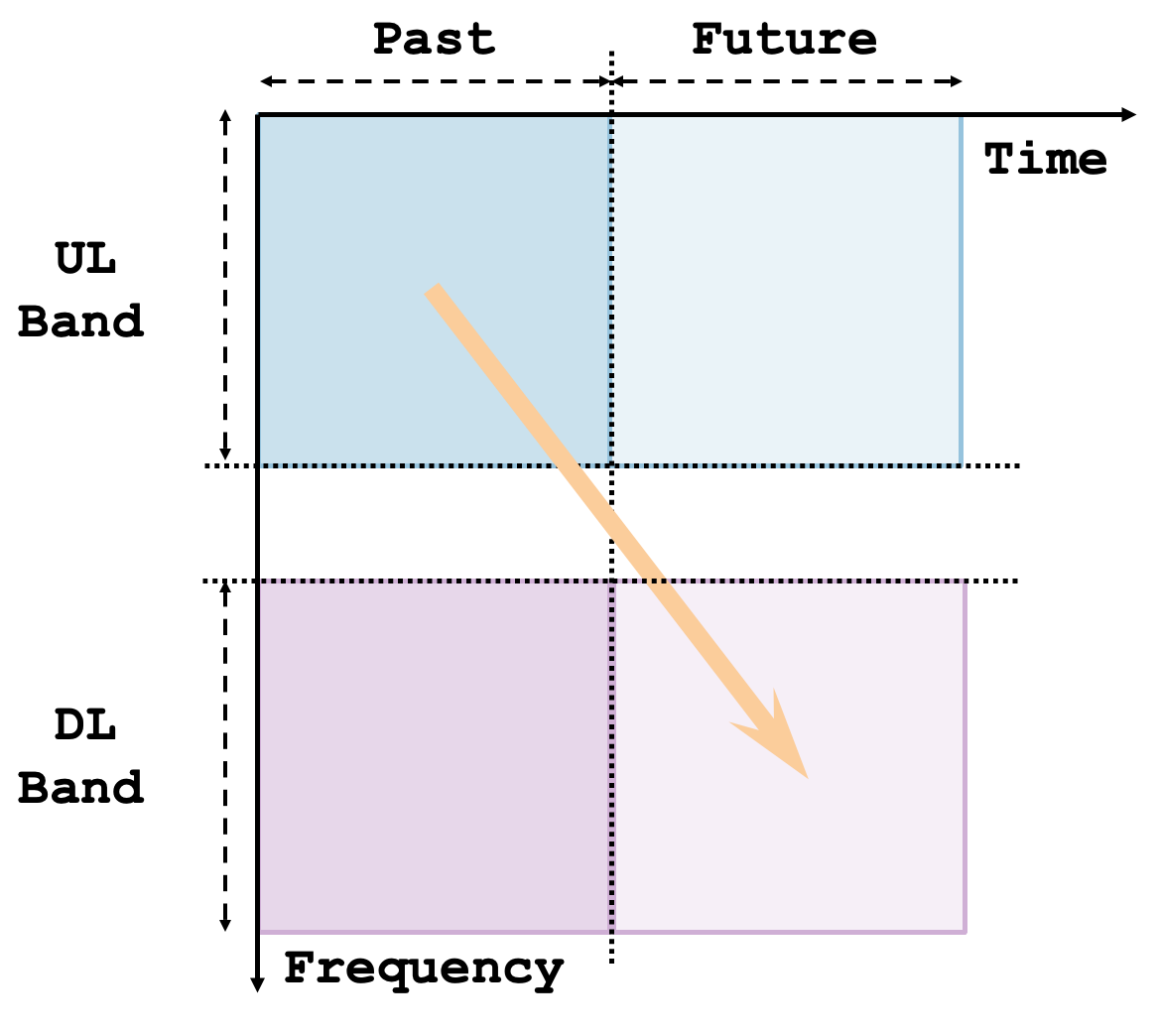}
    \label{fig:fdd_csi_prediction}
  }
  \caption{\textbf{An illustration of the CSI Prediction in the time-frequency domain.}}
  \label{fig:csi_prediction_illustration_tdd_fdd}
\end{figure}

To mitigate channel aging, CSI prediction forecasts near-future DL CSI based on recent pilot-based observations, thereby reducing the pilot-to-use delay and improving the BS's precoding accuracy. The prediction framework aligns with the CSI acquisition pipeline: as illustrated in Fig.~\ref{fig:dl_csi_prediction_schema1}, in TDD systems, where UL and DL share the same frequency band, intra-band prediction leverages historical UL/DL CSI to predict future UL/DL CSI (Fig.~\ref{fig:tdd_csi_prediction}) \cite{gardner-1988}. In contrast, in FDD systems, CSI prediction is utilized to further eliminate the feedback and associated processing overhead by shifting the prediction entirely to the BS side. In this case, the BS uses historical UL CSI (estimated from UL pilots) to directly predict future DL CSI, as shown in Fig.~\ref{fig:dl_csi_prediction_schema3}. Since this approach involves prediction across both time and frequency bands, it is referred to as inter-band prediction (Fig.~\ref{fig:fdd_csi_prediction}) \cite{liang-2018, miretti-2018}.

Consistent with the intra-band and inter-band schemas described above, both settings are unified under a single formulation by treating band-specific details as part of the input-target definitions. Denote the historical CSI sequence as follows: 
\begin{equation}
    \pastCSI^\mathcal{T} = \{ \pastCSI^{t - |\mathcal{T}| + 1}, \pastCSI^{t - |\mathcal{T}| + 2}, ..., \pastCSI^{t} \}
\end{equation}
where $\mathcal{T}$ represents the historical window $\mathcal{T} = \{t - |\mathcal{T}| + 1, \dots, t-1, t\}$.  CSI prediction can be considered as learning the following mapping:
\begin{equation}
  \begin{aligned}
    \predCSI^\mathcal{P} & = f_{\boldsymbol{\Omega}} \left( \pastCSI^\mathcal{T} \right), \\
  \end{aligned}
  \label{eq:csi_prediction_formulation}
\end{equation}
where $\mathcal{P} = \{t+1, \dots, t+|\mathcal{P}| \}$ is the prediction horizon, and $\predCSI^\mathcal{P}\in \CC^{|\mathcal{P}| \times N_{\mathrm{tr}} \times N_{\mathrm{re}} \times N_{\mathrm{sc}}} $ represent the predicted future DL CSI at time slot $t$. The parameters $|\mathcal{T}|$ and $| \mathcal{P}|$ denote the length of the input (past) CSI sequence and the prediction horizon, respectively. The prediction function $f_{\boldsymbol{\Omega}}(.)$ is parameterized by ${\boldsymbol{\Omega}}$.

Under realistic conditions, clean and accurate CSI is generally unattainable due to the presence of transmission noise and inevitable estimation errors. Consequently, the model utilizes noisy CSI as input instead of ideal CSI. The noisy past CSI at time $t$, denoted by $\noisyCSI^{t}$, is defined as:
\begin{equation}
  \noisyCSI^{t} = \pastCSI^{t} + \noise^{t},
  \label{eq:additive_noise}
\end{equation}
\response{where $\noise^{t}$ denotes an effective corruption term at time $t$. Note that this unified observation model is not restricted to independent AWGN and does not imply that $\noise^{t}$ is independent of $\pastCSI^{t}$. Instead, $\noise^{t}$ may be signal-dependent, temporally structured, or non-Gaussian, allowing the same formulation to cover physically non-additive perturbations through an equivalent error representation. Detailed element-wise formulations for various noise types that are considered in this paper, including AWGN, phase noise, burst noise, and packet-drop noise are provided in Appendix~\ref{sec:additive-noise}.}

Consequently, the predicted CSI under the noisy channel is obtained as:
\begin{equation}
  \begin{aligned}
    \predCSI^\mathcal{P} & = f_{\boldsymbol{\Omega}} \left( \noisyCSI^\mathcal{T} \right). \\
  \end{aligned}
  \label{eq:csi_prediction_formulation_noisy}
\end{equation}

This paper introduces a novel deep learning model to learn the mapping $f_{\boldsymbol{\Omega}}$ for CSI prediction. The proposed model is designed to address efficiency, robustness, and generalization and is evaluated against various deep learning models using a comprehensive dataset.

\section{Methodology}
\label{sec:methodology}

This section outlines the proposed method for CSI prediction and describes the proposed deep learning architecture, denoted as \Model{}. 

Let $\mathcal{D}_{\text{train}} = \{ (\noisyCSI^{\mathcal{T}_i}, \futureCSI^{\mathcal{P}_i}) \}_{i=1}^{N}$ be the training dataset, consisting $N=|\mathcal{D}_{\text{train}}|$ pairs of historical and future CSI sequences. The learning process is then formulated as the following optimization problem:
\begin{equation}
  \begin{aligned}
    \min_{\boldsymbol{\Omega}} \quad & \EE_{(\noisyCSI^{\mathcal{T}_i}, \futureCSI^{\mathcal{P}_i}) \sim \mathcal{D}_{\text{train}}} \left[ \mathcal{L} (\predCSI^{\mathcal{P}_i}, \futureCSI^{\mathcal{P}_i}) \right] \\
    \text{s.t.} \quad & \predCSI^{\mathcal{P}_i} = f_{\boldsymbol{\Omega}} \left( \noisyCSI^{\mathcal{T}_i} \right),
  \end{aligned}
  \label{eq:learning_problem}
\end{equation}
where the loss function $\mathcal{L}(.)$ is considered as Normalized Mean Squared Error (NMSE) between the actual and predicted CSI, and is defined as follows:
\begin{equation}
  \text{NMSE} (\predCSI^\mathcal{P}, \futureCSI^\mathcal{P}) = \frac{
    \sum_{t \in \mathcal{P}} \left\| \predCSI^{t} - \futureCSI^{t} \right\|_F^2
  }{
    \sum_{t \in \mathcal{P}} \left\| \futureCSI^{t} \right\|_F^2
  }
  \label{eq:NMSE}
\end{equation}
where$\|\cdot\|_F$ is the Frobenius norm.

\begin{figure*}[!ht]
  \centering
  \includegraphics[width=\textwidth]{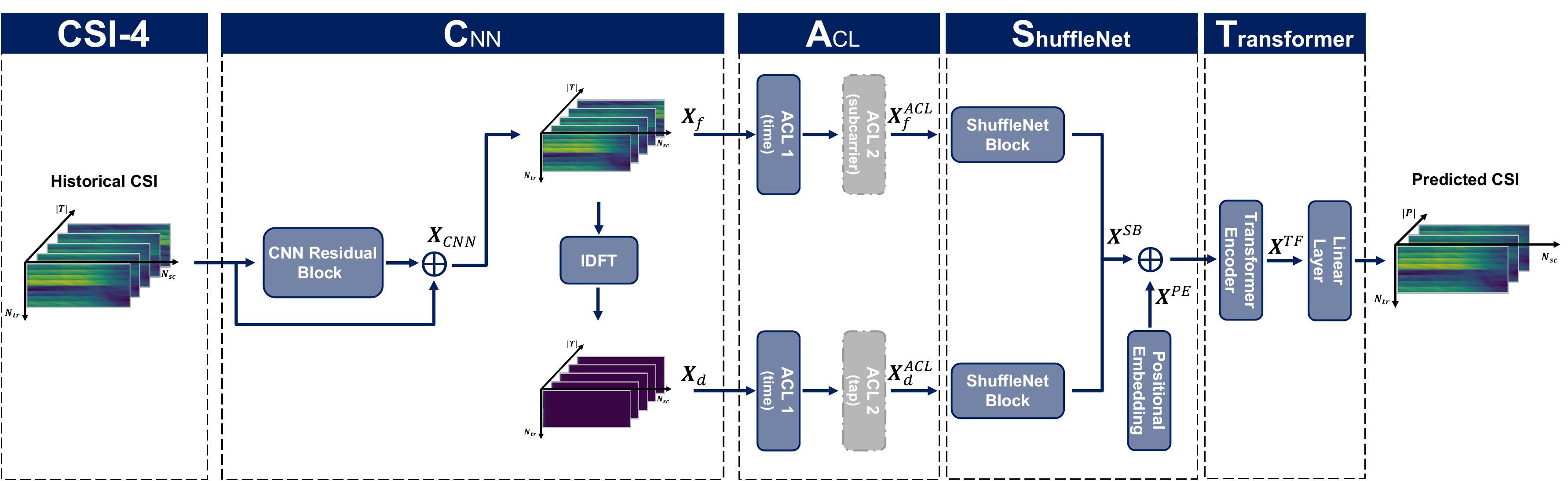}
  \caption{\textbf{The proposed \Model{}.} Historical CSI is first processed by a \underline{C}NN residual block, followed by an inverse DFT (IDFT) to obtain the delay-domain representation. Both frequency- and delay-domain features are then refined by \underline{A}CL layers and passed through a \underline{S}huffleNet block. Finally, the \underline{T}ransformer block maps the embedded features to predict future CSI. The ACL2 layer (in gray) applies only to the FDD.}
  \label{fig:proposed_model}
\end{figure*}

Fig.~\ref{fig:proposed_model} illustrates the schematic of \Model{}. To handle the complexity and high dimensionality of CSI sequences, \Model{} integrates several specialized components-Convolutional neural networks (CNN), Adaptive Correction Layers (ACLs), ShuffleNet Blocks, and Transformer encoders. This design enables efficient feature extraction and robustness in CSI prediction. 

\paragraph{Per-Antenna Modeling for Scalability and Efficiency}
Following standard practice in the literature~\cite{liu-2024}, each transmitter–receiver pair is modeled independently for CSI prediction. This allows \Model{} to operate separately on each pair, improving scalability and efficiency. The input of the model is defined as:
\begin{equation}
  \mathbf{X} = \noisyCSI^{\mathcal{T}}_m \in \CC^{|\mathcal{T}| \times N_{\mathrm{sc}}},
  \label{eq:input_to_model}
\end{equation}
representing the CSI sequence for a single transmitter–receiver pair $m$. To comply with the common requirement that neural networks operate on real-valued tensors, the real and imaginary parts of $\mathbf{X}$ are stacked and presented as:
\begin{equation}
  \mathbf{X}_{\mathbf{r}} = [ \operatorname{Re}(\mathbf{X}),\ \operatorname{Im}(\mathbf{X}) ] \in \RR^{2 \times|\mathcal{T}| \times N_{\mathrm{sc}}}.
  \label{eq:input_to_model_real}
\end{equation}

\paragraph{CNN-based Residual Representation}
Given the input CSI $\mathbf{X}_{\mathbf{r}}$, a CNN-based module, inspired by \cite{chen2024complex, zhang2017beyond}, is employed to extract structured features. By capturing local correlations across time and frequency, the CNN learns residual representations that refine noisy CSI observations, while its inherent smoothing property mitigates measurement inaccuracies, jointly enhancing robustness. The resulting representation is then denoted as $\mathbf{X}_{\mathrm{CNN}} \in \RR^{2 \times|\mathcal{T}| \times N_{\mathrm{sc}}}$.

The module consists of stacked 2D convolutional layers, combined with batch normalization and nonlinear activation functions. The channel configuration evolves across the network depth as $[2, 4, \cdots, 2^\nu, \cdots, 4, 2]$, where the initial value of $2$ reflects the real and imaginary components of the input, and $\nu$ represents the depth of the network. By virtue of carefully selected kernel sizes, padding schemes, and a symmetric channel structure, the output of the CNN module is guaranteed to preserve the dimensionality of the input.

\paragraph{Delay-Domain Representation for Multi-Path Pattern Utilization}
To complement the frequency-domain representation, the CSI is also transformed into the delay domain, where the signal is described in terms of propagation delays rather than subcarrier frequencies. In this view, each tap corresponds to the signal arriving through a distinct propagation path—for example, direct transmission or reflections from surrounding objects. \cite{Rappaport2024, tse-2005} The resulting path-level representation aggregates the per-subcarrier responses into a small number of significant taps that capture the dominant paths and overall delay spread. \response{This form is physically meaningful for multipath propagation and can be more concentrated when the channel is approximately sparse or compressible in the delay domain, which may help the network exploit dominant path structure more effectively \cite{bajwa2010compressed, lu-2020-multi-resolution-csi-feedback, ji2023enhancingdeeplearningperformance}.} Let $\mathbf{X}_{\mathbf{f}} \in \RR^{|\mathcal{T}| \times 2N_{\mathrm{sc}}}$, obtained from the $\mathbf{X}_{\mathrm{CNN}}$ by concatenating the real and imaginary dimensions, denote the real-valued representation of the CSI sequence in the frequency domain. The transformation from frequency to delay domain is accomplished via the inverse discrete Fourier transform (IDFT), expressed as
\begin{equation}
  \mathbf{X}_{\mathbf{d}}^{\CC} = \mathbf{X}_{\mathbf{f}}^{\CC} \mathbf{F}_{\mathbf{d}}^{\dagger} \in \CC^{|\mathcal{T}| \times N_{\mathrm{sc}}}.
  \label{eq:delay_domain_representation}
\end{equation}
Here, $\mathbf{X}_{\mathbf{f}}^{\CC} \in \CC^{|\mathcal{T}| \times N_{\mathrm{sc}}}$ denotes the complex-valued tensor converted from $\mathbf{X}_{\mathbf{f}}$ for the IDFT operation, and $\mathbf{F}_{\mathbf{d}} \in \CC^{N_{\mathrm{sc}} \times N_{\mathrm{sc}}}$ is the unitary DFT matrix. Subsequently, the tensors are converted back into real-valued form: 
$\mathbf{X}_{\mathbf{d}} \in \RR^{|\mathcal{T}| \times 2N_{\mathrm{sc}}}$ is obtained from $\mathbf{X}_{\mathbf{d}}^{\CC}$ by concatenating its real and imaginary components along the last dimension, representing the real-valued CSI sequence in the delay domain.

\paragraph{Adaptive Correction Layer (ACL) for Underlying Structures}
To capture intrinsic dependencies within $\mathbf{X}_{\mathbf{f}}$ and $\mathbf{X}_{\mathbf{d}}$, ACLs are introduced, motivated by~\cite{chen2024modelling5genergyconsumption,zhao2024mining}. \response{The ACL is designed as an adaptive calibration mechanism that complements the backbone feature extractor by explicitly refining intermediate CSI representations along physically meaningful axes. This is needed because different historical time lags and different subcarrier/tap positions are not equally informative for future CSI prediction. Along the temporal axis, CSI exhibits temporal coherence but also channel aging, so the predictive reliability of historical observations is inherently lag dependent. Along the subcarrier/tap axis, the learned CSI representations exhibit structured variation across frequency features or delay taps, reflecting spectral correlation in the frequency domain and multipath-related structure in the delay domain. A fixed feature extractor may capture such effects only implicitly, whereas ACL performs lightweight axis-wise correction that can emphasize informative components and suppress less reliable ones}. Accordingly, in TDD systems, ACL is applied only along the temporal dimension, since prediction is intra-band and mainly governed by temporal evolution. In FDD systems, ACL is applied along both temporal and subcarrier/tap dimensions, because the model must address both temporal dynamics and cross-band spectral mismatch.

For a general case $\mathbf{X} \in \RR^{|\mathcal{T}| \times 2N_{\mathrm{sc}}}$, the ACL is formulated as
  \begin{equation}
    \begin{aligned}
    \mathbf{X}^{\mathrm{ACL1}} [:, k] & = \mathrm{MLP}_1(\mathbf{X}[:, k]) \oplus \mathbf{X} [:, k],  \forall k \in \{1, 2, \cdots, 2N_{\mathrm{sc}} \}  \\
    \mathbf{X}^{\mathrm{ACL2}} [t, :] & = \mathrm{MLP}_2(\mathbf{X}^{\mathrm{ACL1}} [t, :]) \oplus \mathbf{X}^{\mathrm{ACL1}} [t, :],  \forall t \in \mathcal{T}\\
    \mathrm{MLP}_1:\, & \RR^{|\mathcal{T}|}  \rightarrow \RR^{|\mathcal{T}|} \\
    \mathrm{MLP}_2:\, & \RR^{2N_{\mathrm{sc}}}  \rightarrow \RR^{2N_{\mathrm{sc}}}
    \end{aligned}
    \label{eq:arl_x_f}
    \tag{11}
  \end{equation}
where $\mathrm{MLP}_1$ and $\mathrm{MLP}_2$ are multilayer perceptrons, with the number of layers, hidden dimensions, and activation functions treated as tunable hyperparameters. In the first line of Eq.~\eqref{eq:arl_x_f}, the same $\mathrm{MLP}_1$ is shared across all $k$, such that each historical sequence $\mathbf{X}[:,k]$ is processed along the time axis. Therefore, ACL1 performs lag-dependent calibration of the historical CSI, allowing the model to learn how different time lags contribute to future prediction. \response{This is particularly important in TDD CSI prediction, where the historical and future channels lie in the same band, and the main issue is not spectral mismatch but the unequal predictive reliability of different historical lags.} The corrected representations are obtained by combining the MLP outputs with the original inputs using an element-wise operation $\oplus$ (addition or multiplication), which is treated as a tunable hyperparameter. \response{This mechanism enables the model to adaptively refine the learned representation along physically meaningful axes of CSI, namely temporal evolution and frequency/tap structure, while preserving the input dimensionality.} For TDD, the final corrected representation is $\mathbf{X}^{\mathrm{ACL}}_f = \mathbf{X}^{\mathrm{ACL1}}_f$, while for FDD, it is $\mathbf{X}^{\mathrm{ACL}}_f = \mathbf{X}^{\mathrm{ACL2}}_f$. Similarly, $\mathbf{X}^{\mathrm{ACL}}_d$ is derived in an analogous manner. Therefore, ACL does not change the dimensionality of the input, i.e., $\mathbf{X}^{\mathrm{ACL}}_f,\mathbf{X}^{\mathrm{ACL}}_d \in \RR^{|\mathcal{T}| \times 2N_{\mathrm{sc}}}$. \response{The physical intuition and learned behavior of ACL are further supported by intermediate-feature visualizations and targeted ablation results in Appendix~\ref{sec:appendix-ablation-acl}.}

\paragraph{ShuffleNet Block for Feature Extraction}
Inspired by \cite{Zhang_2018_CVPR, Ma_2018_ECCV}, the ShuffleNet Block is employed to perform efficient and expressive feature extraction on each input tensor $\mathbf{X}$ obtained from the ACL. \response{In CSI prediction, the intermediate representation contains structured local patterns across time and subcarriers, while useful dependencies must also be exchanged across learned feature channels. Compared with alternative lightweight backbones such as MobileNet, ShuffleNet organizes channel interaction through grouped point-wise transformations followed by channel shuffle, enabling progressive cross-group feature exchange rather than relying on dense point-wise channel mixing within each block \cite{Zhang_2018_CVPR, Ma_2018_ECCV, howard2017mobilenetsefficientconvolutionalneural, sandler2019mobilenetv2invertedresidualslinear}. This mechanism is empirically better matched to the present feature extraction stage, as further validated by the ablation study in Section~\ref{sec:ablation-study}.} The input is first reshaped into $\mathbb{R}^{2 \times |\mathcal{T}| \times N_{\mathrm{sc}}}$ and projected into $\rho$ feature maps using $\mathrm{Conv1d}$, resulting in $\mathbf{X}_{\rho} \in \mathbb{R}^{\rho \times |\mathcal{T}| \times N_{\mathrm{sc}}}$. A sequence of ShuffleNet Blocks is then applied to extract higher-level representations. 

Each ShuffleNet Block integrates several lightweight yet synergistic operations:
\textit{Point-wise convolution (PW):} A $\mathrm{Conv1d}$ grouped convolution with group size $\eta$ is first applied, enabling intra-group cross-channel interaction while remaining computationally efficient.
\textit{Channel Shuffle (CS):} A permutation step redistributes channels across groups, allowing later grouped convolutions to incorporate information from diverse groups, thereby enriching feature representation.
\textit{Depth-wise convolution (DW):} After shuffling, a depth-wise convolution with kernel size $\mu \times \mu$ is applied, where each channel is convolved independently to capture spatial structure within channels while preserving their count. This operation offers spatial expressiveness with low computational cost.
\textit{Second point-wise convolution:} A final $\mathrm{Conv1d}$ grouped convolution further refines intra-group features post depth-wise processing.
The overall feature extraction procedure can be expressed as Eq.~\eqref{eq:shuffle_block_fe}, resulting in $\mathbf{X}^{\mathrm{FE}} \in \mathbb{R}^{\rho \times |\mathcal{T}| \times N_{\mathrm{sc}}}$. To adaptively emphasize informative channels, a \textit{Squeeze-and-Excitation (SE)} module is applied to $\mathbf{X}^{\mathrm{FE}}$, producing channel attention weights $\mathbf{X}^{\mathrm{SE}} \in \mathbb{R}^{\rho \times 1 \times 1}$ as in Eq.~\eqref{eq:shuffle_block_se}. Finally, the channel-wise Hadamard product ($\odot$) combines $\mathbf{X}^{\mathrm{SE}}$ and $\mathbf{X}^{\mathrm{FE}}$ to yield the refined feature map $\mathbf{X}^{\mathrm{SB}} \in \mathbb{R}^{\rho \times |\mathcal{T}| \times N_{\mathrm{sc}}}$ as shown in Eq.~\eqref{eq:shuffle_block_sb}.  

\begin{subequations}
  \label{eq:shuffle_block_input_output}
  \begin{align}
    \mathbf{X}^{\mathrm{FE}} & = \mathrm{PW} ( \mathrm{DW} ( \mathrm{CS}(\mathrm{PW} (\mathbf{X}_{\mathbf{\rho}}) ) ) ) \label{eq:shuffle_block_fe} \\
    \mathbf{X}^{\mathrm{SE}} & = \mathrm{SE} (\mathbf{X}^{\mathrm{FE}}) \label{eq:shuffle_block_se} \\
    \mathbf{X}^{\mathrm{SB}} & = \mathbf{X}^{\mathrm{SE}} \odot \mathbf{X}^{\mathrm{FE}} \label{eq:shuffle_block_sb}
  \end{align}
\end{subequations}

After that, the resulting $\mathbf{X}^{\mathrm{SB}}_f, \mathbf{X}^{\mathrm{SB}}_d \in \RR^{|\mathcal{T}| \times 2N_{\mathrm{sc}}}$ are obtained by projecting the outputs of the ShuffleNet Blocks back to the original space using $\mathrm{Conv1d}$ convolutions and then concatenating the real and imaginary parts. The delay and frequency representation are then added to feed into the subsequent modules, $\mathbf{X}^{\mathrm{SB}} = \mathbf{X}^{\mathrm{SB}}_f + \mathbf{X}^{\mathrm{SB}}_d \in \RR^{|\mathcal{T}| \times 2N_{\mathrm{sc}}}$.

\paragraph{Position Embedding \& Transformer Encoder}
A Transformer encoder is employed to model long-range and non-local temporal dependencies in the CSI sequence. Through multi-head self-attention, it adaptively aggregates information across all historical time steps. Due to the permutation-invariant nature of the Transformer architecture, it lacks an inherent sense of sequence order. Position embedding (PE) is therefore crucial for incorporating relative positional information. Let $\gamma$ denote the latent dimension of the Transformer, $u$ the index along the latent dimension, and $v$ the position index of the input sequence. The position embedding $\mathbf{X}^{\mathrm{PE}} \in \RR^{|\mathcal{T}| \times \gamma}$ is defined as
\begin{equation}
  \mathbf{X}^{\mathrm{PE}} (u, v) = \begin{cases}
    \sin \left( \frac{v}{|\mathcal{T}|^{u/\gamma}} \right), & \, u = 0, 2, \cdots, 2 \lfloor \gamma/2 \rfloor \\
    \cos \left( \frac{v}{|\mathcal{T}|^{(u-1)/\gamma}} \right), & \, u = 1, 3, \cdots, 2 \lfloor \gamma/2 \rfloor -1
  \end{cases}
  \label{eq:position_embedding}
\end{equation}

Simultaneously, token embeddings (TE) are obtained by projecting $\mathbf{X}^{\mathrm{SB}}$ into the latent space of dimension $\gamma$ using $1\times1$ convolutions, as shown in \eqref{eq:token_embedding_conv}. Following the standard Transformer approach \cite{vaswani-2017}, the position embeddings are added to the token embeddings to form the input to the Transformer encoder. As illustrated in \eqref{eq:transformer}, the resulting embedded sequence is passed through a stack of Transformer encoder layers, yielding the final representation of the input CSI sequence.
\begin{subequations}
  \label{eq:token_embedding}
  \begin{align}
    \mathbf{X}^{\mathrm{TE}} & = \mathrm{Conv1d} (\mathbf{X}^{\mathrm{SB}}) \in \RR^{|\mathcal{T}| \times \gamma} \label{eq:token_embedding_conv} \\
    \mathbf{X}^{\mathrm{TF}} & = \mathrm{Transformer} (\mathbf{X}^{\mathrm{TE}} + \mathbf{X}^{\mathrm{PE}})  \label{eq:transformer}
  \end{align}
\end{subequations}

\paragraph{Prediction Module}

The final prediction module comprises two MLPs that transform the learned CSI embeddings into the predicted CSI sequence. The first MLP maps the Transformer embeddings from the latent dimension back to the original subcarrier dimension, $\RR^{\gamma} \rightarrow \RR^{2N_{\mathrm{sc}}}$. The second MLP projects the historical time dimension into the predicted time dimension, $\RR^{|\mathcal{T}|} \rightarrow \RR^{|\mathcal{P}|}$. The final predicted CSI sequence is reconstructed by converting the real-valued tensor back into a complex-valued tensor through stacking the real and imaginary parts.
\section{Experiments}
\label{sec:experiments}
This section outlines the experimental setup for CSI prediction, covering data generation for training and testing, baseline models, and evaluation metrics.
\subsection{Dataset}
\label{sec:experiment-dataset}

\begin{table}[ht]
  \centering
  \caption{\textbf{System Configuration}}
  \label{tab:system-config}
  \begin{tabular}{cc}
  \toprule
  \textbf{Parameter}               & \textbf{Value}            \\
  \midrule
  BS antenna              & dual-polarized [4,4] UPA    \\
  UE antenna              & Single omnidirectional antenna  \\
  Carrier frequency       & 2.4\,GHz                           \\
  Subcarriers (UL / DL)   & 300 for UL/DL                  \\
  Subcarrier spacing      & 30\,kHz                           \\
  \bottomrule
  \end{tabular}
\end{table}

\begin{table}[ht]
  \centering
  \caption{\textbf{Training Dataset Configuration}}
  \label{tab:training-config}
  \begin{tabular}{cc}
  \toprule
  \textbf{Parameter}               & \textbf{Value}            \\
  \midrule
  Channel models  & CDL-A, CDL-C, CDL-D                                        \\
  Delay spreads   & 30, 100, 300\,ns                                           \\
  User velocities & 1, 10, 30\,m/s                \\
  Noise Type       & AWGN                                    \\
  Noise SNR       & uniformly distributed in  [0,\,25]\,dB \\
  \bottomrule
  \end{tabular}
\end{table}

\begin{table}[ht]
\centering
\caption{\textbf{Testing Dataset Configurations}}
\label{tab:testing-config}
\renewcommand{\arraystretch}{1.1}
\begin{tabularx}{\linewidth}{@{} l  >{\raggedright\arraybackslash}p{0.34\linewidth}
                                  >{\raggedright\arraybackslash}X @{}}
\toprule
\textbf{Test type} & \textbf{Parameter} & \textbf{Value} \\
\midrule
\multicolumn{3}{@{}l}{\textbf{Regular}} \\
\cmidrule(lr){1-3}
 & Channel models & CDL-A/C/D \\
 & Delay spreads  & 30, 100, 300\,ns \\
 & User velocities & 1, 10, 30\,m/s \\
 & Noise Type     & AWGN \\
 & Noise SNR      & [0, 5, 10, 15, 20, 25]\,dB \\
\midrule
\multicolumn{3}{@{}l}{\textbf{Robustness}} \\
\cmidrule(lr){1-3}
 & Channel models & CDL-A/C/D \\
 & Delay spreads  & 30, 100, 300\,ns \\
 & User velocities & 1, 10, 30\,m/s \\
 & Noise Type     & phase, burst, packet-drop \\
 & Noise SNR      & [10, 15, 20, 25]\,dB for both phase and burst \\
 & Packet drop probability & [0.01, 0.02, \ldots, 0.10] \\
\midrule
\multicolumn{3}{@{}l}{\textbf{Generalization}} \\
\cmidrule(lr){1-3}
 & Channel models & CDL-A/B/C/D/E \\
 & Delay spreads  & 30, 50, 100, 200, 300, 400\,ns \\
 & User velocities & 3, 6, \ldots, 45\,m/s; 1, 10\,m/s \\
 & Noise Type     & AWGN \\
 & Noise SNR      & [0, 5, 10, 15, 20, 25]\,dB \\
\bottomrule
\end{tabularx}
\end{table}

\paragraph{Data generation and system configurations}
This study uses the Sionna library \cite{sionna} to synthesize time-varying CSI. The system configurations are summarized in Table~\ref{tab:system-config}. Specifically, the system configured an OFDM link with a dual-polarized $[4,4]$ UPA at the BS ($N_{\mathrm{tr}}=4\times4\times2$) and a single omnidirectional receive antenna at the UE ($N_{\mathrm{re}}=1$) with the carrier frequency 2.4\,GHz. Both TDD and FDD duplexing modes are considered. The OFDM grid consists of 750 subcarriers with 30 kHz subcarrier spacing—300 assigned to the UL and 300 to the DL—corresponding to 9 MHz bandwidth for each link. A guard band of 150 subcarriers (4.5 MHz) separates the UL and DL bands. CSI reports are spaced by 5 slots, corresponding to 2.5 ms under the considered system parameters. The historical CSI window and prediction horizon lengths are considered as $|\mathcal{T}| = 16$, and $|\mathcal{P}|=4$. As a result, each CSI snapshot is represented by a $32 \times 1 \times 750$ complex tensor, corresponding to the $N_{\mathrm{tr}}$, receive antennas $N_{\mathrm{re}}$, and total number of subcarriers ($2N_{\mathrm{sc}}+150$). After separating the UL and DL subcarriers, the model input and prediction target are shaped as $32 \times 1 \times 16 \times 300$ and $32 \times 1 \times 4 \times 300$, respectively. These system configurations are carefully aligned with the 3GPP specifications \cite{3gpp-38211}, ensuring consistency with standardized practices and relevance to real-world deployment settings. 

A large set of scenarios are considered for training and evaluation of the models under various channel conditions and noise types. The set of generated scenarios is denoted as:
\begin{equation}
  \begin{aligned}
    \Scenarios = \Bigl\{ & [\speed, \delay, \cm, \noisetype, \noisedegree] \,\Big|\, 
    \speed \in \mathcal{V}, \; \delay \in \Sigma, \\
    & \cm \in \mathfrak{M}, \; \noisetype \in \mathfrak{N}, \; \noisedegree \in \mathfrak{D} 
    \Bigr\}
    \label{eq:scenarios}
  \end{aligned}
\end{equation}
which enumerates the combinations of user speed ($\speed$), delay spread ($\delay$), and channel model ($\cm$), along with the noise type ($\noisetype$) and noise degree ($\noisedegree$) that control the characteristics of additional synthesized noise. The noise component is introduced to emulate real-world channel conditions. Specifically, $\noisedegree$ denotes the packet drop probability for packet drop noise, and the SNR for other noise types. These parameters govern the statistical properties of the synthesized CSI sequence \eqref{eq: definition of CSI}.

\paragraph{Training set}
Table~\ref{tab:training-config} summarizes the training configurations. The 3GPP TR~38.901 \cite{3gpp-38901} CDL-A, CDL-C, and CDL-D models with delay spreads $[30,100,300]$\,ns and user velocities $[1,10,30]$\,m/s are adopted. For each scenario 1{,}000 samples are generated, yielding $3\times3\times3=27$ configurations and 27{,}000 samples in total. To model noisy CSI observations, AWGN (Appendix \ref{sec:additive-white-gaussian-noise}) is added to the historical inputs with SNR uniformly distributed in $[0,25]$\,dB. Note that the noise is only added to the historical sequence, and the clean future CSI is used for training and evaluation. The training set is further split into a training set and a validation set with a ratio of 9:1.

\paragraph{Testing suite}
The testing suite comprises three scenarios (Table~\ref{tab:testing-config}). 
\begin{itemize}
  \item \Regular{} retains the same 27 configurations and AWGN setting as the training data. For each configuration, AWGN with SNRs $[0, 5, 10, 15, 20, 25]$\,dB is evaluated, resulting in a total of $27 \times 6 = 162$ scenarios. For each scenario, 100 samples are generated (this sample size is consistently used across all subsequent testing scenarios and will not be restated), resulting in a total of 16,200 pairs of historical and future CSI instances. This setup provides an exact in-distribution evaluation of model performance.
  \item \Robustness{} keeps the same 27 configurations as training, but replaces the AWGN assumption with three realistic noise types. This setting evaluates the robustness of the model against realistic noises. The total number of scenarios in \Robustness{} is $27 \times (4 + 4 + 10) = 486$, detailed as follows:
  \begin{itemize}
    \item \textbf{Phase noise (Appendix~\ref{sec:phase-noise})}: simulates irregular fluctuations in the channel phase. Evaluated at $[10, 15, 20, 25]$\,dB. 
    \item \textbf{Burst noise (Appendix~\ref{sec:burst-noise})}: models short, high-amplitude, pulse-like disturbances in the channel. Evaluated at $[10, 15, 20, 25]$\,dB. 
    \item \textbf{Packet drop noise (Appendix~\ref{sec:packet-drop-noise})}: represents random erasures of CSI matrices within the sequence. Evaluated at drop rates $[0.01, 0.02, \ldots, 0.10]$.
  \end{itemize}
  \item \Generalization{} retains the AWGN setting used in training but extends the channel model, delay spread, and user velocity configurations to evaluate the model's generalization capability. The final dataset includes 5 channel models, 6 delay spreads, and 17 user velocities. For each of the 510 ($5 \times 6 \times 17$) configurations, AWGN is evaluated at SNRs $[0, 5, 10, 15, 20, 25]$\,dB. Accordingly, the total number of scenarios in \Generalization{} is $5 \times 6 \times 17 \times 6 = 3{,}060$.
  \begin{itemize}
    \item \textbf{Channel model}: According to \cite{sionna, 3gpp-38901}, CDL-A/B/C correspond to Non-Line-of-Sight (NLOS) channels, where signals reach the receiver through reflection, scattering, and diffraction, whereas CDL-D/E represent Line-of-Sight (LOS) channels characterized by a dominant direct path between the BS and UE. In the experimental setup, CDL-A/C/D are included during training while CDL-B/E are reserved for evaluation, enabling the model to learn from both channel types and to demonstrate its generalization across NLOS and LOS conditions.
    \item \textbf{Delay spread}: The training dataset includes delay spreads of 30, 100, and 300 ns, while additional values of 50, 200, and 400 ns are used for evaluation. This allows assessment with both within-range values (50 ns and 200 ns) and an outside-range value (400 ns).
    \item \textbf{User velocity}: In addition to the training velocities of 1, 10, and 30 m/s, the dataset includes velocities ${3, 6, 9, \dots, 45}$ m/s. This setup enables evaluation both within the training range (velocities below 30 m/s) and outside the training range (velocities above 30 m/s), similar to the delay-spread configuration.
  \end{itemize}
\end{itemize}
\paragraph{\response{Data Partitioning and Leakage Prevention}}
\response{To ensure reproducible evaluation, fixed random seeds are used in dataset generation and in the train/validation split. Each \dataset{} sample is generated as an independent fixed-length CSI sequence containing both the historical input window and the prediction horizon, rather than being extracted from a long trajectory via overlapping sliding windows. This construction prevents temporal overlap between training, validation, and test samples. In addition, the train/validation split is performed only within the training partition, while the \Regular{}, \Robustness{}, and \Generalization{} partitions are loaded separately and used only for final evaluation. For the \Generalization{} track, unseen channel conditions are explicitly reserved for testing. For the \Regular{} track, the same channel configurations and AWGN setting as in training are retained, but the underlying samples remain strictly disjoint from the training/validation data. For the \Robustness{} track, the same base channel configurations are retained to isolate robustness to corruption, but the corruption family is changed from the AWGN used in training to phase noise, burst noise, and packet drop noise; therefore, the robustness evaluation does not reuse the training corruption type.}

\subsection{Baseline Models}
\label{sec:experiment-baseline-models}

To evaluate the proposed model, this study compares against the following baselines:
\begin{itemize}
  \item \textbf{LLM4CP \cite{liu-2024}:} Together with CSI-BERT2 \cite{zhao2024mining}, LLM4CP is a representative example of recent CSI predictors that leverage (pre-trained) large language model (LLM) layers. LLM4CP is included due to reported strong generalization and publicly available code.
  \item \textbf{STEMGNN \cite{mourya-2024}:} An advanced deep learning architecture that incorporates a graph neural network (GNN) component to capture spatiotemporal dependencies. The authors of \cite{mourya-2024} applied STEMGNN to CSI prediction and demonstrated strong performance. It should be noted that \cite{mourya-2024} integrated the encoder-decoder structure from STNet \cite{Mourya_2023}, allowing STEMGNN to operate in the latent space between the encoder and decoder. For a fair comparison, this study applies the STEMGNN predictor directly, without incorporating the additional encoder-decoder structure. 
  \item \textbf{RNN \cite{jiang-2019A}:} A recurrent neural network baseline that models temporal dependencies; among the earliest deep learning approaches introduced for CSI prediction, showing promising results.
  \item \textbf{CNN \cite{safari-2020}:} A convolutional baseline that leverages structural similarity between CSI tensors and images, particularly for FDD scenarios.
  \item \textbf{No Prediction (NP):} A naive baseline that repeats the last observed CSI across the four-step prediction horizon (i.e., persistent model). This provides a direct measure of channel aging and offers an intuitive understanding of task difficulty across scenarios.
  \item \response{\textbf{AR \cite{duel2007fading}:} A vector autoregressive baseline that models the temporal evolution within the same band from past CSI to future CSI through linear dependence on previous observations.}
  \item \response{\textbf{PAD \cite{yin-2020}:} A Prony-based predictor operating in the angular-delay domain, which exploits the inherent angle-delay-Doppler structure of multipath channels. It is an advanced model-based channel prediction method proposed to mitigate the curse of mobility in TDD systems.}
  \item \response{\textbf{Wiener \cite{arnold2019enabling}:} A classical Wiener filter baseline that estimates CSI in the time and frequency domains, commonly referred to as the linear minimum mean square error (LMMSE) predictor \cite{li2000pilot}.}
\end{itemize}

\response{Note that the AR and PAD approaches are evaluated only on the TDD task, whereas Wiener is evaluated on both TDD and FDD tasks. Moreover, the PAD implementation used in this work incorporates a safe fallback mechanism. Further details and discussions are provided in the Appendix~\ref{sec:appendix-model-based-baselines}. All methods (if applicable) were trained on the same dataset and underwent extensive hyperparameter tuning, as explained in the following sections.}

\subsection{Training Configuration}\label{sec:training-configuration}

\paragraph{Learning-based baselines}
For the deep learning baselines, hyperparameter tuning is performed with Optuna \cite{optuna_2019} for \Model{} and all baselines in both TDD and FDD modes. For each (model, duplexing) pair, an Optuna study explores a predefined search space for up to 30 hours. Depending on the model's complexity, between 1 and 3 NVIDIA H200 GPUs are allocated in parallel to accelerate the search. Each trial is budgeted for 10-20 training epochs, with the length chosen empirically according to the typical convergence speed of the model. Trials are evaluated on the validation set, recording accuracy (validation NMSE) together with efficiency in terms of FLOPs. The detailed hyperparameter ranges and training configurations are provided in Appendix~\ref{sec:appendix-training-configuration}.

After tuning, all trial outcomes are mapped to the accuracy-efficiency plane, and the set of \emph{non-dominated} configurations is extracted as the Pareto frontier. Every configuration on this frontier is then \emph{retrained from scratch} under a full schedule: at most 50 epochs with early stopping. In practice, models typically converge and stop between 15 and 30 epochs, so the actual training duration does not deviate substantially from the trial budgets. For each (model, duplexing) setting, the final checkpoint is the one achieving the lowest validation NMSE among these full trainings. The Pareto frontier approach is employed to explicitly capture the trade-off between computational complexity and predictive performance.

\paragraph{Model-based baselines}
\response{AR, Wiener, and PAD baselines use the same data pipeline as the learning-based baselines. For AR (TDD), a per-subcarrier vector AR model of order $p$ is fitted by ridge-regularized least squares; the order used at inference is selected by minimizing validation NMSE. For Wiener (TDD/FDD), a per-subcarrier ridge-regularized linear map from flattened history to flattened target is fitted in closed form. Unlike other baselines, PAD does not rely on an offline parameter-estimation stage; its Prony coefficients are identified online from the currently available historical CSI and used immediately for recursive prediction \cite{yin-2020}.}

\subsection{Evaluation Metrics}
\label{sec:evaluation-metrics}

In order to evaluate the performance of the proposed model on \dataset{}, the following metrics are employed:
\begin{itemize}
  \item \textbf{NMSE} As shown in \eqref{eq:NMSE}, NMSE serves as a straightforward numerical indicator of prediction performance. Moreover, the increasing or decreasing trend of NMSE under changes in the channel conditions offers intuitive insights into the mechanisms by which different factors influence prediction performance.

  \item \textbf{Spectral Efficiency (SE)} In addition to NMSE, SE (Appendix~\ref{sec:appendix-se}) quantifies the practical performance of the predicted CSI in terms of achievable data rate. It measures how prediction accuracy translates into overall system efficiency.
   
  \item \textbf{Rank Score and the Percentage of Models with Rank 1} The evaluation dataset contains thousands of diverse scenarios that vary substantially in difficulty. As a result, NMSE values are not directly comparable across scenarios—for instance, the NMSE on CDL-A may be an order of magnitude larger than on CDL-D under identical settings (Table~\ref{tab:performance-channel-models-regular}). Consequently, averaging NMSE across scenarios may introduce bias toward the harder cases and complicate fair comparisons (see Sections~\ref{sec:nmse-based-evaluation} and \ref{sec:robustness-performance}). To mitigate this issue, the scenario-wise rank distribution is introduced to better reflect model performance within specific subsets of scenarios, and is defined as:
  \begin{equation}
    \mathrm{rank} (\pi, \mathbf{s}) \in \{ 1, \ldots, |\Pi| \},
    \label{eq:scenario-wise-model-rank}
  \end{equation}
  where $\pi$ denotes the model, $\mathbf{s}$ the scenario, and $\Pi$ the set of all models. Based on this definition, for a selected subset of scenarios $\mathbf{S'} \subset \Scenarios$, the mean rank score and the percentage of rank-1 occurrences are introduced to summarize overall model performance:

  \begin{subequations}
    \label{eq:mean-rank}
    \begin{align}
      & \mathrm{MeanRank} (\pi, \mathbf{S'}) \\
      & = \frac{1}{|S'|} \sum_{\mathbf{s} \in \mathbf{S'}} \mathrm{rank} (\pi, \mathbf{s}) \in [1, |\Pi|]
    \end{align}
  \end{subequations}

  \begin{subequations}
    \label{eq:mean-rank-score}
    \begin{align}
      & \mathrm{RankScore} (\pi, \mathbf{S'}) \\
      & = |\Pi| - \mathrm{MeanRank} (\pi, \mathbf{S'}) \in [0, |\Pi|-1]
    \end{align}
  \end{subequations}
  \begin{subequations}
    \label{eq:rank-1-percentage}
    \begin{align}
      & \mathbf{P}_{\mathrm{rank1}} (\pi, \mathbf{S'}) \\
      & = \frac{1}{|S'|} \sum_{\mathbf{s} \in \mathbf{S'}} \mathbf{1} \left\{ \mathrm{rank} (\pi, \mathbf{s}) = 1 \right\} \in [0, 1]
    \end{align}
  \end{subequations}
  Here, $\mathbf{S'}$ is assigned to the \Regular{}, \Robustness{}, and \Generalization{} sets, so that $\mathrm{RankScore}$ and $\mathbf{P}_{\mathrm{rank1}}$ reflect performance in the corresponding evaluation tracks. For both metrics, larger values indicate stronger performance.
  \item \textbf{Efficiency} Model efficiency is evaluated in terms of FLOPs, total parameters, inference time, and \response{static memory footprint}. The efficiency score is defined as the normalized improvement relative to the most resource-demanding model on each metric. Let $\mathrm{c}$ denote the cost metric, with
\begin{equation}
  \begin{aligned}
    \mathrm{c} \in \{& \text{FLOPs}, \text{Total Params}, \\ 
    & \text{Inference Time}, \response{\text{Static Memory Footprint}}\}
  \end{aligned}
\end{equation}
The efficiency score is then given by
  \begin{subequations}
    \label{eq:efficiency-score}
    \begin{align}
      & \mathrm{EffScore} (\pi, \mathrm{c}) \\
      & =1-\frac{\mathrm{c}(\pi)}{\max_{\pi \in \Pi} \mathrm{c}(\pi)} \in [0,1].
    \end{align}
  \end{subequations}
  By design, larger values of the efficiency score indicate stronger performance, namely better computational efficiency, consistent with the definitions of $\mathrm{RankScore}$ and $\mathbf{P}_{\mathrm{rank1}}$.
\end{itemize}
\begin{figure*}[!ht]
  \centering
  \subfloat[\Regular{}]{
      \includegraphics[width=0.32\textwidth]{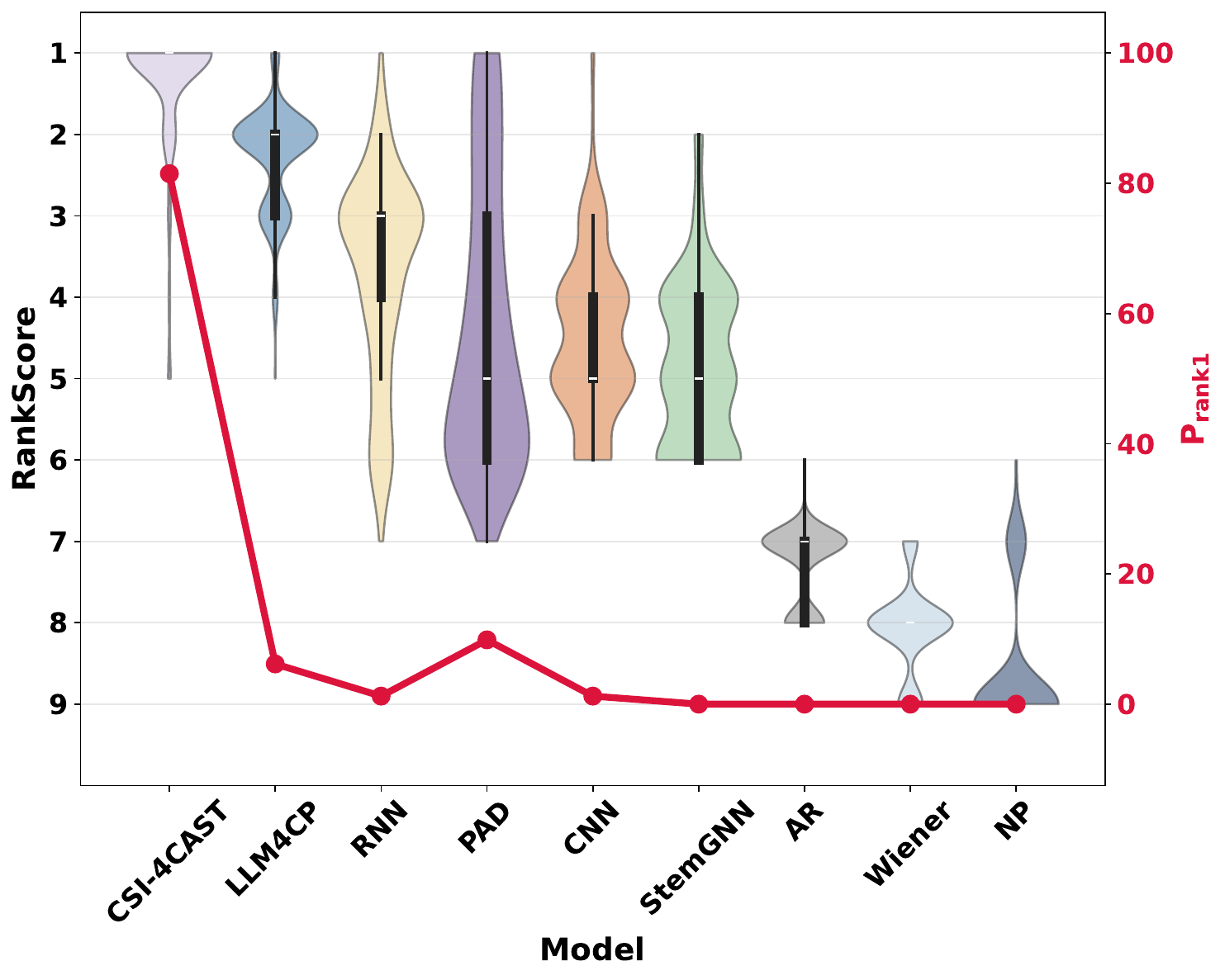}
      \label{fig:violin_regular_tdd}
  }
  \subfloat[\Robustness{}]{
      \includegraphics[width=0.32\textwidth]{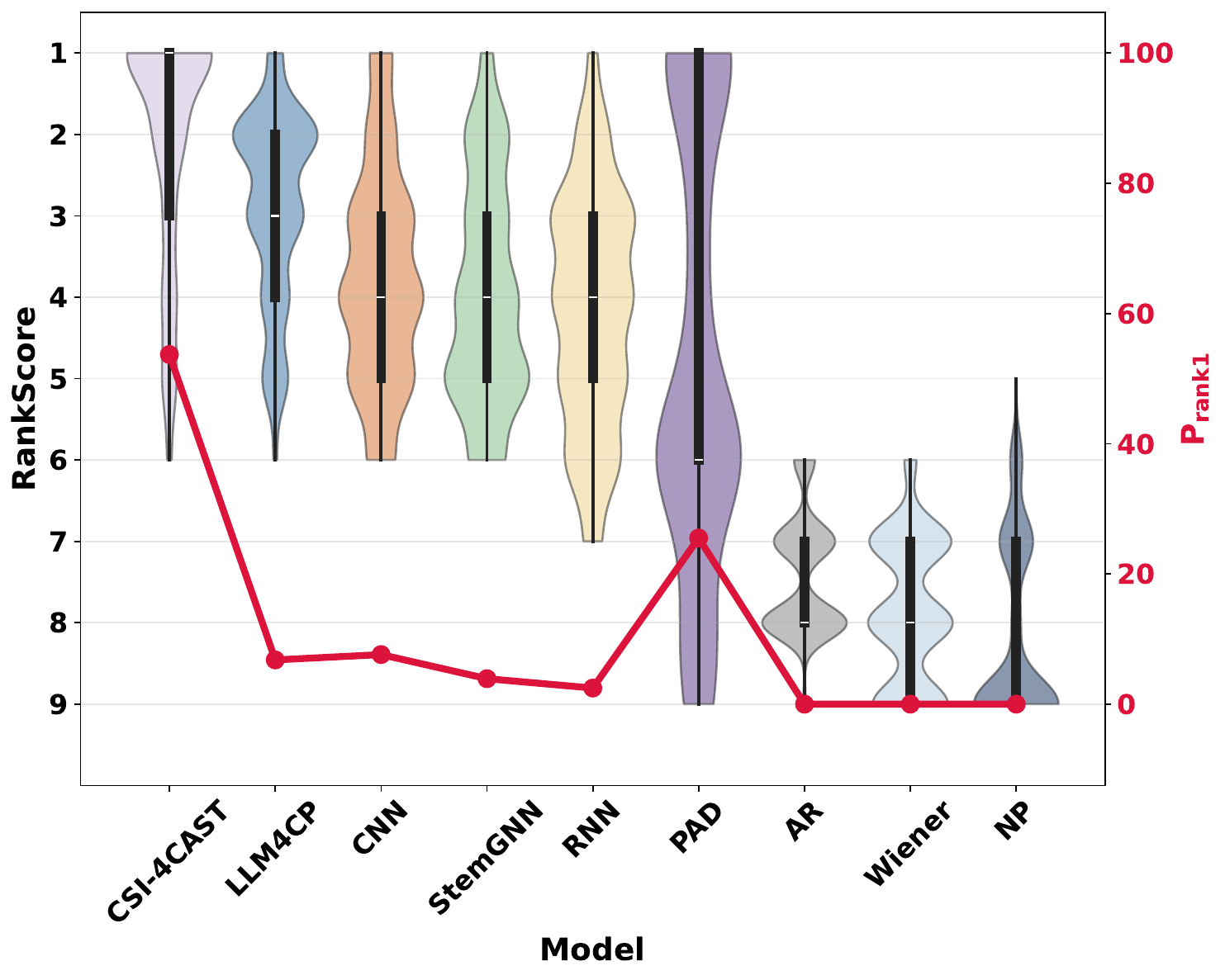}
      \label{fig:violin_robustness_tdd}
  }
  \subfloat[\Generalization{}]{
      \includegraphics[width=0.32\textwidth]{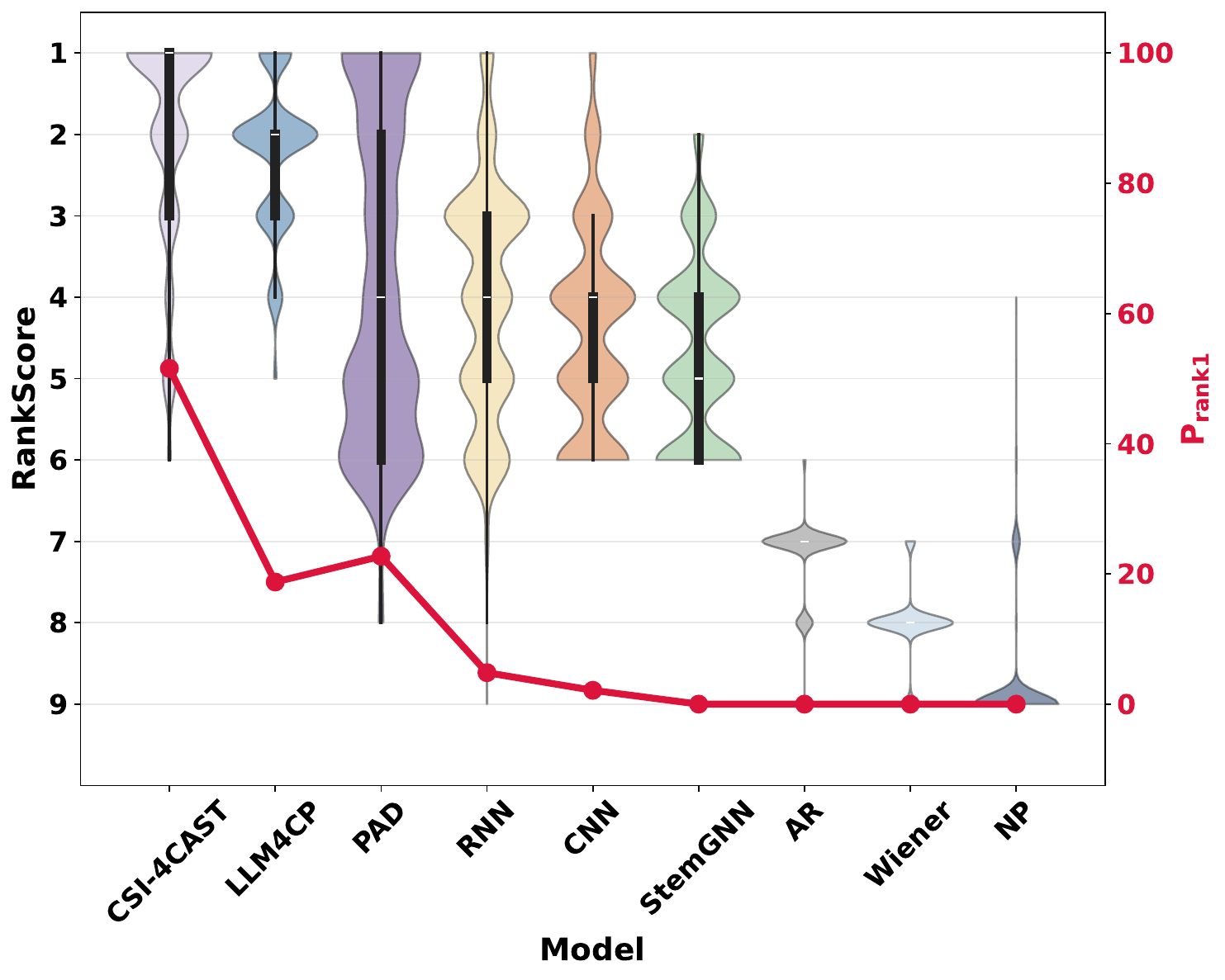}
      \label{fig:violin_generalization_tdd}
  }
  \caption{\textbf{TDD: NMSE rank distribution of \Regular{}, \Robustness{}, and \Generalization{}.} Within each panel, models are ordered left to right by their mean rank, $\mathrm{MeanRank}$ in \eqref{eq:mean-rank} (lower is better). Rank distributions are shown as violin plots, while top-1 percentages, $\mathbf{P}_{\mathrm{rank1}}$ in \eqref{eq:rank-1-percentage}, are plotted as a red line graph.}
  \label{fig:performance-regular-robustness-generalization-tdd}
\end{figure*}

\begin{figure*}[!ht]
  \centering
  \subfloat[\Regular{}]{
      \includegraphics[width=0.32\textwidth]{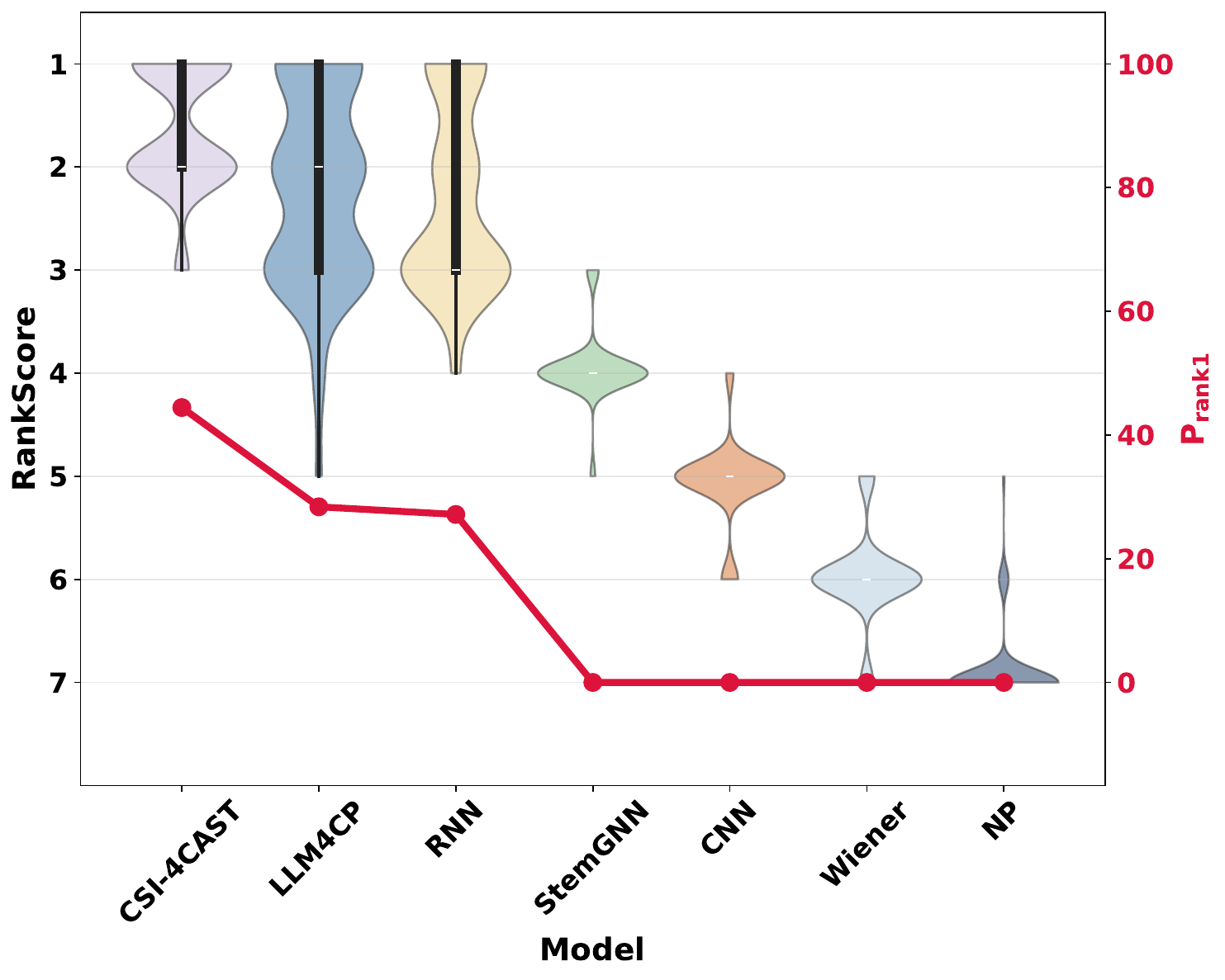}
      \label{fig:violin_regular_fdd}
  }
  \subfloat[\Robustness{}]{
      \includegraphics[width=0.32\textwidth]{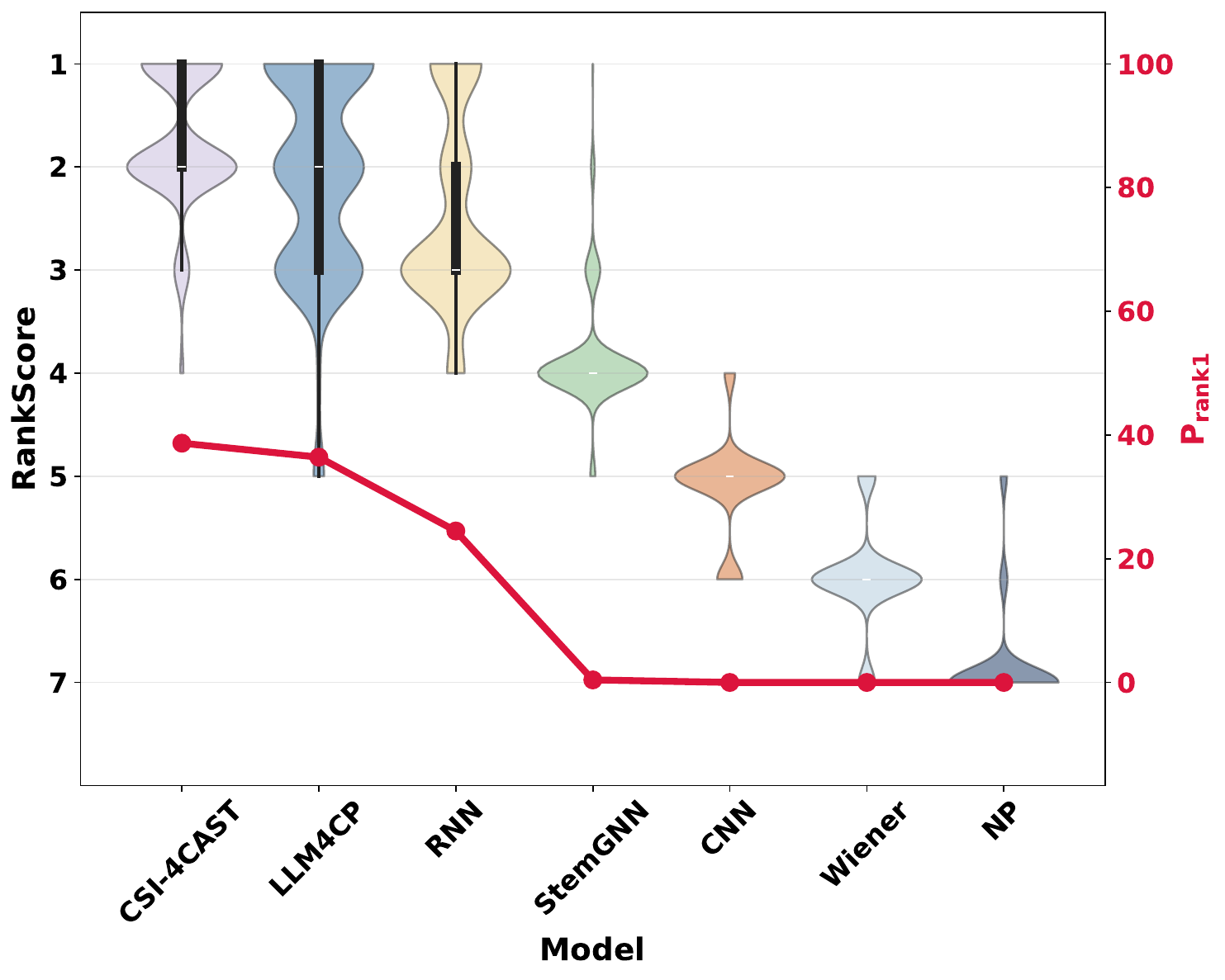}
      \label{fig:violin_robustness_fdd}
  }
  \subfloat[\Generalization{}]{
      \includegraphics[width=0.32\textwidth]{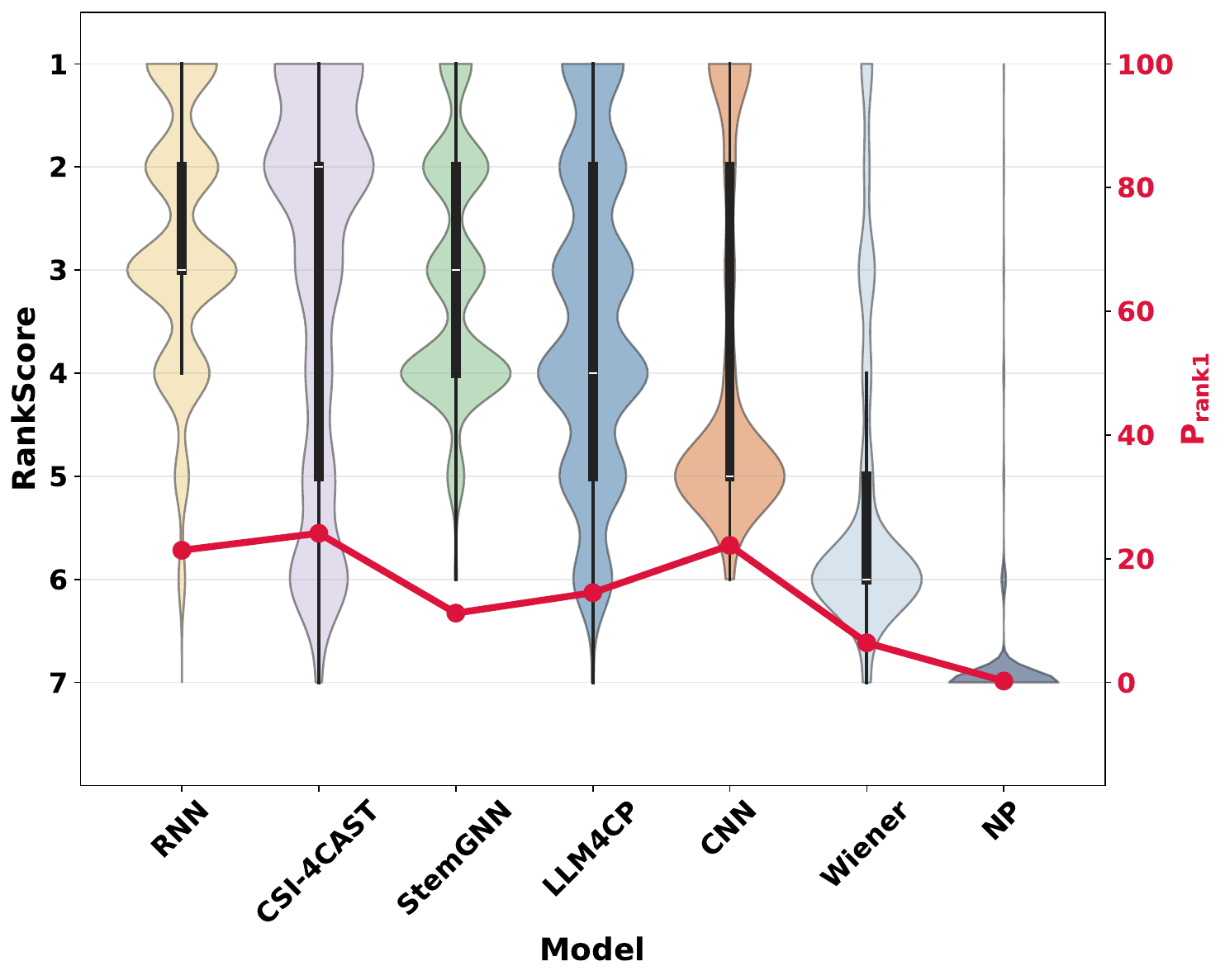}
      \label{fig:violin_generalization_fdd}
  }
  \caption{\textbf{FDD: NMSE rank distribution of \Regular{}, \Robustness{}, and \Generalization{}.} The plotting conventions follow those in Fig.~\ref{fig:performance-regular-robustness-generalization-tdd}.}
  \label{fig:performance-regular-robustness-generalization-fdd}
\end{figure*}

\section{Performance Evaluation}
\label{sec:performance-evaluation}

This section presents the experimental results, comparing the proposed model with baseline methods for CSI prediction across a comprehensive set of testing scenarios. As outlined in Section~\ref{sec:experiment-dataset}, the scenarios generated in \dataset{} are grouped into three sets: testing under the same distribution as training (\Regular{}), testing under realistic noise conditions (\Robustness{}), and testing on unseen scenarios (\Generalization{}). Section~\ref{sec:rank-based-evaluation} reports the rank-based evaluation of the models on the \Regular{}, \Robustness{}, and \Generalization{} sets; Section~\ref{sec:nmse-based-evaluation} presents the NMSE-based evaluation across varying SNRs, user velocities, delay spreads, and channel models; Section~\ref{sec:robustness-performance} focuses on NMSE under various realistic noise conditions; and Section~\ref{sec:overall-performance} provides an overall assessment of the models, considering the trade-off between prediction performance and computational overhead. \response{Section~\ref{sec:ablation-study} presents an ablation study that validates the effectiveness of the proposed design and justifies the choice of each module.}

Both TDD and FDD duplexing modes are evaluated, and the results are reported separately. This separation is necessary due to the substantial differences in the prediction tasks (Section~\ref{sec:problem-definition}) and the independent training of models for the two duplexing modes (Section~\ref{sec:training-configuration}).

\subsection{Rank-Based Evaluation: \Regular{}, \Robustness{}, \& \Generalization{}}
\label{sec:rank-based-evaluation}

This section summarizes the model comparisons across the \Regular{}, \Robustness{}, and \Generalization{} tracks in Fig.~\ref{fig:performance-regular-robustness-generalization-tdd} for TDD and Fig.~\ref{fig:performance-regular-robustness-generalization-fdd} for FDD. For each track, both the scenario-wise rank distribution and the top-1 rank percentage, $\mathbf{P}_{\mathrm{rank1}}$ in \eqref{eq:rank-1-percentage}, are reported. The $\mathrm{MeanRank}$ in \eqref{eq:mean-rank} is also reflected in each panel, with models arranged from left to right in ascending order.

\begin{figure*}[!htbp]
  \centering
  \subfloat[TDD]{
    \includegraphics[width=0.4\textwidth]{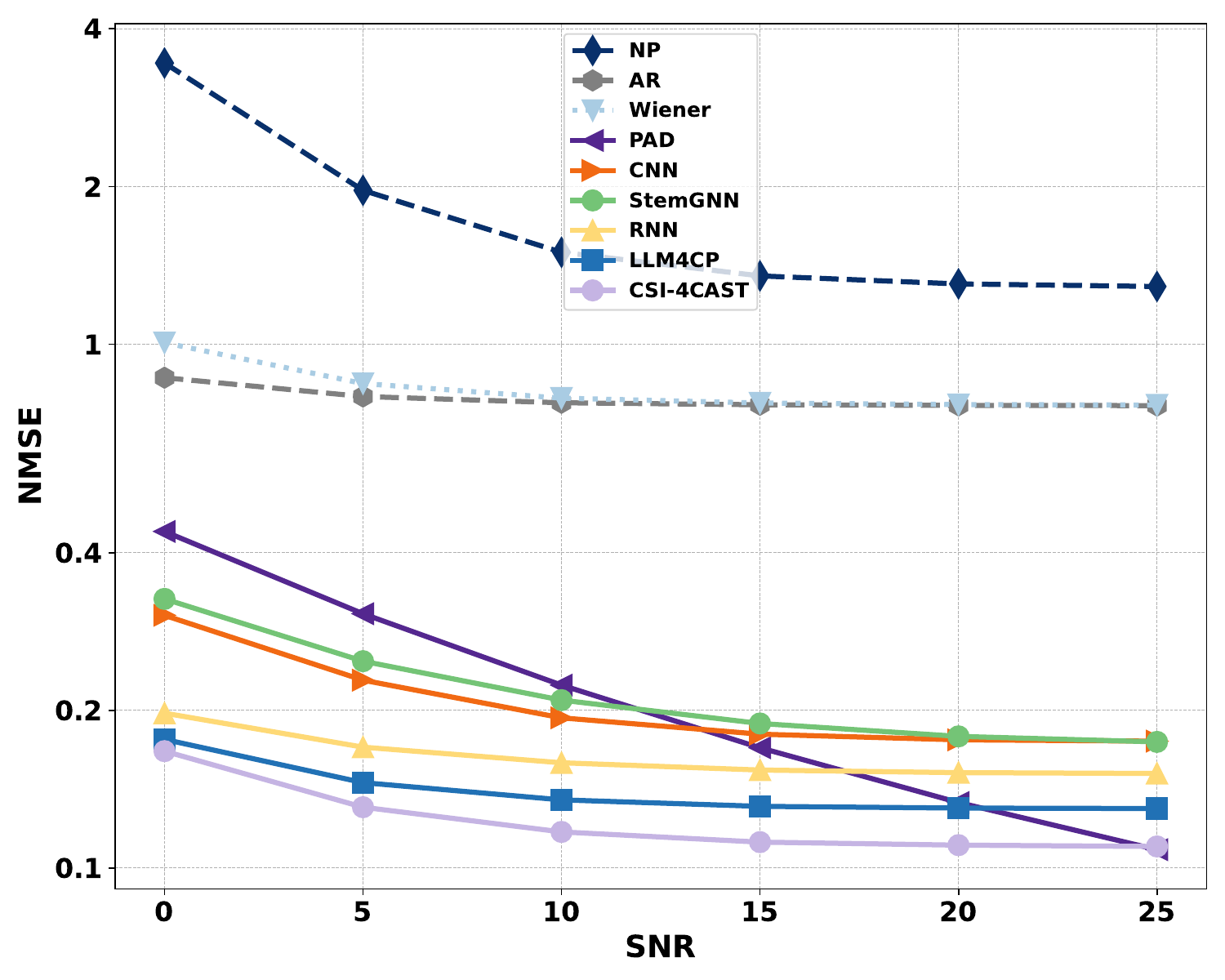}
    \label{fig:tdd_gaussian_across_snr_nmse_regular}
  }\
  \subfloat[FDD]{
    \includegraphics[width=0.4\textwidth]{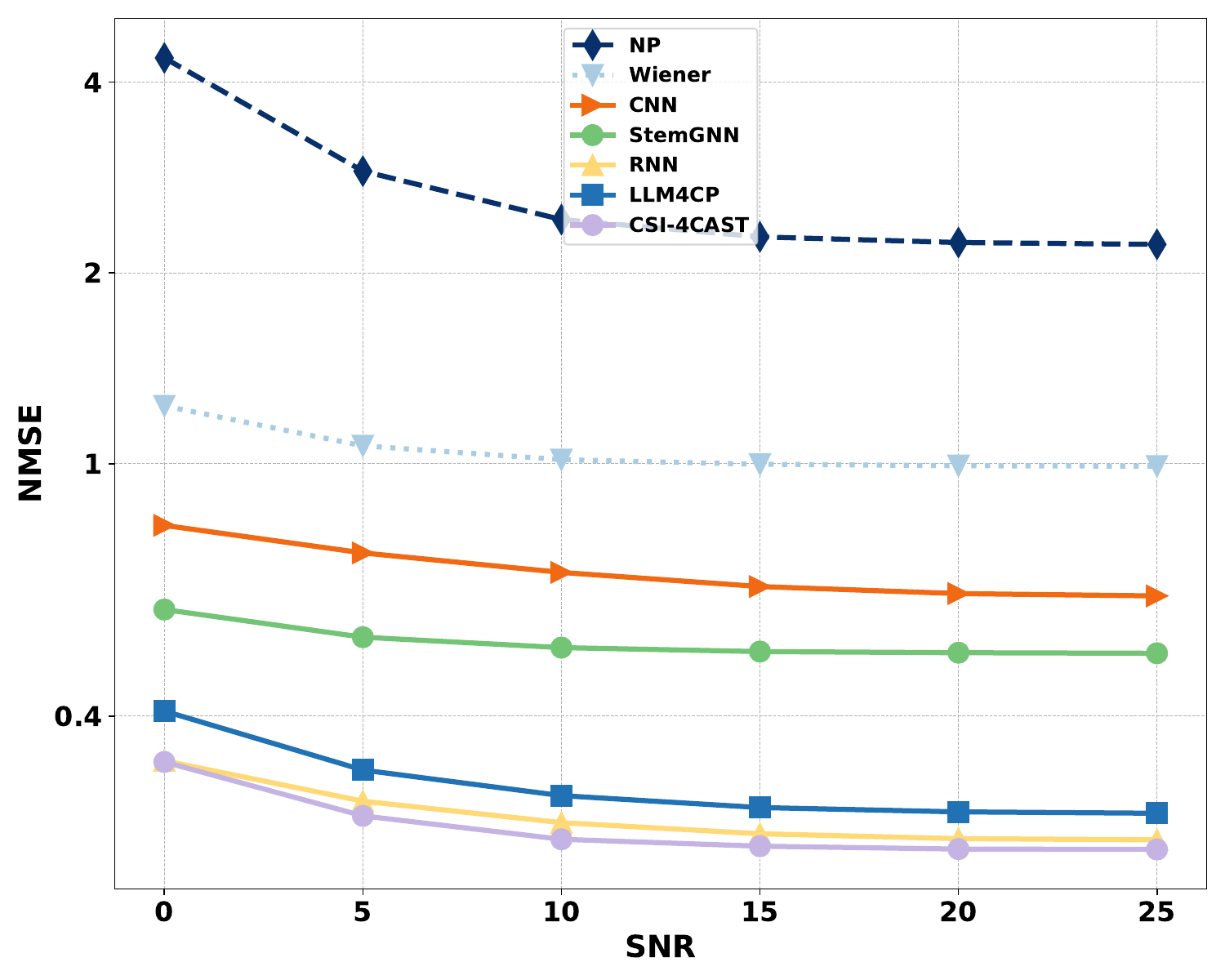}
    \label{fig:fdd_gaussian_across_snr_nmse_regular}
  }
  \caption{\textbf{NMSE under varying SNR (dB).} Model performance under (a) TDD and (b) FDD is evaluated across SNR from 0 to 25~dB using AWGN.}
  \label{fig:performance-awgn-noises-nmse-regular}
\end{figure*}

For TDD system (Fig.~\ref{fig:performance-regular-robustness-generalization-tdd}), the proposed \Model{} achieves the best $\mathrm{MeanRank}$ and $\mathbf{P}_{\mathrm{rank1}}$ across all three scenarios, consistently outperforming the baselines. Under \Regular{} testing, it achieves a $\mathrm{MeanRank}$ of \response{1.31} and the $\mathbf{P}_{\mathrm{rank1}}$ of \response{81.5\%}. Under \Robustness{} testing, the \Model{} attains a $\mathrm{MeanRank}$ of \response{2.07} and a $\mathbf{P}_{\mathrm{rank1}}$ of \response{53.7\%}. For \Generalization{} testing, it achieves a $\mathrm{MeanRank}$ of \response{1.99} and a $\mathbf{P}_{\mathrm{rank1}}$ of \response{51.6\%}. The performance margin over LLM4CP (the next-best model) in $\mathbf{P}_{\mathrm{rank1}}$ is \response{75.3\%, 46.9\%, and 32.8\%} for \Regular{}, \Robustness{}, and \Generalization{}, respectively. Moreover, the proposed \Model{} exhibits narrower and more concentrated rank distributions across all three tracks, highlighting its leading position in all tracks. In contrast, while LLM4CP shows stable rank distributions in \Regular{} and \Generalization{}, its performance in the \Robustness{} track is scattered. \response{PAD, CNN, STEMGNN, and RNN, are generally distributed across ranks 3--6 in all tracks. Simple and lightweight statistical models, such as AR and Wiener, typically rank near the bottom across all tracks and outperform only the NP baseline. For PAD, however, its $\mathbf{P}_{\mathrm{rank1}}$ always exhibits a spike pattern, indicating that $\mathbf{P}_{\mathrm{rank1}}$ is abnormally high relative to its $\mathrm{MeanRank}$. This is because PAD is a strong and highly expressive model when the channel noise is low (i.e., high SNR scenarios), which aligns with its exponential smoothing assumptions. As shown in Fig.~\ref{fig:tdd_gaussian_across_snr_nmse_regular} and Fig.~\ref{fig:tdd_phase_nmse_noise_degree}, PAD achieves the lowest, or nearly the lowest, NMSE at the highest SNR under AWGN noise and phase noise, respectively. Further details are discussed in the subsequent sections and in Appendix~\ref{sec:appendix-pad-safe-fallback}.}

For FDD system (Fig.~\ref{fig:performance-regular-robustness-generalization-fdd}), the proposed \Model{} achieves the highest $\mathrm{MeanRank}$ in both \Regular{} and \Robustness{} scenarios, with a $\mathrm{MeanRank}$ of \response{1.62} and a $\mathbf{P}_{\mathrm{rank1}}$ of \response{44.4\%} in \Regular{} testing, and a $\mathrm{MeanRank}$ of \response{1.71} and a $\mathbf{P}_{\mathrm{rank1}}$ of \response{38.7\%} in \Robustness{} testing. Under \Generalization{} testing, however, all models exhibit significant degradation compared to TDD. Due to the inherent difficulty of FDD prediction, which requires inter-band inference, none of the methods achieve reliable generalization to unseen scenarios, resulting in wide rank distributions across all models. \response{Similar to TDD, Wiener ranks near the bottom in FDD across all tracks, indicating that simple statistical models are insufficient for both duplexing modes.}

The overall comparison between TDD and FDD underscores the greater complexity of cross-band prediction in FDD and highlights the need for scenario-specific modeling and adaptive training strategies, such as distribution-shift detection and retraining \cite{jiang2024enhancing, yang2019deep}, to ensure reliable inference performance.

\subsection{NMSE-Based Evaluation: SNR, User Velocity, Delay Spread, \& Channel Model}
\label{sec:nmse-based-evaluation}

This section reports the NMSE comparisons between the proposed \Model{} and the baselines across various factors. Specifically, Sections~\ref{sec:performance-awgn-noises}--\ref{sec:performance-channel-models} analyze sensitivity to (i) AWGN SNR, (ii) user velocity, (iii) channel model, and (iv) delay spread, respectively. The objective is to provide a detailed examination of how each factor influences the CSI prediction performance across different models.

Unless stated otherwise, when one factor is examined, performance is \emph{aggregated over all remaining factors}. For example, in Section~\ref{sec:performance-awgn-noises}, for each SNR the NMSEs are averaged across all channel models, delay spreads, and user velocities; the same convention applies in the other sections.

\begin{figure*}[!htbp]
  \centering
  \subfloat[TDD]{
    \includegraphics[width=0.39\textwidth]{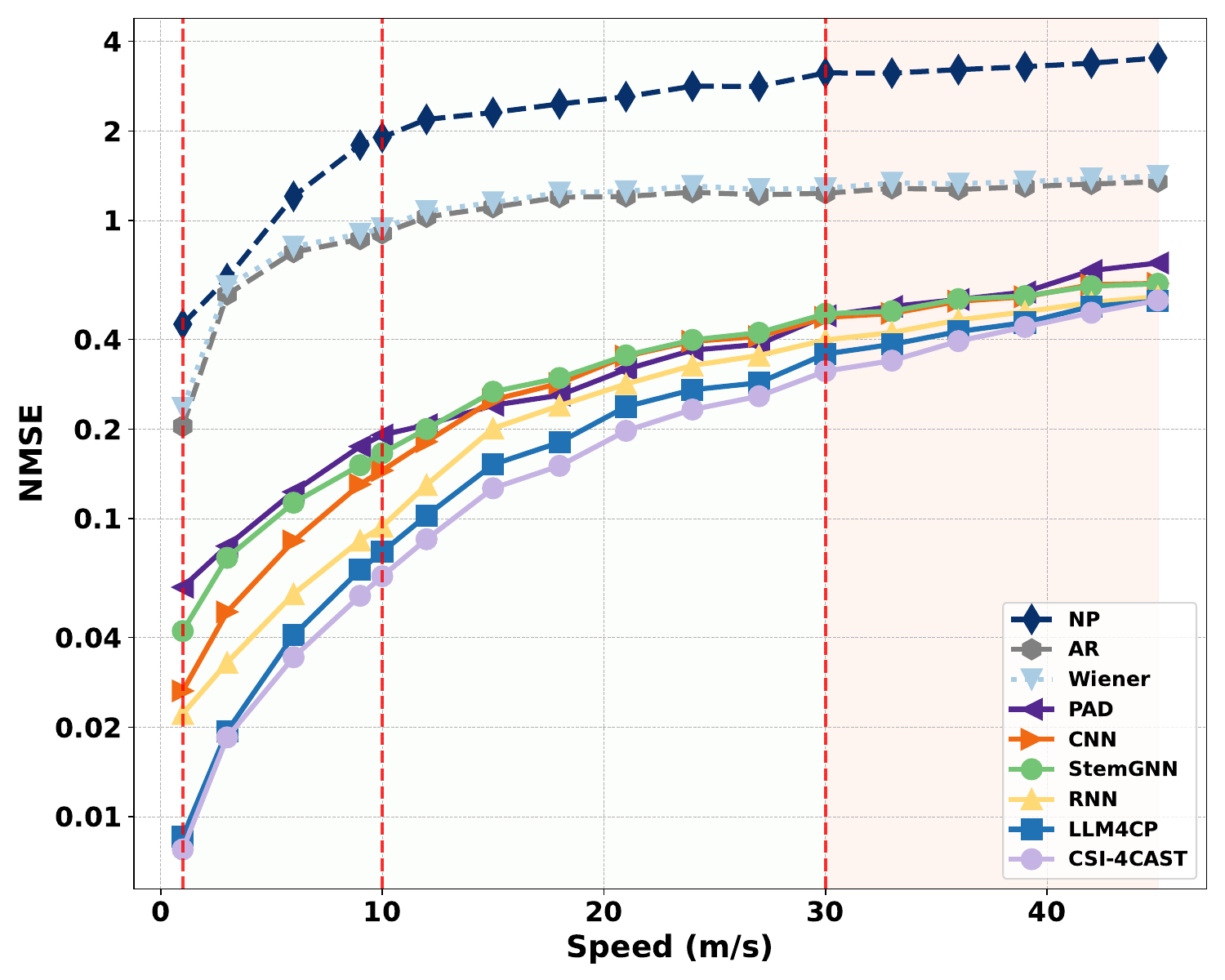}
    \label{fig:tdd_vanilla_nmse_ms_no_std_res_generalization}
  }\
  \subfloat[FDD]{
    \includegraphics[width=0.4\textwidth]{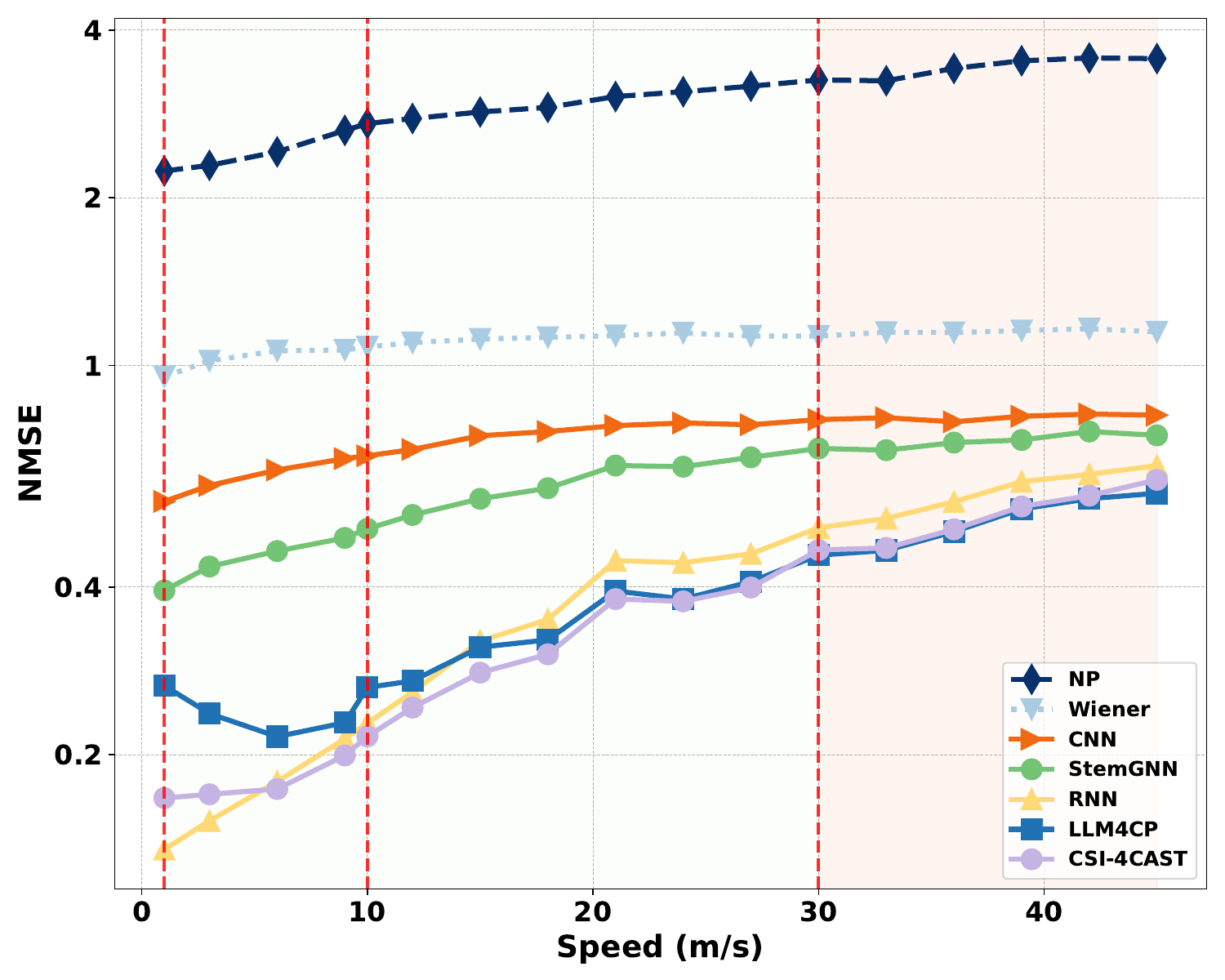}
    \label{fig:fdd_vanilla_nmse_ms_no_std_res_generalization}
  }
  \caption{\textbf{NMSE across user velocities.} Red dashed vertical lines mark the velocities included in the \regular{} set; all other velocities belong to the \generalization{} set. Light green shading denotes the interpolation region (velocities within the \regular{} range), whereas light red denotes the extrapolation region (velocities outside that range).}
  \label{fig:generalization-across-user-velocities-nmse}
\end{figure*}

\subsubsection{Performance Under Varying AWGN SNRs}
\label{sec:performance-awgn-noises}

Fig.~\ref{fig:performance-awgn-noises-nmse-regular} presents NMSE versus AWGN SNR injected into the \emph{input} historical CSI for TDD and FDD. This setup reflects practical conditions in which the observed CSI is corrupted by varying levels of AWGN, requiring prediction models to operate reliably across all such cases. Across all models and both duplexing modes, \Model{} consistently achieves the lowest NMSE at every SNR, demonstrating strong robustness to input AWGN. \response{While PAD achieves nearly the same NMSE as \Model{} at SNR$=25$\,dB, it also exhibits the largest performance degradation from SNR$=0$\,dB to SNR$=25$\,dB. This behavior highlights PAD's strong expressivity when channel noise is low and aligns with its model assumptions, but also reveals its vulnerability to input noise.}

Comparing TDD and FDD performance, NMSE is consistently higher under FDD. The larger NMSE of NP in FDD directly reflects that prediction under FDD is intrinsically more challenging than under TDD, consistent with the analysis in Section~\ref{sec:problem-definition}, where FDD was identified as an inter-band prediction task requiring prediction across both time and frequency. Similarly, all other deep learning models exhibit higher NMSE in FDD than in TDD. For example, at $\text{SNR}=0$\,dB, the deep learning models (\Model{}, LLM4CP, CNN) achieve roughly $2\times$ higher NMSE in FDD than in TDD.

\subsubsection{Performance Under Varying User Velocities}
\label{sec:performance-user-velocities}
  
Fig.~\ref{fig:generalization-across-user-velocities-nmse} reports NMSE versus user velocity for both TDD and FDD. The training dataset and the \Regular{} track include velocities of 1, 10, and 30 m/s, marked by the red dashed vertical lines. The \Generalization{} track extends to velocities from 3 to 45 m/s with finer granularity, covering both interpolation (velocities within the training range, shaded light green) and extrapolation (velocities outside the training range, shaded light red). Considering user velocity is crucial in the current mobile communication era, as UEs operate across a wide range of speeds, from walking to driving or transporting with various speeds.

Across all duplexing modes and user velocities, \Model{} achieves the lowest NMSE or matches the next best baseline. For both TDD and FDD, NMSE increases with user velocity, consistent with high-speed-induced \emph{temporal decorrelation}, which reduces the predictability of future CSI. This interpretation is supported by the autocorrelation analysis in Appendix~\ref{sec:appendix-acf-user-velocities}.

Comparing duplexing modes, FDD consistently produces higher NMSE than TDD. However, the performance degradation from lowest to highest user speed is more severe in TDD. For \Model{} and baselines, NMSE increases by more than $10\times$ between 1 m/s and 45 m/s, whereas in FDD the increase is less than $3\times$ over the same range. The results imply that TDD prediction leverages temporal channel correlation more directly, and as coherence time shortens rapidly with high velocity, its performance becomes significantly degraded. In contrast, the FDD task is inherently more challenging, its prediction accuracy is already limited at low user speeds, so the relative degradation with increasing velocity appears less pronounced.

\begin{table}[!t]
  \caption{\textbf{\response{NMSE under varying delay spreads.}} \textbf{Bold} \textnormal{denotes the best value and \underline{underline} the second-best (lower NMSE is better).}}
  \label{tab:performance-delay-spreads}
  \centering
  \setlength{\tabcolsep}{3pt}        
  \renewcommand{\arraystretch}{1.1}  
  \begin{tabularx}{\linewidth}{@{} l Y Y Y | Y Y Y @{}}
    \toprule
    \textbf{Models} &
      \multicolumn{3}{c}{\textbf{Regular}} &
      \multicolumn{3}{c}{\textbf{Generalization}} \\
    & \textbf{30 ns} & \textbf{100 ns} & \textbf{300 ns} &
      \textbf{50 ns} & \textbf{200 ns} & \textbf{400 ns} \\
    \midrule
    \multicolumn{7}{@{}l}{\textbf{TDD}}\\
    \cmidrule(lr){1-7}
     NP       & 2.246 & 1.714 & 1.526 & 1.891 & 1.499 & 1.466 \\
     AR       & 0.843 & 0.767 & 0.738 & 0.799 & 0.733 & 0.724 \\
     WIENER   & 0.897 & 0.801 & 0.764 & 0.848 & 0.762 & 0.747 \\
     PAD      & 0.359 & 0.216 & 0.156 & 0.272 & 0.154 & \underline{0.151} \\
     CNN      & 0.261 & 0.211 & 0.174 & 0.230 & 0.173 & 0.172 \\
     STEMGNN  & 0.253 & 0.230 & 0.212 & 0.222 & 0.207 & 0.214 \\
     RNN      & \underline{0.191} & 0.168 & 0.156 & 0.176 & 0.164 & 0.282 \\
     LLM4CP   & 0.195 & \underline{0.145} & \underline{0.103} & \underline{0.163} & \textbf{0.110} & \textbf{0.141} \\
     \Model{}    & \textbf{0.176} & \textbf{0.125} & \textbf{0.084} & \textbf{0.148} & \underline{0.119} & 0.174 \\
     \midrule
     \multicolumn{7}{@{}l}{\textbf{FDD}}\\
     \cmidrule(lr){1-7}
     NP       & 2.798 & 2.920 & 2.488 & 3.045 & 2.312 & 2.025 \\
     WIENER   & 1.104 & 1.022 & 1.036 & 1.090 & 1.061 & 1.014 \\
     CNN      & 0.631 & 0.800 & 0.626 & 0.744 & \textbf{0.797} & \underline{0.994} \\
     STEMGNN  & 0.493 & 0.597 & 0.523 & 0.562 & \underline{0.818} & \textbf{0.894} \\
     RNN      & \textbf{0.301} & \textbf{0.318} & 0.255 & \textbf{0.425} & 0.930 & 1.330 \\
     LLM4CP   & 0.334 & 0.467 & \textbf{0.184} & 0.557 & 1.003 & 1.421 \\
     \Model{}    & \underline{0.316} & \underline{0.342} & \underline{0.190} & \underline{0.495} & 0.976 & 1.417 \\
     \bottomrule
    \end{tabularx}
\end{table}

\subsubsection{Performance Under Varying Delay Spreads}
\label{sec:performance-delay-spreads}

Table~\ref{tab:performance-delay-spreads} reports NMSE across various delay spreads: 30, 100, and 300 ns, which are included in training; 50 and 200 ns, used for interpolation within the training range; and 400 ns, used for extrapolation beyond the training range. The delay spread reflects the multipath richness of the channel, which varies significantly across environments (e.g., urban, indoor, rural), motivating the need for evaluation under diverse delay spreads.

For TDD systems, \Model{} achieves the lowest NMSE on the \Regular{} track and remains the best or a close second-best on the \Generalization{} track. The deep learning models generally show degraded performance on unseen delay spreads. In particular, at 400 ns (outside the training range), \Model{}, LLM4CP, and RNN experience substantial performance drops compared with 300 ns, the nearest seen condition. For example, NMSE increases by about 50\% for LLM4CP and nearly 100\% for RNN and \Model{}. These results highlight the challenge of generalizing to unseen delay spreads. \response{Interestingly, although PAD is not a competitive model in TDD overall, it becomes the second-best method at 400 ns. A plausible explanation is that PAD performs parameter estimation during inference, enabling it to adapt to the instantaneous multipath structure under unseen delay spreads. This improved generalization performance, however, comes at the cost of significantly higher inference complexity and latency (Appendix~\ref{sec:appendix-computational-overhead}).}

For FDD systems, \Model{} achieves the second-best NMSE on the \Regular{} track for the seen delay spreads. On the \Generalization{} track, however, performance degrades substantially for the unseen delay spreads. At 400 ns, the degradation is particularly severe: NMSE increases by more than $5\times$ for \Model{}, LLM4CP, and RNN. This sharper degradation in FDD arises from the added challenge of cross-frequency prediction. Since delay spread is inversely proportional to coherence bandwidth, a larger delay spread reduces frequency-domain correlation, making inter-band prediction more difficult in the FDD setting.

The overall results across different delay spreads underscore that generalizing to unseen delay spreads remains challenging for deep learning models. One reason is the complex and intricate influence of delay spread on channel properties. Unlike user velocity, which primarily smoothly affects temporal ACF, delay spread interacts with the channel model to jointly shape the temporal-frequency ACF, as supported by the analysis in Appendix~\ref{sec:appendix-acf-delay-spreads}. The dynamics of multipath structure and scattering geometry are difficult to model and predict, leading to poor generalization. This observation highlights the importance of developing more scenario-specific models, particularly for FDD.

\begin{table}[!t]
  \caption{\textbf{\response{NMSE under varying channel models.}} \textbf{Bold} \textnormal{denotes the best value and \underline{underline} the second-best (lower NMSE is better).}}
  \label{tab:performance-channel-models-regular}
  \centering
  \setlength{\tabcolsep}{3pt}        
  \renewcommand{\arraystretch}{1.1}  
  \begin{tabularx}{\linewidth}{@{} l Y Y Y | Y Y @{}}
    \toprule
    \textbf{Models} &
      \multicolumn{3}{c}{\textbf{Regular}} &
      \multicolumn{2}{c}{\textbf{Generalization}} \\
    & \textbf{CDL-A} & \textbf{CDL-C} & \textbf{CDL-D} &
      \textbf{CDL-B} & \textbf{CDL-E} \\
    \midrule
    \multicolumn{6}{@{}l}{\textbf{TDD}}\\
    \cmidrule(lr){1-6}
     NP       & 2.167 & 1.948 & 1.372 & 1.902 & 1.353 \\
     AR       & 0.825 & 0.804 & 0.718 & 0.822 & 0.731 \\
     WIENER   & 0.879 & 0.849 & 0.734 & 0.880 & 0.744 \\
     PAD      & 0.327 & 0.338 & 0.065 & 0.449 & 0.066 \\
     CNN      & 0.246 & 0.335 & 0.065 & 0.405 & 0.074 \\
     STEMGNN  & 0.283 & 0.364 & 0.049 & 0.453 & 0.058 \\
     RNN      & 0.207 & 0.263 & 0.044 & 0.394 & 0.057 \\
     LLM4CP   & \underline{0.168} & \underline{0.245} & \underline{0.031} & \textbf{0.349} & \underline{0.043} \\
     \Model{}    & \textbf{0.156} & \textbf{0.207} & \textbf{0.022} & \underline{0.376} & \textbf{0.036} \\
     \midrule
     \multicolumn{6}{@{}l}{\textbf{FDD}}\\
     \cmidrule(lr){1-6}
     NP       & 3.857 & 2.935 & 1.415 & 3.029 & 1.419 \\
     WIENER   & 1.188 & 1.154 & 0.822 & 1.212 & 0.818 \\
     CNN      & 0.903 & 0.925 & 0.230 & \textbf{1.091} & 0.278 \\
     STEMGNN  & 0.677 & 0.855 & 0.081 & \underline{1.107} & 0.106 \\
     RNN      & \textbf{0.367} & 0.443 & 0.063 & 1.190 & \textbf{0.091} \\
     LLM4CP   & 0.498 & \underline{0.431} & \underline{0.059} & 1.268 & 0.106 \\
     \Model{}    & \underline{0.385} & \textbf{0.410} & \textbf{0.052} & 1.308 & \underline{0.092} \\
     \bottomrule
    \end{tabularx}
\end{table}

\subsubsection{Performance Under Varying Channel Models}
\label{sec:performance-channel-models}

\begin{figure*}[!ht]
  \centering
  \subfloat[TDD: Phase noise]{
    \includegraphics[width=0.32\textwidth]{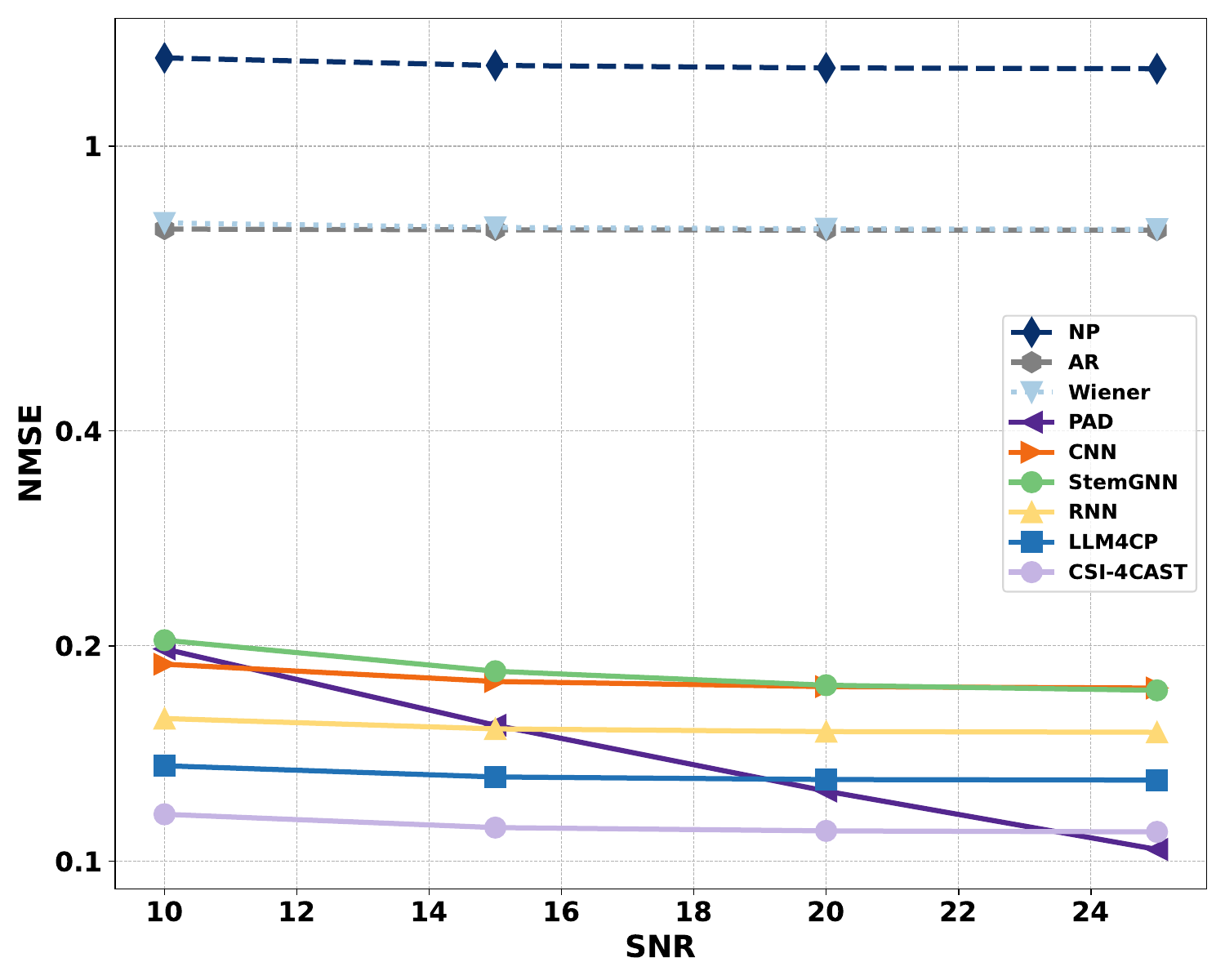}
    \label{fig:tdd_phase_nmse_noise_degree}
  }
  \subfloat[TDD: Burst noise]{
    \includegraphics[width=0.32\textwidth]{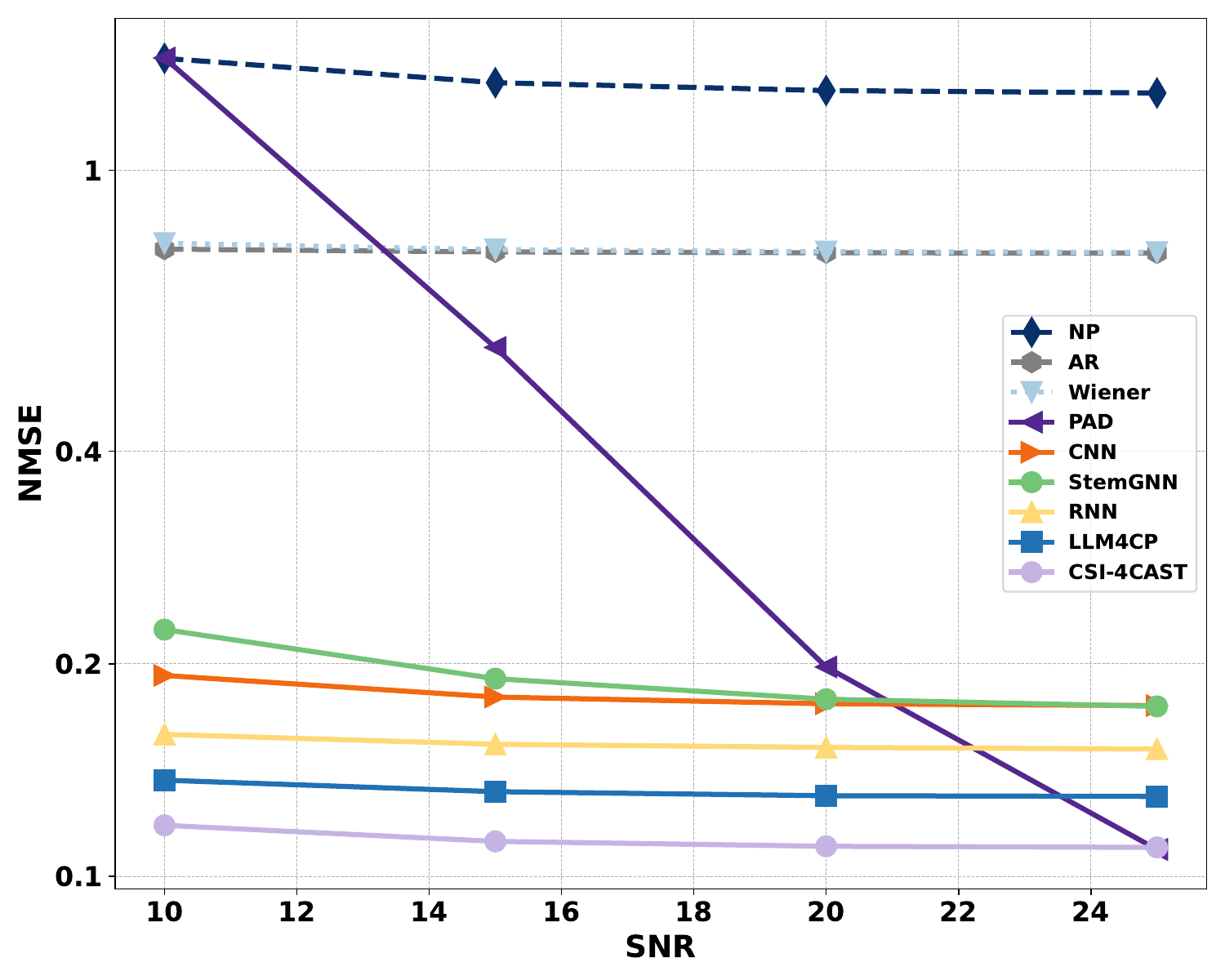}
    \label{fig:tdd_burst_nmse_noise_degree}
  }
  \subfloat[TDD: Packet drop noise]{
    \includegraphics[width=0.32\textwidth]{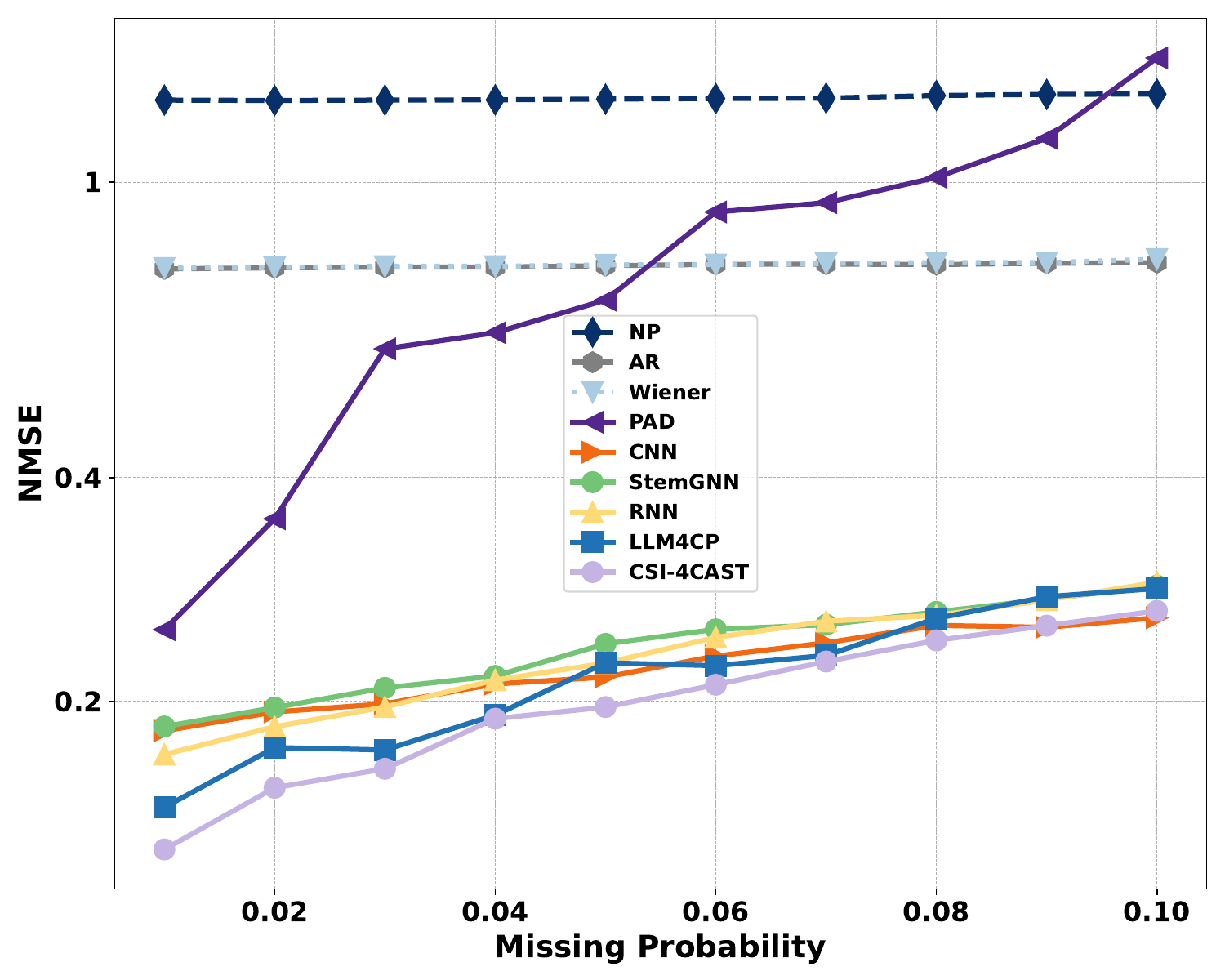}
    \label{fig:tdd_packagedrop_nmse_noise_degree}
  }\
  \subfloat[FDD: Phase noise]{
    \includegraphics[width=0.32\textwidth]{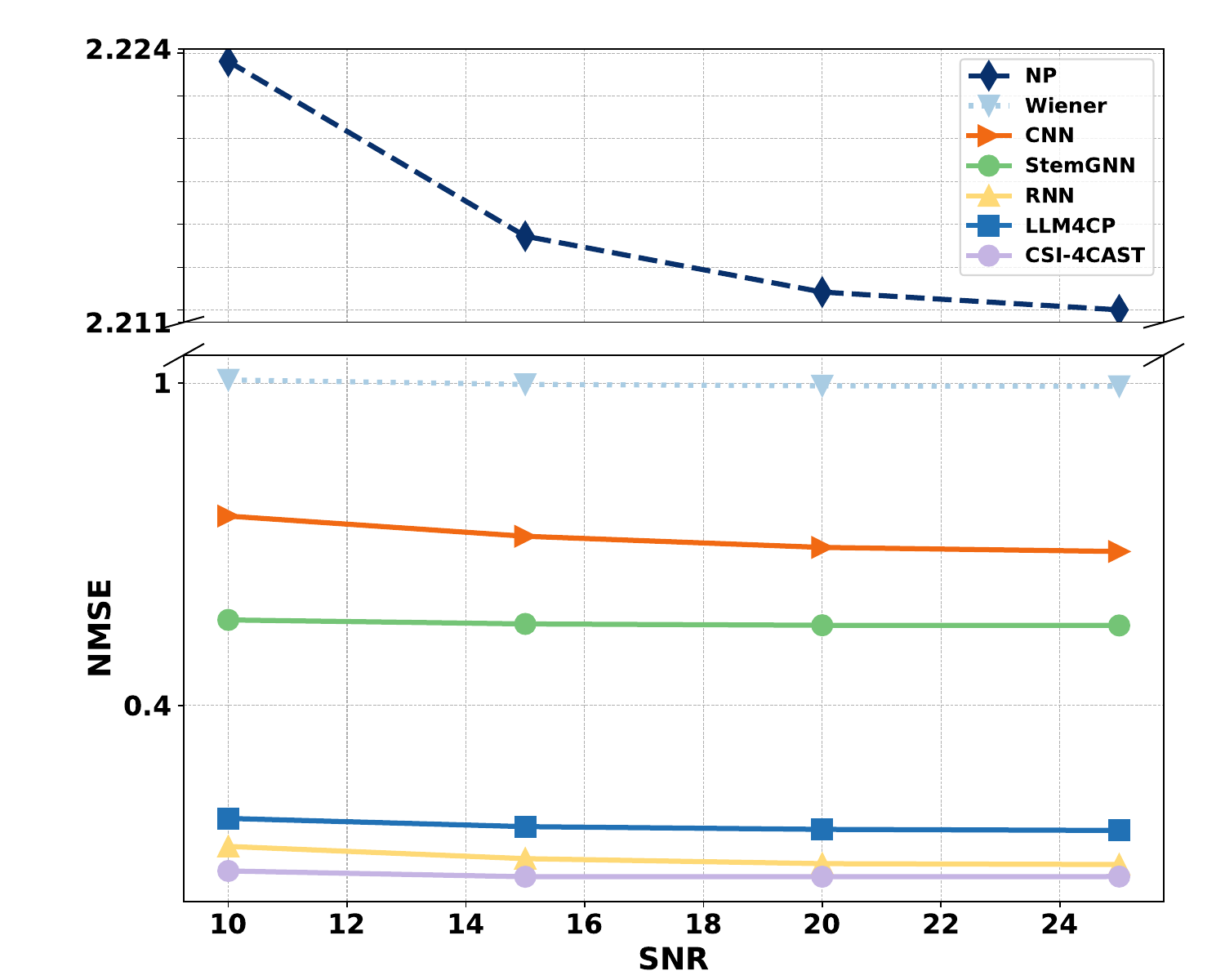}
    \label{fig:fdd_phase_nmse_noise_degree}
  }
  \subfloat[FDD: Burst noise]{
    \includegraphics[width=0.32\textwidth]{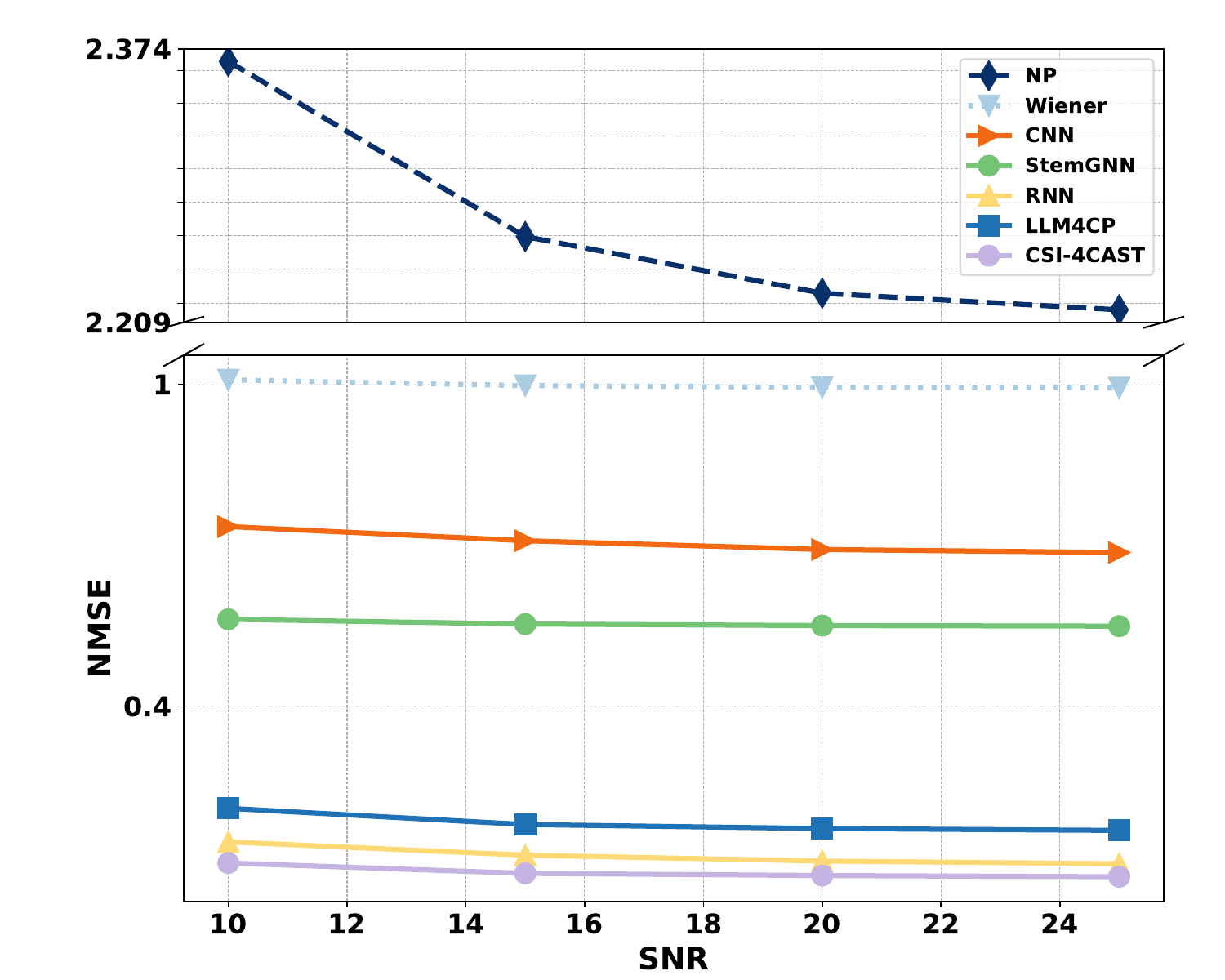}
    \label{fig:fdd_burst_nmse_noise_degree}
  }
  \subfloat[FDD: Packet drop noise]{
    \includegraphics[width=0.32\textwidth]{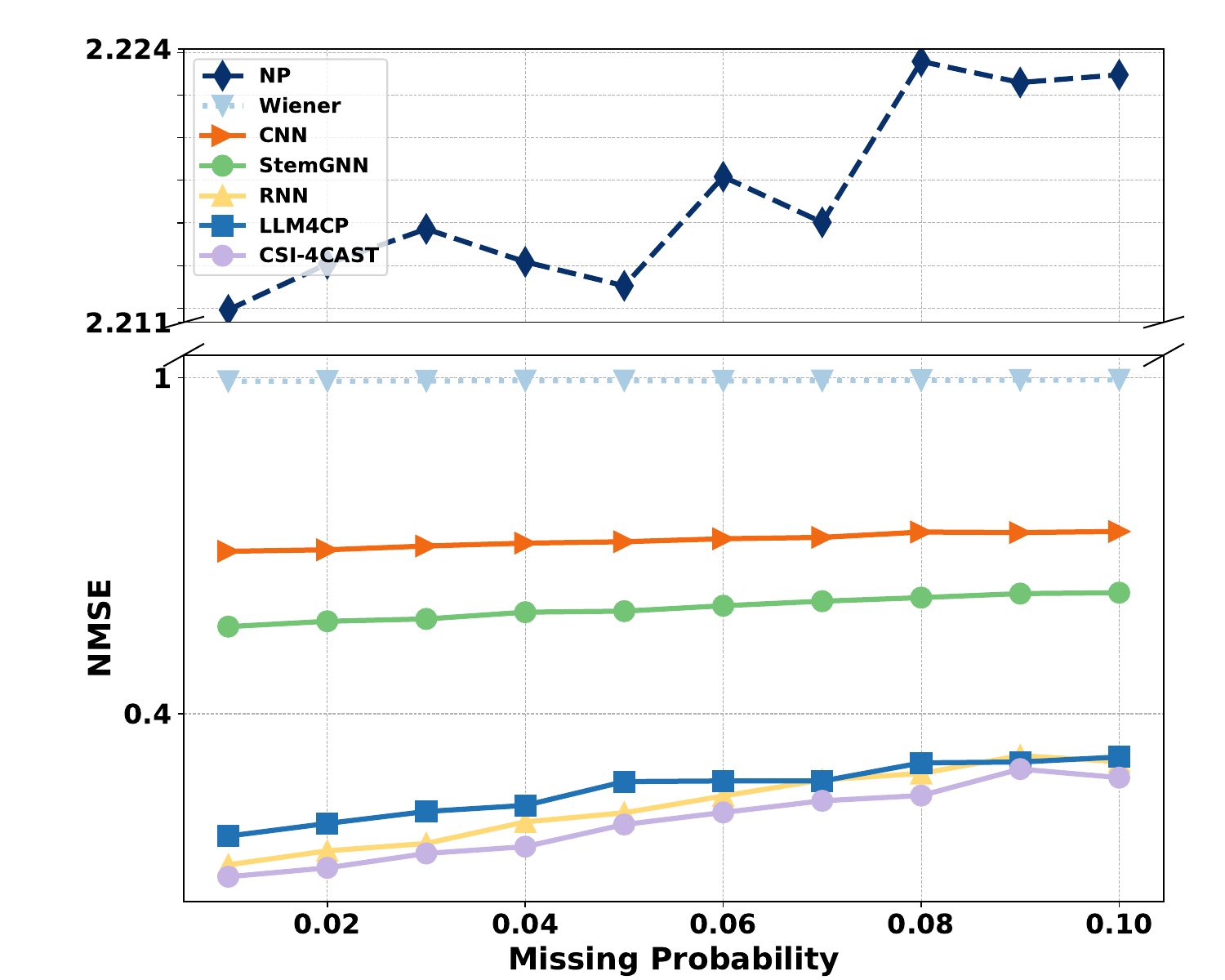}
    \label{fig:fdd_packagedrop_nmse_noise_degree}
  }\
  \caption{\textbf{NMSE under varying realistic additional noises: TDD and FDD.}}
  \label{fig:performance-realistic-additional-noises-nmse}
\end{figure*}

Table~\ref{tab:performance-channel-models-regular} presents NMSE results across different channel models for both TDD and FDD. CDL-A/C/D are included in training (\Regular{} track), while CDL-B and CDL-E are used for generalization (\Generalization{} track). CDL-A/B/C typically represent \emph{NLOS} conditions with diffuse multipath and rich angular dispersion, often corresponding to dense urban streets or cluttered indoor offices. In contrast, CDL-D and CDL-E represent \emph{LOS} conditions with a strong dominant direct path, commonly found in open outdoor spaces or rural macro scenarios \cite{3gpp-38901}. Including diverse channel models in evaluation is essential to capture real-world environmental dynamics, as mobile communication scenarios span a wide range of conditions that can be abstracted by different channel models.

\Model{} is the overall best performer, ranking within the top two across all channel models for both TDD and FDD, with the only exception being FDD CDL-B. This highlights the strong prediction performance and generalization ability of \Model{}. A detailed comparison shows that CDL-A/B/C (NLOS) are more challenging than CDL-D/E (LOS) across all models and duplexing modes, with NLOS yielding NMSE values an order of magnitude larger than LOS in both the \Regular{} and \Generalization{} tracks. This observation aligns with the above analysis of the channel characteristics: complex scattering and rapid fluctuations in NLOS conditions increase prediction difficulty.

In particular, FDD with CDL-B poses an especially difficult generalization task, as the combination of FDD and NLOS amplifies prediction challenges. Strong models such as RNN, LLM4CP, and \Model{}, which perform best under other conditions, show severe degradation. Compared with CDL-A/C, NMSE in CDL-B increases by more than 100\% for all three models.

\subsection{Robustness Analysis Across Various Noise Types and Degrees}
\label{sec:robustness-performance}

This section evaluates the robustness of \Model{} and the baselines under three realistic noise types: \emph{phase noise}, \emph{burst-type corruption}, and \emph{packet drops}, which are previously introduced in Section~\ref{sec:introduction}. Complete definitions, visualizations, parameter ranges, generation procedures, and experiment details for all additional noises are provided in Appendix \ref{sec:additive-noise}.

The NMSE of the models across different noise types are presented in Fig.~\ref{fig:performance-realistic-additional-noises-nmse} for both TDD and FDD. Across all noise types and levels, \Model{} consistently achieves the lowest, \response{or nearly the lowest}, NMSE, demonstrating the strongest robustness. Interestingly, CNN performs well under packet drop noise, particularly at high drop probabilities, while \Model{}, which has CNN at the front end, shows the best overall robustness. These results support the intuition that CNN-style residual representations are effective for handling channel corruption and extracting structural features \cite{chen2024complex, zhang2017beyond}. Besides, the performance degradation due to packet drop noise is also more severe in TDD than in FDD, consistent with the expectation that temporal correlation is more critical in TDD.

\response{Additionally, PAD’s performance under TDD deserves attention. It shows the largest performance variation across noise intensities. This behavior becomes even more pronounced under burst noise and packet-drop noise, where the noise structure deviates from PAD’s underlying assumptions. Further discussion of the underlying mechanisms is provided in Appendix~\ref{sec:appendix-pad-safe-fallback}.}

\begin{figure}[!t]
  \centering
  \subfloat[TDD]{
    \includegraphics[width=0.47\linewidth]{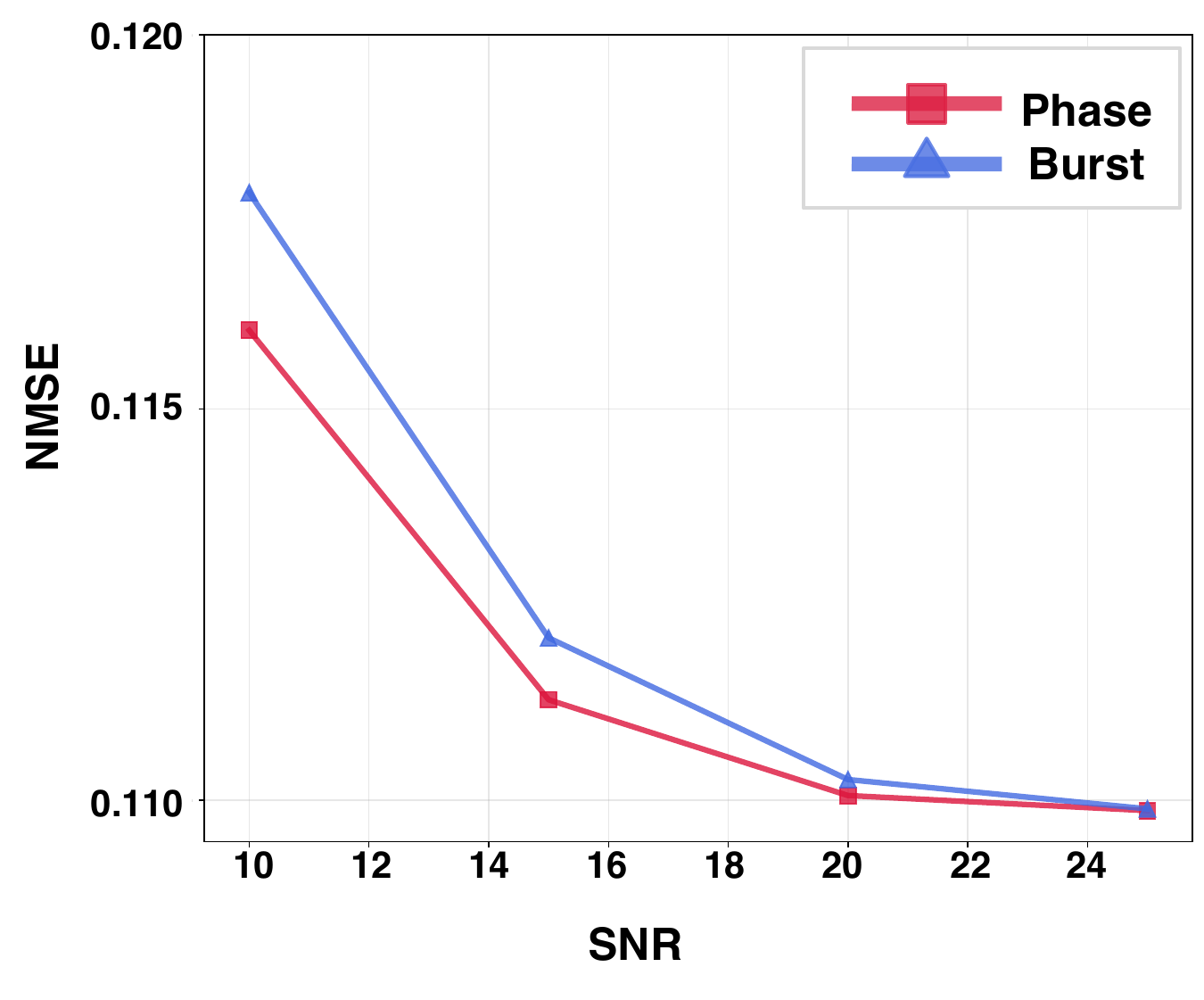}
    \label{fig:tdd_nmse_model_different_noise}
  }
  \subfloat[FDD]{
    \includegraphics[width=0.47\linewidth]{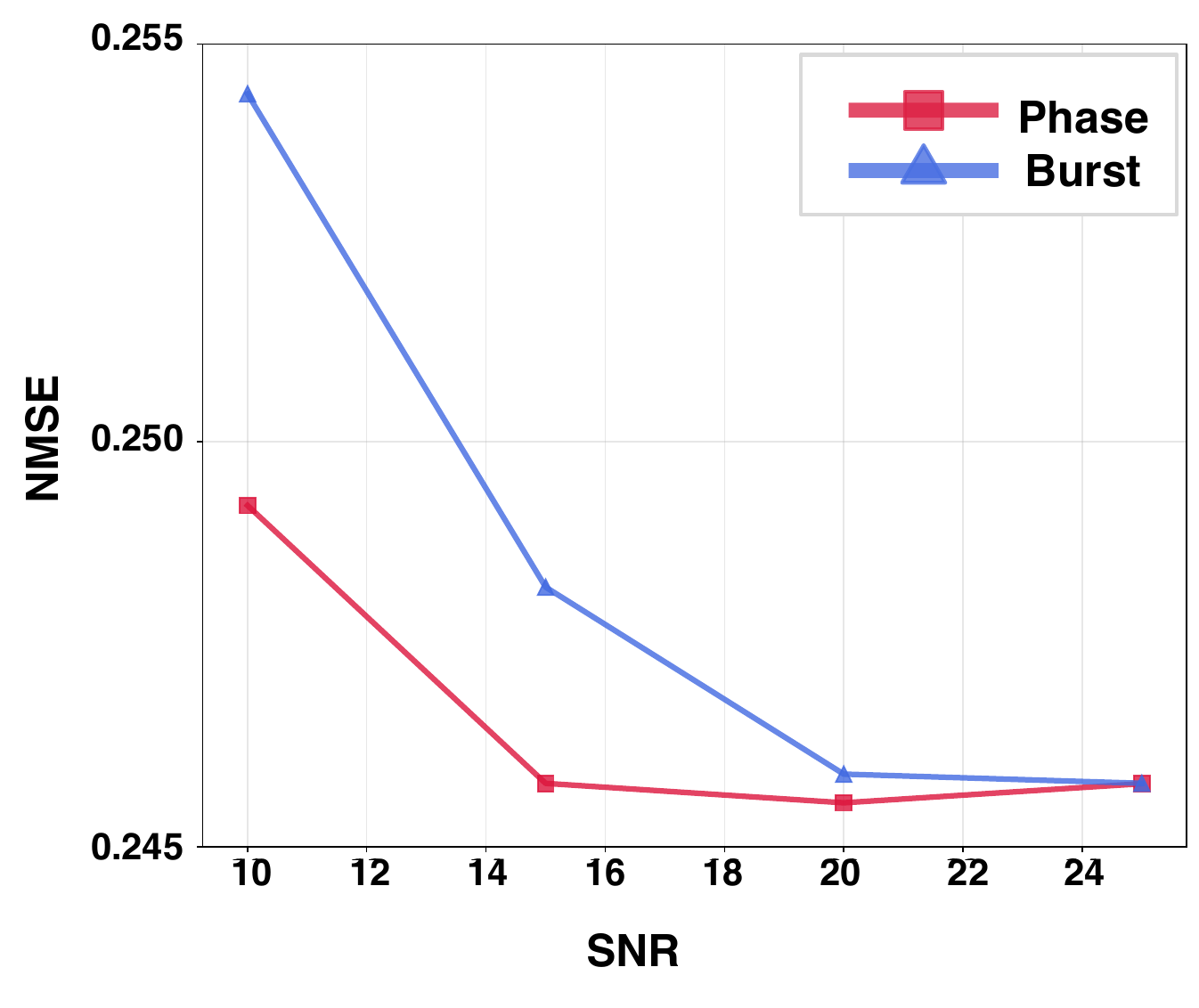}
    \label{fig:fdd_nmse_model_different_noise}
  }
  \caption{\textbf{\Model{}: NMSE under Phase and Burst Noise.}}
  \label{fig:performance-realistic-additional-noises-nmse-model-phase-burst}
\end{figure}

Moreover, Fig.~\ref{fig:performance-realistic-additional-noises-nmse-model-phase-burst} compares \Model{}'s performance under phase noise and burst noise at matched SNRs. Despite equal SNR, the two noise types have different effects: burst noise causes greater NMSE degradation than phase noise. Phase noise introduces smooth, continuous perturbations that partly resemble the AWGN used in training, keeping \Model{} relatively stable. By contrast, burst noise is abrupt and high-energy over short windows, disrupting the temporal coherence that \Model{} relies on and leading to sharper performance losses. \response{These results indicate that robustness depends not only on the SNR level but also on the noise structure, affecting both model-based baselines and learning-based approaches.}

\subsection{Overall Performance: Prediction Performance and Computational Considerations}
\label{sec:overall-performance}

\begin{figure*}[!ht]
  \centering
  \subfloat[TDD]{
    \includegraphics[width=0.48\linewidth]{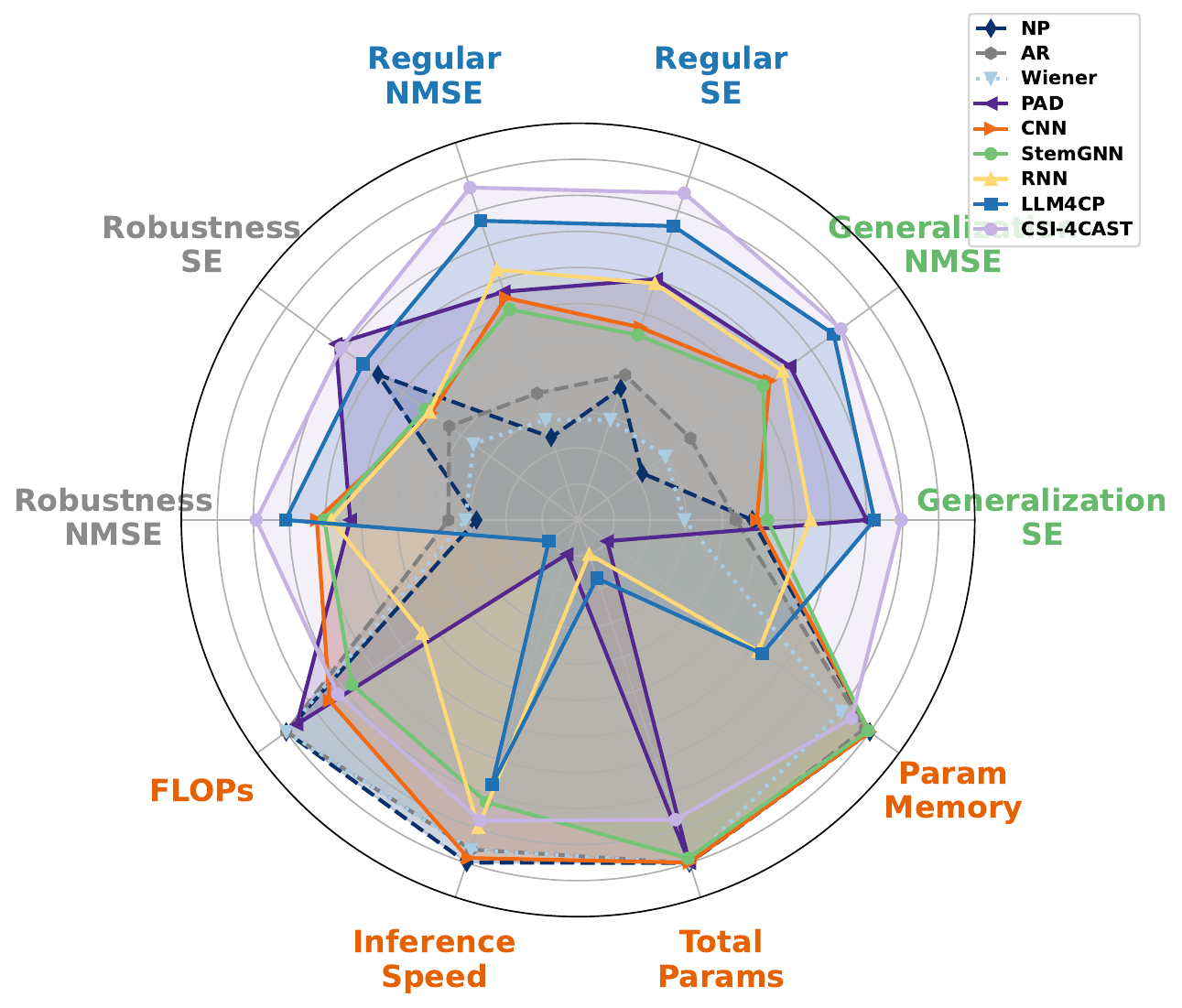}
    \label{fig:radar_tdd_no_median_no_trainable_no_training_time_w_offset}
  }
  \subfloat[FDD]{
    \includegraphics[width=0.48\linewidth]{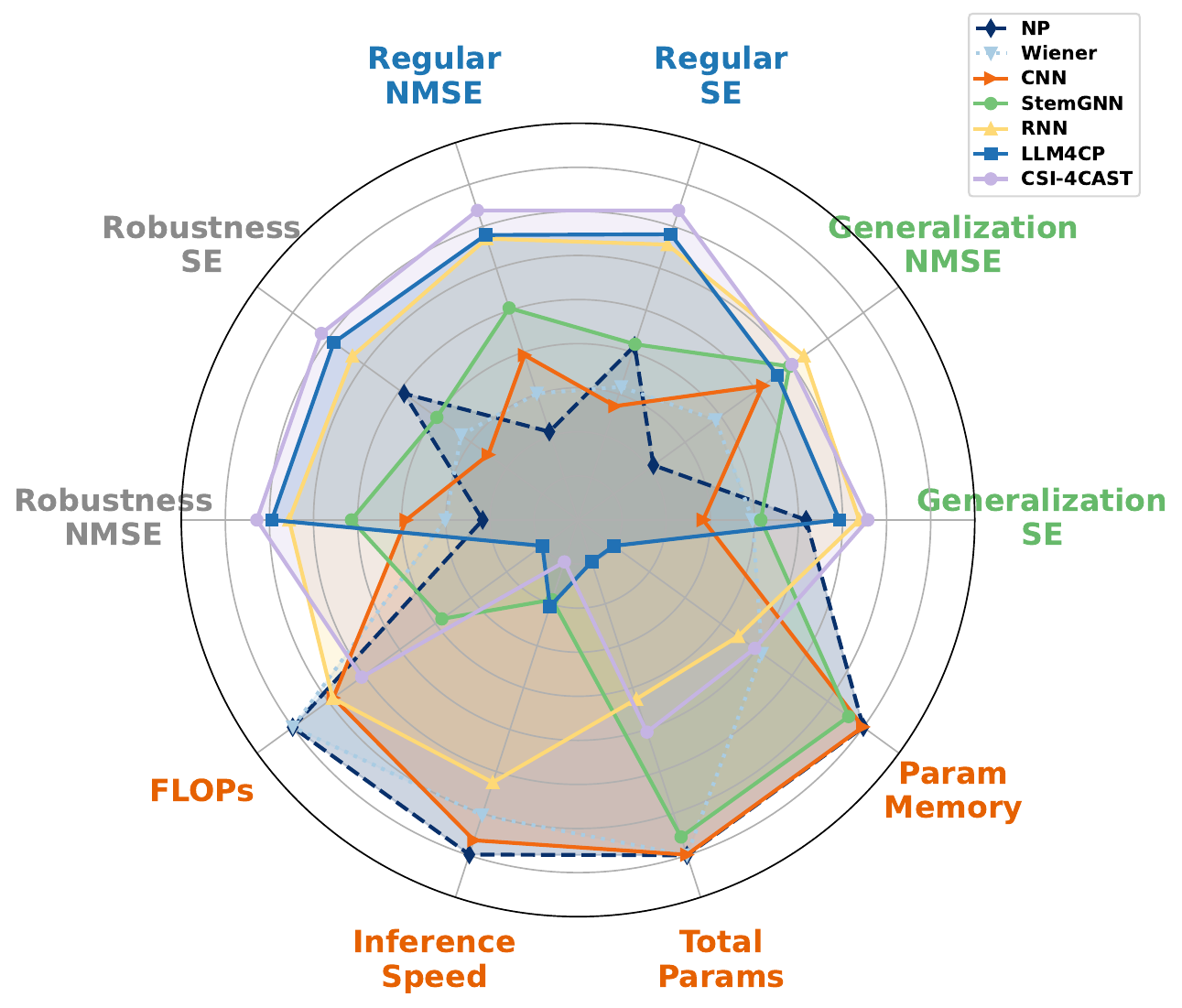}
    \label{fig:radar_fdd_no_median_no_trainable_no_training_time_w_offset}
  }
  \caption{\textbf{Overall performance.} The \reg{blue}, \rob{gray}, and \gen{green} axes represent \emph{prediction performance} for \reg{\Regular{}}, \rob{\Robustness{}}, and \gen{\Generalization{}} tracks, measured using the $\mathrm{RankScore}$ \eqref{eq:mean-rank-score}, with separate axes for results based on NMSE-rank and Spectral Efficiency (SE)-rank. The orange axes indicate \emph{computational cost}, quantified by the $\mathrm{EffScore}$ \eqref{eq:efficiency-score}. Each colored polygon corresponds to a model. By construction, larger scores correspond to better results; hence, values farther from the center indicate stronger performance along that axis. A larger polygon therefore reflects a more favorable overall accuracy-efficiency trade-off. Since axes are scaled independently, comparisons should be made \emph{only along the same axis}.}
  \label{fig:performance-radar-overall-no_median_no_trainable_no_training_time_w_offset}
\end{figure*}

Fig.~\ref{fig:performance-radar-overall-no_median_no_trainable_no_training_time_w_offset} illustrates the overall performance of \Model{} compared to baseline models, jointly evaluating prediction accuracy on \dataset{} and computational overhead. The prediction performance axes are presented based on the $\mathrm{RankScore}$ \eqref{eq:mean-rank-score}. Unlike earlier sections that focused solely on NMSE, both NMSE-rank and Spectral Efficiency (SE)-rank are included here. This choice is motivated by two factors: (i) high SE is a central goal in wireless communication systems, and (ii) NMSE and SE reflect distinct aspects of model performance, where a low NMSE does not necessarily correspond to high SE. Additional details and a breakdown of SE performance are provided in Appendix~\ref{sec:appendix-se}. NMSE-rank and SE-rank are reported across the \Regular{}, \Robustness{}, and \Generalization{} tracks. The orange axes reflect computational efficiency, assessed via the $\mathrm{EffScore}$ \eqref{eq:efficiency-score}, which accounts for FLOPs, total parameters, inference time, \response{and memory footprint}. Notably, $\mathrm{RankScore}$ and $\mathrm{EffScore}$ are defined such that higher values indicate better performance. Thus, in the figure, larger values along individual axes represent stronger performance in that aspect, and a larger overall polygon reflects a more favorable trade-off between accuracy and efficiency.

For TDD, \Model{} achieves the strongest overall performance, dominating all three evaluation tracks and both performance metrics while maintaining significantly lower computational complexity. Only LLM4CP provides competitive performance on \Generalization{} w.r.t. NMSE-rank, whereas all other baselines fall consistently behind across every track and both metrics. In terms of efficiency, \Model{} is particularly advantageous: its FLOPs are comparable to CNN and only about $1/5$ of LLM4CP; its parameter count is reduced to roughly $1/7$ of LLM4CP; and its inference time is similar to RNN, requiring only about $1/2$ as long as LLM4CP. \response{Comparisons with model-based baselines further highlight the strengths of \Model{}. Despite their conceptual simplicity, AR, Wiener, and PAD exhibit substantial performance gaps relative to \Model{}. Moreover, these methods introduce non-negligible computational overhead. For instance, although Wiener has relatively low FLOPs, it requires a large memory footprint (154\% of \Model{}). PAD is even more computationally demanding, with over 7$\times$ higher inference latency and more than 14$\times$ the memory footprint of \Model{}. Considering both performance and efficiency, the strong overall performance of \Model{} justifies its computational cost. Detailed efficiency statistics are provided in Appendix~\ref{sec:appendix-computational-overhead}.}

For FDD, \Model{} continues to lead on the \Regular{} and \Robustness{} tracks w.r.t. both metrics, although with smaller margins. LLM4CP performs closely, and RNN also shows competitive performance—matching LLM4CP on \Regular{} and trailing slightly on \Robustness{}. On the \Generalization{} track, all models exhibit notable performance degradation due to the inherent difficulty of inter-band prediction. For NMSE-rank, the RankScores are tightly clustered, with CNN as the outlier. For SE-rank, \Model{}, LLM4CP, and RNN form a leading group with relatively close performance, clearly separated from the remaining models. These results underscore that generalization in FDD remains a significant challenge, with no model demonstrating clear superiority. \response{From a deployment perspective, this should be regarded as an important limitation: a fixed offline-trained FDD predictor is unlikely to remain reliable under strong scenario shifts without post-deployment adaptation.} Furthermore, the noticeable differences between SE-rank and NMSE-rank—particularly on the \Generalization{} track—highlight the distinct characteristics captured by each metric. From an efficiency perspective, \Model{} requires about $1/2$ the FLOPs of STEMGNN and $1/3$ of LLM4CP, with parameter count comparable to RNN and nearly $1/2$ of LLM4CP. However, due to the additional subcarrier-wise ACL layer and heavier hyperparameterization in FDD, \Model{}'s inference time aligns more closely with LLM4CP and STEMGNN and is slightly slower, in contrast to the clear speed advantage observed in TDD.

\subsection{Ablation Study}
\label{sec:ablation-study}

\begin{figure*}[!ht]
  \centering
  \subfloat[TDD]{
    \includegraphics[width=0.45\linewidth]{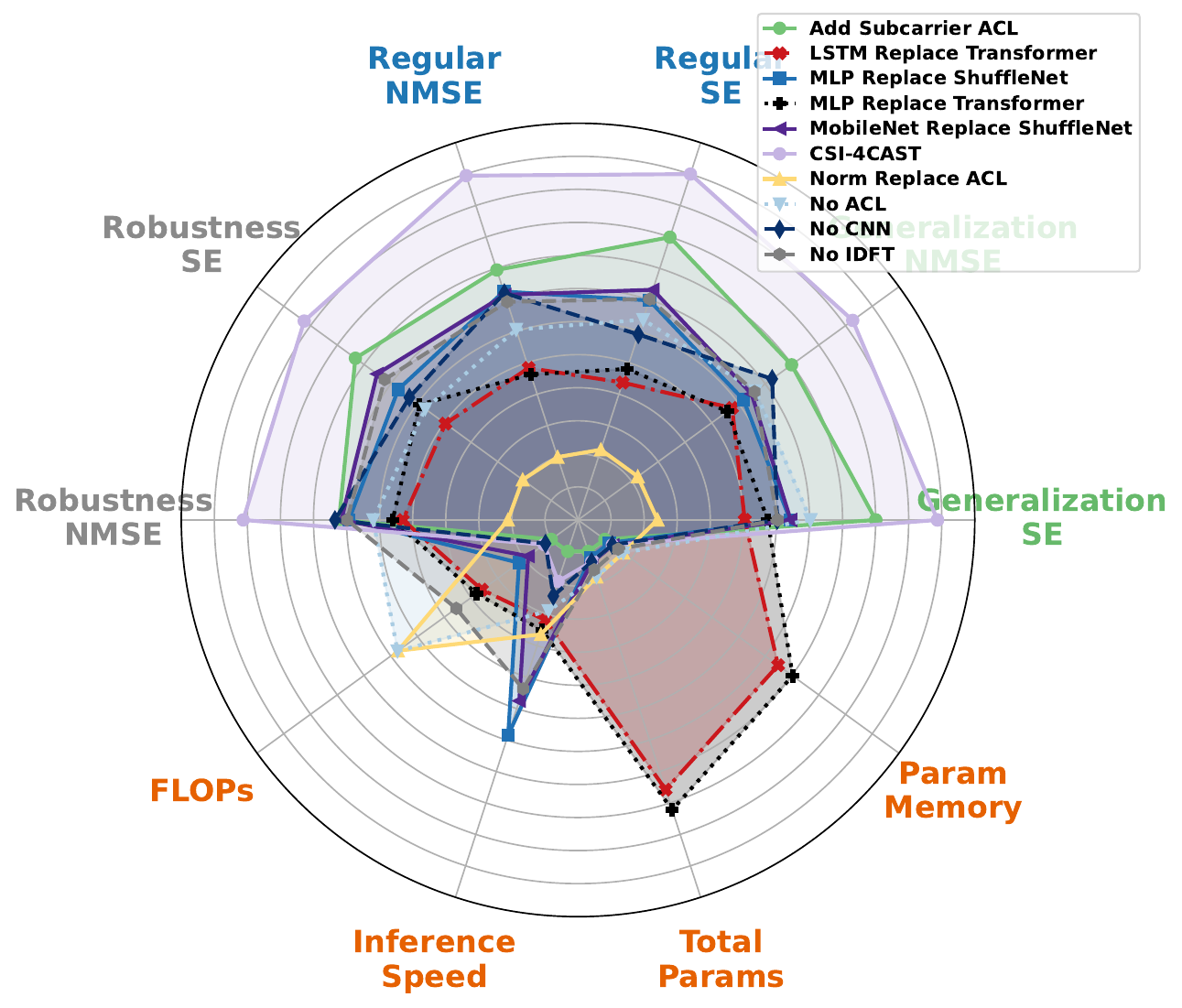}
    \label{fig:ablation_combined_radar_tdd}
  }
  \subfloat[FDD]{
    \includegraphics[width=0.45\linewidth]{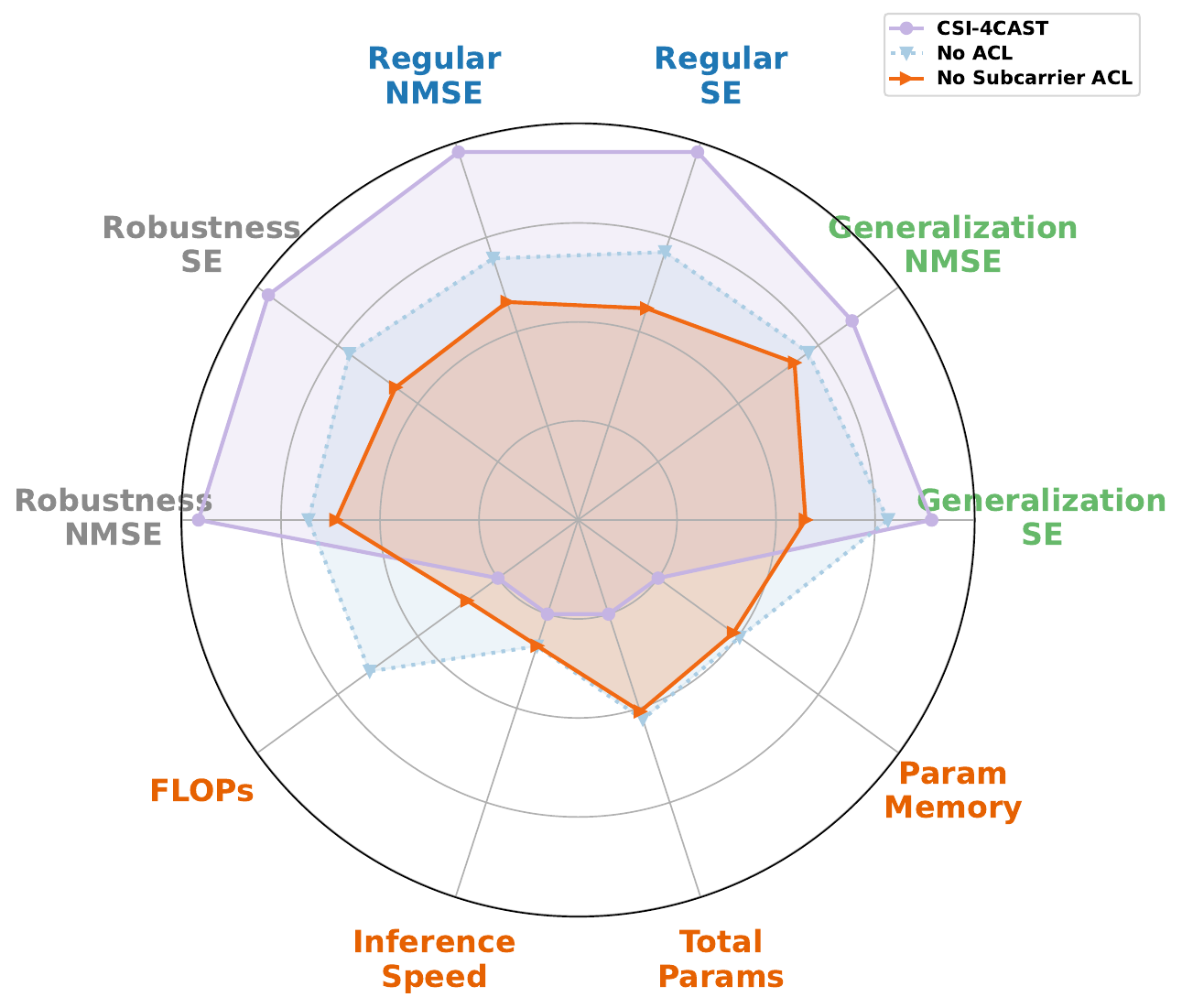}
    \label{fig:ablation_combined_radar_fdd}
  }
  \caption{\textbf{\response{Overall comparison of all ablation variants.}} \textnormal{The \reg{blue}, \rob{gray}, and \gen{green} axes represent \emph{prediction performance} for \reg{\Regular{}}, \rob{\Robustness{}}, and \gen{\Generalization{}} tracks, measured using the $\mathrm{RankScore}$ \eqref{eq:mean-rank-score}, with separate axes for results based on NMSE-rank and Spectral Efficiency (SE)-rank. The orange axes indicate \emph{computational cost}, quantified by the $\mathrm{EffScore}$ \eqref{eq:efficiency-score}. Each colored polygon corresponds to a model. By construction, larger scores correspond to better results; hence, values farther from the center indicate stronger performance along that axis. A larger polygon therefore reflects a more favorable overall accuracy-efficiency trade-off. Since axes are scaled independently, comparisons should be made \emph{only along the same axis}.}}
  \label{fig:ablation_combined_radar}
\end{figure*}

\response{To validate the effectiveness of the design choices in \Model{}, this section presents the results of comprehensive ablation experiments in which specific modules of \Model{} are removed or replaced. For modules shared by both the TDD and FDD settings, namely CNN, IDFT, ShuffleNet, and Transformer, ablation studies are conducted only under the TDD setting. For ACL, which has different structures in TDD and FDD, separate ablation experiments are performed for each duplexing mode. Each ablation variant is defined by changing exactly one module of the full model while leaving all other modules unchanged. The main findings are summarized in Table~\ref{tab:ablation-study} and Fig.~\ref{fig:ablation_combined_radar}. Further details, including precise definitions of the ablation models, the experimental setup, and discussions of the modules' underlying mechanisms, are provided in Appendix~\ref{sec:appendix-ablation}.}

\begin{table*}[!t]
  \caption{\textbf{\response{Ablation study on \Model{}.}} \textnormal{For the efficiency metrics and the mean NMSE columns, relative changes (\%) are reported with respect to \Model{}. The $\mathrm{MeanRank}$ and $\mathbf{P}_{\mathrm{rank1}}$ values are obtained by ranking the original \Model{} together with all ablation variants to evaluate the relative importance of each module (lower NMSE and $\mathrm{MeanRank}$ indicate better performance, while higher $\mathbf{P}_{\mathrm{rank1}}$ indicates better performance).}}
  \label{tab:ablation-study}
  \centering
  \scriptsize
  \setlength{\tabcolsep}{3pt}
  \renewcommand{\arraystretch}{1.1}
  \begin{tabularx}{\linewidth}{@{} l l c c c c Y Y Y @{}}
    \toprule
    \textbf{Duplexing Mode} & \textbf{Model} &
    \makecell{\textbf{$\Delta$ Inference}\\\textbf{Time} \\ \textbf{(\%)}} &
    \makecell{\textbf{$\Delta$ FLOPs}\\\textbf{(\%)}} &
    \makecell{\textbf{$\Delta$ Total}\\\textbf{Params} \\ \textbf{(\%)}} &
    \makecell{\textbf{$\Delta$ Memory}\\\textbf{Footprint} \\ \textbf{(\%)}} &
    \makecell{\textbf{$\Delta$ NMSE}\\\textbf{(Reg / Rob / Gen)} \\ \textbf{(\%)}} &
    \makecell{$\mathrm{MeanRank}$\\\textbf{(Reg / Rob / Gen)}} &
    \makecell{$\mathbf{P}_{\mathrm{rank1}}$\\\textbf{(Reg / Rob / Gen)} \\ \textbf{(\%)}} \\
    \midrule
    \multicolumn{9}{@{}l}{\textbf{TDD}}\\
    \cmidrule(lr){1-9}
    & \Model{}             &   0.0 &   0.0  &   0.0  &   0.0  &  0.0 /  0.0 /  0.0   & 1.05 / 1.87 / 1.73 & 98.77 / 78.19 / 85.33 \\
    & No CNN               &  -5.1 &   -0.0  &  -0.0  &  -0.0  & 29.27 / 17.27 / 10.54 & 4.78 / 4.65 / 4.74 &  0.00 /  3.29 /  0.88 \\
    & No IDFT              & -37.8 & -33.8  &  -2.6  &  -2.2  & 30.99 / 19.84 / 13.14 & 5.06 / 5.01 / 5.40 &  0.00 /  1.85 /  0.13 \\
    \multicolumn{9}{@{}l}{\emph{ACL variants}}\\
    & No ACL               & -10.3 & -56.3  &  -4.9  &  -4.2  & 33.66 / 22.14 / 11.89 & 5.94 / 5.80 / 5.36 &  0.00 /  1.44 /  3.43 \\
    & Norm Replace         & -18.6 & -56.1  &  -4.9  &  -4.2  &119.05 /104.43 / 43.51 &10.00 / 9.88 / 9.76 &  0.00 /  0.00 /  0.00 \\
    & Add Subcarrier       &  10.5 &   2.7  &   3.4  &   2.9  & 22.54 / 14.39 /  9.19 & 4.05 / 4.77 / 4.02 &  0.00 /  4.12 /  0.20 \\
    \multicolumn{9}{@{}l}{\emph{ShuffleNet variants}}\\
    & MLP Replace & -54.0 &  -9.7  &   1.3  &   1.1  & 26.86 / 17.54 / 13.79 & 4.73 / 5.07 / 5.82 &  0.62 /  3.29 /  0.42 \\
    & MobileNet Replace    & -42.0 &  -6.1  &  -0.2  &  -0.2  & 29.03 / 17.35 / 12.68 & 4.84 / 4.84 / 5.51 &  0.00 /  2.88 /  0.29 \\
    \multicolumn{9}{@{}l}{\emph{Transformer variants}}\\
    & MLP Replace   & -17.2 & -25.9  & -81.5  & -69.4  & 36.51 / 20.07 / 11.37 & 7.37 / 6.39 / 6.41 &  0.62 /  2.67 /  9.05 \\
    & LSTM Replace         & -13.6 & -23.9  & -74.9  & -63.8  & 36.04 / 22.90 / 12.41 & 7.17 / 6.70 / 6.24 &  0.00 /  2.26 /  0.26 \\
    \midrule
    \multicolumn{9}{@{}l}{\textbf{FDD}}\\
    \cmidrule(lr){1-9}
    & \Model{}             &   0.0 &   0.0  &   0.0  &   0.0  &  0.0 /  0.0 /  0.0   & 1.09 / 1.17 / 1.58 & 92.59 / 86.42 / 64.87 \\
    \multicolumn{9}{@{}l}{\emph{ACL variants}}\\
    & No ACL               & -11.4 & -53.4  & -37.3  & -33.9  & 38.79 / 41.29 /  3.29 & 2.22 / 2.28 / 2.12 &  6.17 /  9.47 / 11.08 \\
    & No Subcarrier        & -11.2 & -12.9  & -34.5  & -31.3  & 45.77 / 43.55 /  2.73 & 2.69 / 2.56 / 2.30 &  1.23 /  4.12 / 24.05 \\
    \bottomrule
  \end{tabularx}
\end{table*}

\response{Overall, removing any of the main modules consistently degrades performance across the \Regular{}, \Robustness{}, and \Generalization{} tracks. Removing the initial CNN block (i.e., \textit{No CNN} ablation) increases the NMSE by 29.27\%/17.27\%/10.54\% on the \Regular{}, \Robustness{}, and \Generalization{} tracks, respectively. However, the computational cost remains almost the same, with only 5\% improvement in inference time and without a observable change in FLPOS and memory. This suggests that the CNN is a crucial component and can, to a large extent, be regarded as a ``free lunch'' for the performance improvement of \Model{}.}

\response{Although removing the IDFT reduces inference latency by 37.8\% and FLOPs by 33.8\%, the resulting performance degradation is substantial: NMSE increases by 30.99\%, 19.84\%, and 13.14\% on the \Regular{}, \Robustness{}, and \Generalization{} tracks, respectively. This result indicates that delay-domain information is an important complement to frequency-domain information and provides significant benefits for \Model{}'s prediction.}

\response{Among all components, ACL is the most complicated, and modifying it in an inappropriate manner leads to the most severe performance degradation. First, the \textit{No ACL} variant in both the TDD and FDD settings produces considerable performance degradation (33.66\%/22.14\%/11.89\% for TDD and 38.79\%/41.29\%/3.29\% for FDD), which clearly justifies the effectiveness of ACL. Additionally, both introducing the subcarrier ACL in TDD and removing the subcarrier ACL in FDD lead to performance degradation, which reveals that the design of the ACL module must align with the characteristics of the specific duplexing mode. In TDD, the frequency channel in the historical window is exactly the same as that in the future window, and introducing the subcarrier ACL mixes the channel information and breaks the correspondence between the historical and future channels. In FDD, since the prediction is also cross-band, it is highly important to learn features from all subcarriers and taps for accurate prediction, and removing the subcarrier ACL breaks this property. Moreover, it is particularly notable that the \textit{Norm Replace ACL} variant in the TDD setting causes a collapse in performance, increasing the NMSE by 119.05\%/104.43\%/43.51\%, and ranking the bottom among all 10 TDD ablations. In brief, using normalization along the temporal dimension to replace ACL can be regarded as an ``aggressive'' active distortion, which is more destructive than simply omitting ACL. It removes the temporal mean, absolute scale, and relative temporal variance of each subcarrier/tap channel, which is exactly the reverse of ACL: it makes the representation flatter and removes informative structure. More detailed discussion and visualization can be found in Appendix~\ref{sec:appendix-ablation-norm-replace-acl}.}

\response{In addition, replacing ShuffleNet also leads to performance degradation. The poor performance of the \textit{MobileNet Replace ShuffleNet} variant, with NMSE increases of 29.03\%/17.35\%/12.68\%, reveals that not all lightweight convolutional backbones can achieve a favorable balance between accuracy and efficiency; empirically, the specific channel interaction mechanism in ShuffleNet is better suited to the task considered in this paper. Although \textit{MobileNet Replace ShuffleNet} reduces inference time by 42\% and FLOPs by 6.2\%, its substantially worse performance compared with \Model{} makes it an unattractive option. Similarly, the \textit{MLP Replace ShuffleNet} variant also leads to considerable performance degradation (26.86\%/17.54\%/13.79\%), indicating that ignoring the local 2D structure across time and subcarriers is unsuitable for the CSI prediction task.}

\response{Furthermore, the ablation variants related to the Transformer module result in the largest performance drops, except for the \textit{Norm Replace ACL} variant. Both the \textit{MLP Replace Transformer} and \textit{LSTM Replace Transformer} variants rank around 7th among the TDD ablations and exhibit more than 36\% NMSE increase on the \Regular{} track. These results highlight the importance of the Transformer and further demonstrate that attention mechanisms and sequence modeling are crucial for the CSI prediction task.}

\response{In summary, the ablation results demonstrate that \Model{} benefits from the complementarity of its design choices: CNN-based residual refinement for noisy inputs, IDFT-based delay-domain representation, ACL-based adaptive calibration, and a ShuffleNet-Transformer backbone for efficient local-global feature extraction.}
\section{Conclusion and Discussion}
\label{sec:conclusion}

This paper introduced \Model{}, a lightweight hybrid deep learning model for CSI prediction, together with \dataset{}, a large-scale benchmark comprising more than 300,000 instances across 3,060 scenarios for both training and evaluation. \response{The architecture of \Model{} is designed to align with the structure of CSI and leverages CSI-specific priors to achieve high predictive accuracy while maintaining deployment efficiency.} Experimental results show that \Model{} consistently outperforms baseline methods under diverse testing conditions in both TDD and FDD. In particular, \Model{} attains the lowest NMSE in \response{81.5\%} of TDD scenarios and \response{44.4\%} of FDD scenarios, while reducing FLOPs by factors of 5 and 3 compared with the strongest competing model in TDD and FDD, respectively.

\response{The detailed evaluation also provided several substantive insights into CSI prediction. Variations in SNR and user velocity produced relatively smooth performance trends, whereas shifts in channel model and delay spread were substantially more disruptive. NLOS channels were consistently more difficult than LOS channels, and unseen large delay spreads caused pronounced degradation, particularly in FDD. The comparison between duplexing modes further confirmed that inter-band UL-to-DL prediction (FDD) is fundamentally harder than intra-band prediction (TDD). In addition, the robustness analysis showed that performance depends not only on noise intensity but also on noise structure: burst noise caused larger degradation than phase noise at matched SNR, and packet-drop noise was more harmful in TDD, highlighting the importance of temporal continuity for same-band prediction.}

\response{The ablation study further showed that the observed gains do not arise from simply stacking generic modules. Removing the CNN front-end, the delay-domain branch, or the ACL consistently degraded performance, and replacing ShuffleNet or the Transformer with alternative modules also reduced accuracy. The duplex-aware ACL design was especially important: adding subcarrier-wise correction in TDD hurt performance, whereas removing it in FDD also hurt performance. These results support the architectural rationale of \Model{} and indicate that the performance gains arise from the coordinated use of channel-informed representation, axis-wise correction, and efficient local-global modeling.}

\response{One of the limitations of the present study is that \Model{} does not explicitly incorporate noise-type-specific modules for phase noise, burst noise, or packet drops. Instead, robustness is achieved through a unified, noise-agnostic architecture that combines a CNN residual front-end with delay-domain transformation, adaptive correction, and temporal sequence modeling. This design is practically meaningful when the active corruption type is unknown at inference time, but it does not explicitly exploit the distinct statistical structure of each non-Gaussian impairment. A natural next step is therefore to develop lightweight corruption-aware extensions that preserve the generality of the current framework while improving robustness under severe or previously unseen disturbances, for example by integrating noise-aware adaptation or specialized refinement mechanisms into the existing architecture.}

\response{A further limitation is that the present study remains based on a standardized simulation benchmark rather than measured field data. Although this setting enables controlled, reproducible, and large-scale evaluation across diverse channel models, delay spreads, user velocities, duplexing modes, and non-Gaussian perturbations, a sim-to-real gap is still expected in practical deployments. Measured CSI may reflect additional sources of distribution shift, including site-specific propagation effects beyond stochastic abstractions, hardware and measurement-chain impairments, implementation-dependent CSI extraction and calibration, and, in FDD, quantization and reporting constraints. Consequently, while the reported results provide strong evidence of comparative robustness and generalization under standardized simulation, absolute NMSE/SE—and potentially even relative model ranking—may change on measured CSI. Narrowing this gap will require both improved data generation and data-efficient adaptation. Promising directions include hardware-in-the-loop dataset generation with real RF components, site-specific channel synthesis using radio-frequency digital twins or ray-tracing, and calibration of synthetically pretrained models using limited real data through fine-tuning, meta-learning, or domain adaptation.}

\response{In addition, another limitation lies in the limited cross-scenario generalization observed in FDD. Owing to the inherent cross-band nature of FDD prediction, all models suffer notable performance degradation when evaluated on unseen delay spreads and channel models, especially under challenging shifts such as NLOS generalization and extrapolation. None of the models learn a universal cross-band mapping that transfers reliably across deployment scenarios. These findings suggest that robust FDD generalization remains an open problem, likely requiring further research into distribution-shift detection, online adaptation, or scenario-specific updating.}

\response{Overall, the present results establish a strong benchmark-driven foundation for robust and efficient CSI prediction, while also making the remaining challenges more explicit, including cross-scenario generalization, real-world deployment, and the design of noise-specific modules. Future research will extend this work by incorporating measured CSI datasets, lightweight corruption-aware extensions, and data-efficient adaptation strategies—such as domain adaptation, fine-tuning, continual updating, and active learning—to enable more adaptive, reliable, and deployment-oriented CSI prediction in practical wireless systems.}

\section*{Acknowledgments}
This research was partly supported by NSF award 2112533.


\appendices

\section{Computational Overhead}
\label{sec:appendix-computational-overhead}

\begin{table*}[!t]
  \caption{\textbf{\response{Computational overhead analysis.}}}
  \label{tab:computational-overhead}
  \centering
  \scriptsize
  \setlength{\tabcolsep}{3pt} 
  \renewcommand{\arraystretch}{1.1}
  \begin{tabularx}{\linewidth}{@{} l l *{6}{Y} @{}}
    \toprule
    \makecell{\textbf{Duplexing}\\\textbf{Mode}} & \textbf{Model} &
    \makecell{\textbf{Trainable}\\\textbf{Params} \\ \textbf{(M)}} &
    \makecell{\textbf{Total}\\\textbf{Params} \\ \textbf{(M)}} &
    \makecell{\textbf{FLOPs}\\\textbf{(G)}} &
    \makecell{\textbf{Inference}\\\textbf{Time} \\ \textbf{(ms)}} &
    \makecell{\textbf{Training}\\\textbf{Time} \\ \textbf{(ms)}} &
    \makecell{\textbf{Memory}\\\textbf{Footprint}\\\textbf{(MiB)}} \\
    \midrule
    \multicolumn{8}{@{}l}{\textbf{TDD}}\\
    \cmidrule(lr){1-8}
     & NP       & 0 & 0 & 0      & 0.042  & -  & 0      \\
     & AR       & 0 & 0 & 0.108  & 2.818  & -  & 25.854 \\
     & WIENER   & 0 & 0 & 0.157  & 2.792  & -  & 151.465 \\
     & PAD      & 0 & 0 & 14.75  & 61.808 & - & 1406.25 \\
     & CNN      & 0.197 & 0.197 & 60.45  & 0.968  & 0.976  & 0.752 \\
     & RNN      & 156.312 & 156.312 & 190.04 & 6.881  & 6.936  & 596.284 \\
     & STEMGNN  & 2.345 & 2.345 & 91.62  & 11.918 & 12.105 & 8.943 \\
     & LLM4CP   & 4.532 & 144.140 & 366.96 & 15.194 & 15.528 & 576.383 \\
     & \Model{} & 21.914 & 21.914 & 71.90 & 8.099  & 10.668 & 98.255 \\
     \midrule
     \multicolumn{8}{@{}l}{\textbf{FDD}}\\
     \cmidrule(lr){1-8}
     & NP       & 0 & 0 & 0      & 0.042  & -  & 0      \\
     & WIENER   & 0 & 0 & 0.157  & 2.792  & -  & 151.465 \\
     & CNN      & 0.197 & 0.197 & 60.45  & 0.968  & 0.976  & 0.752 \\
     & RNN      & 48.981 & 48.981 & 59.54 & 4.539  & 5.220  & 186.847 \\
     & STEMGNN  & 5.754 & 5.754 & 222.34 & 16.469 & 16.677 & 21.948 \\
     & LLM4CP   & 3.811 & 92.002 & 372.21 & 16.118 & 16.466 & 372.608 \\
     & \Model{} & 38.636 & 38.636 & 101.64 & 18.698 & 20.486 & 162.059 \\
    \bottomrule
  \end{tabularx}
\end{table*}

\response{Table~\ref{tab:computational-overhead} presents the computational overhead of the proposed model in comparison with the baseline models. The reported metrics include trainable parameters (M), total parameters (M), FLOPs (G), inference time (ms), training time (ms), and static memory footprint (MiB) for both TDD and FDD duplexing modes. The FLOPs and runtime metrics are reported for one whole CSI sequence. Runtime values are obtained by averaging each measurement over 100 repeated executions under the same hardware/software environment. The memory footprint corresponds to the static memory required to store the predictor together with its persistent buffers/state, and therefore serves as a deployment-feasibility indicator complementary to FLOPs and latency. Because hyperparameter tuning is performed independently for models trained on the TDD and FDD datasets, the same model may yield different results across the two duplexing modes.}
\section{Autocorrelation Function (ACF) across different user velocities}
\label{sec:appendix-acf-user-velocities}

\begin{figure}[h]
  \centering
  \subfloat[1\,m/s]{
    \includegraphics[width=0.45\linewidth]{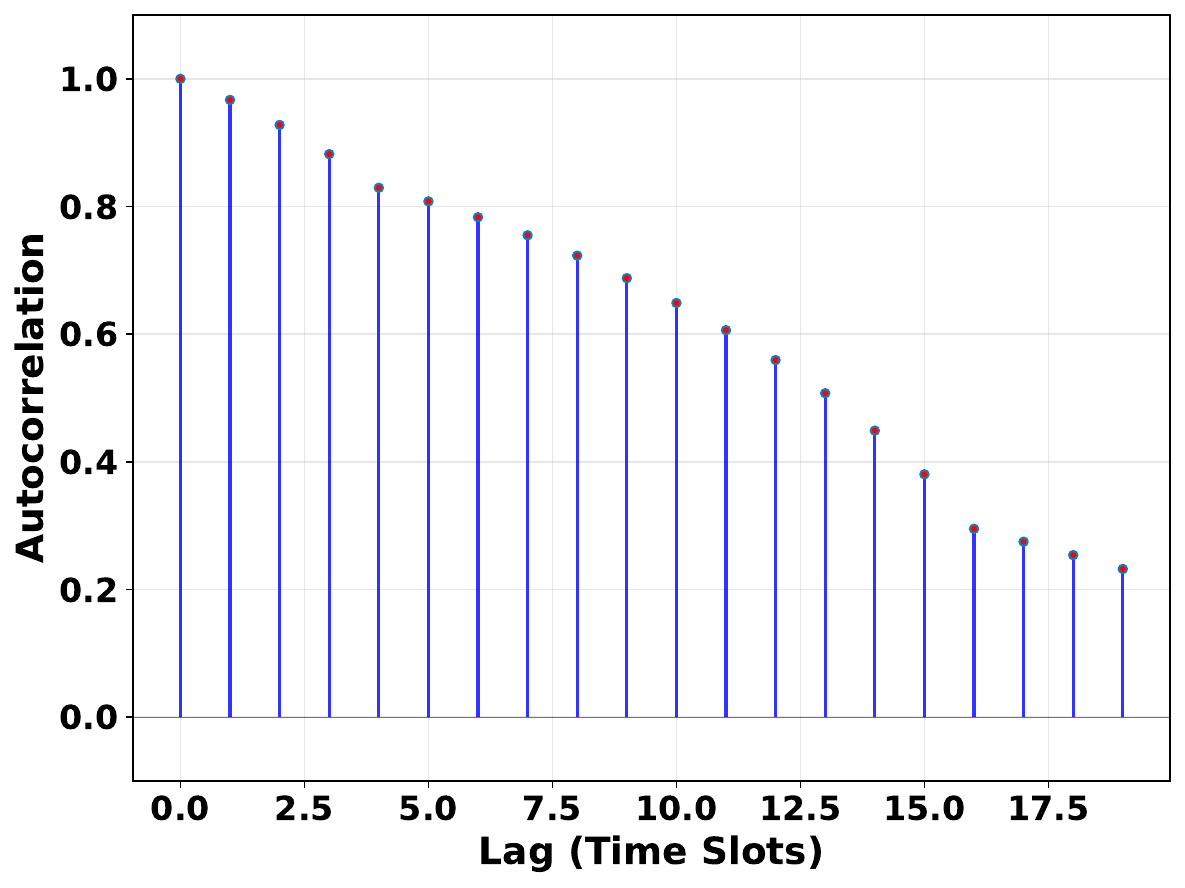}
    \label{fig:fdd_acf_cm_a_ds_30ns_ms_1mps_gen}
  }
  \subfloat[3\,m/s]{
    \includegraphics[width=0.45\linewidth]{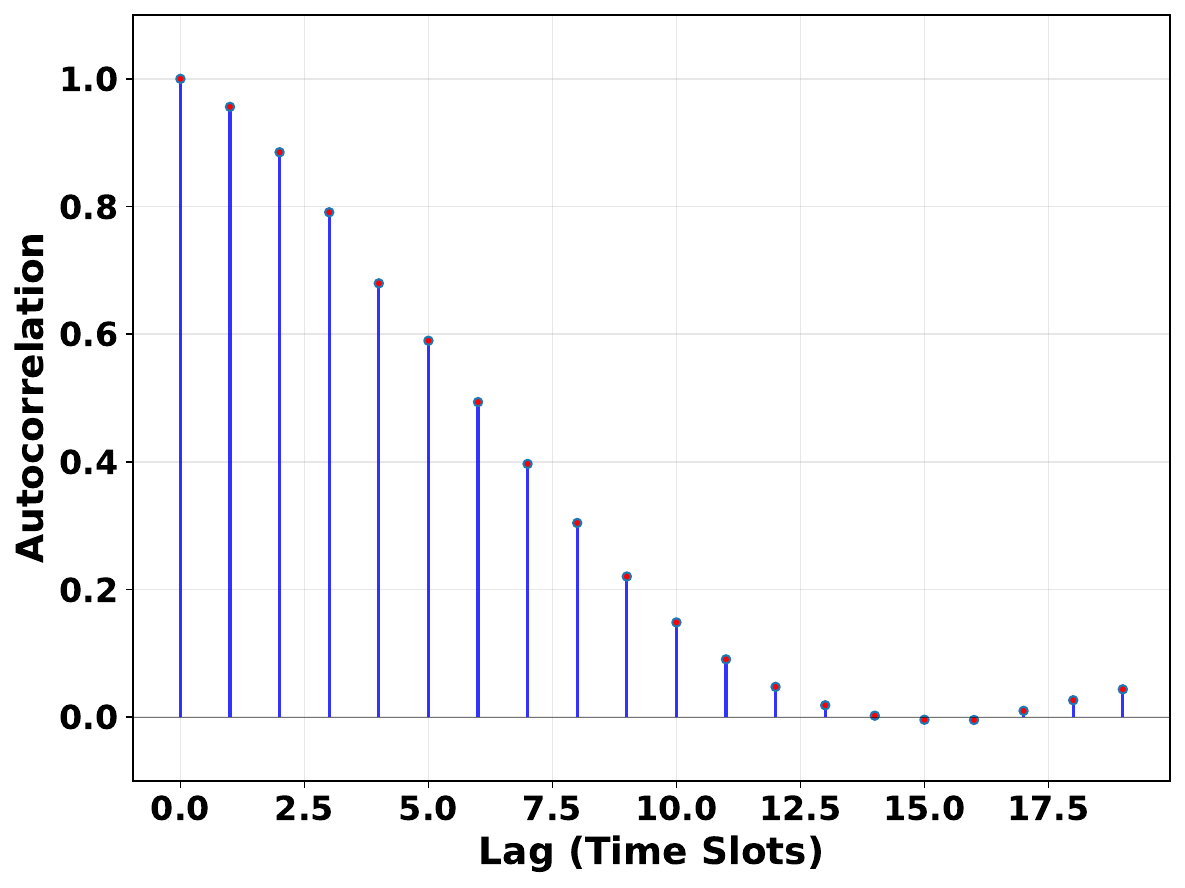}
    \label{fig:fdd_acf_cm_a_ds_30ns_ms_3mps_gen}
  }\
  \subfloat[6\,m/s]{
    \includegraphics[width=0.45\linewidth]{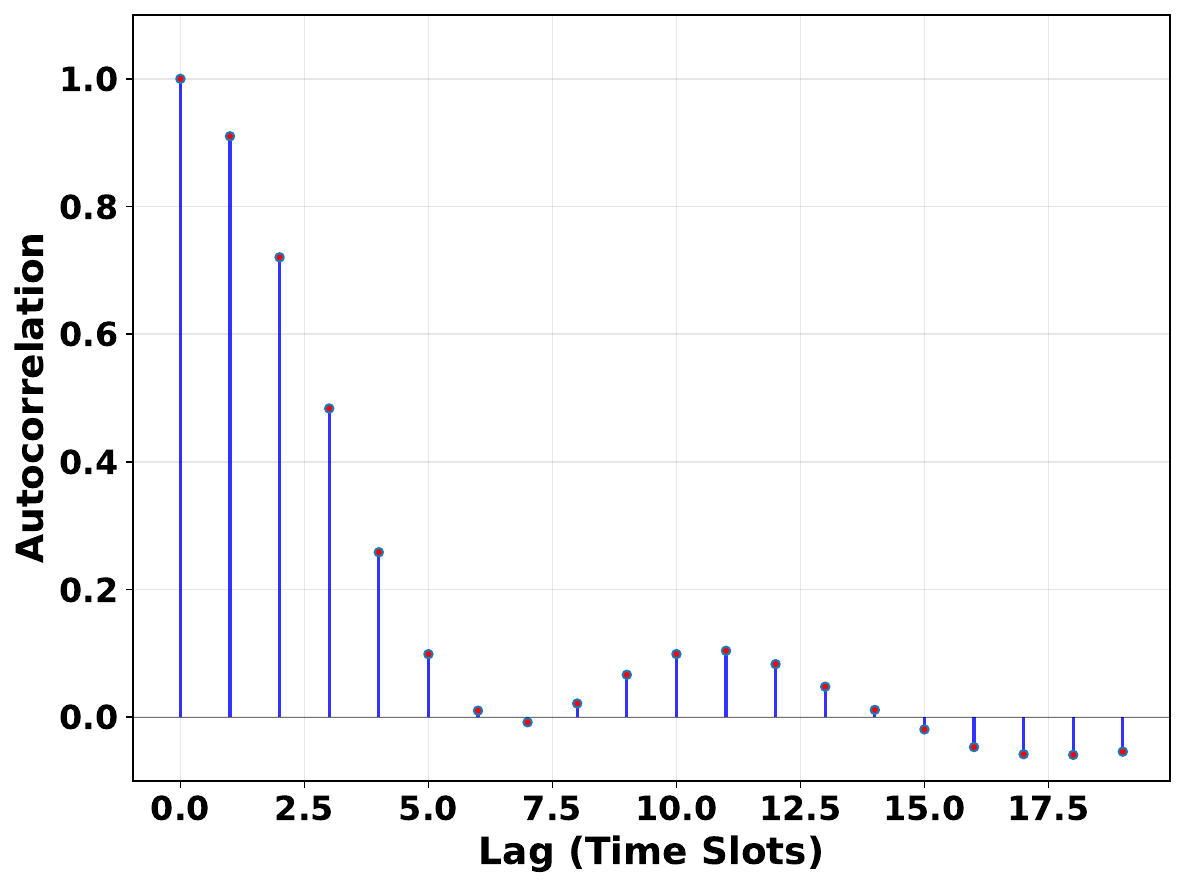}
    \label{fig:fdd_acf_cm_a_ds_30ns_ms_6mps_gen}
  }
  \subfloat[12\,m/s]{
    \includegraphics[width=0.45\linewidth]{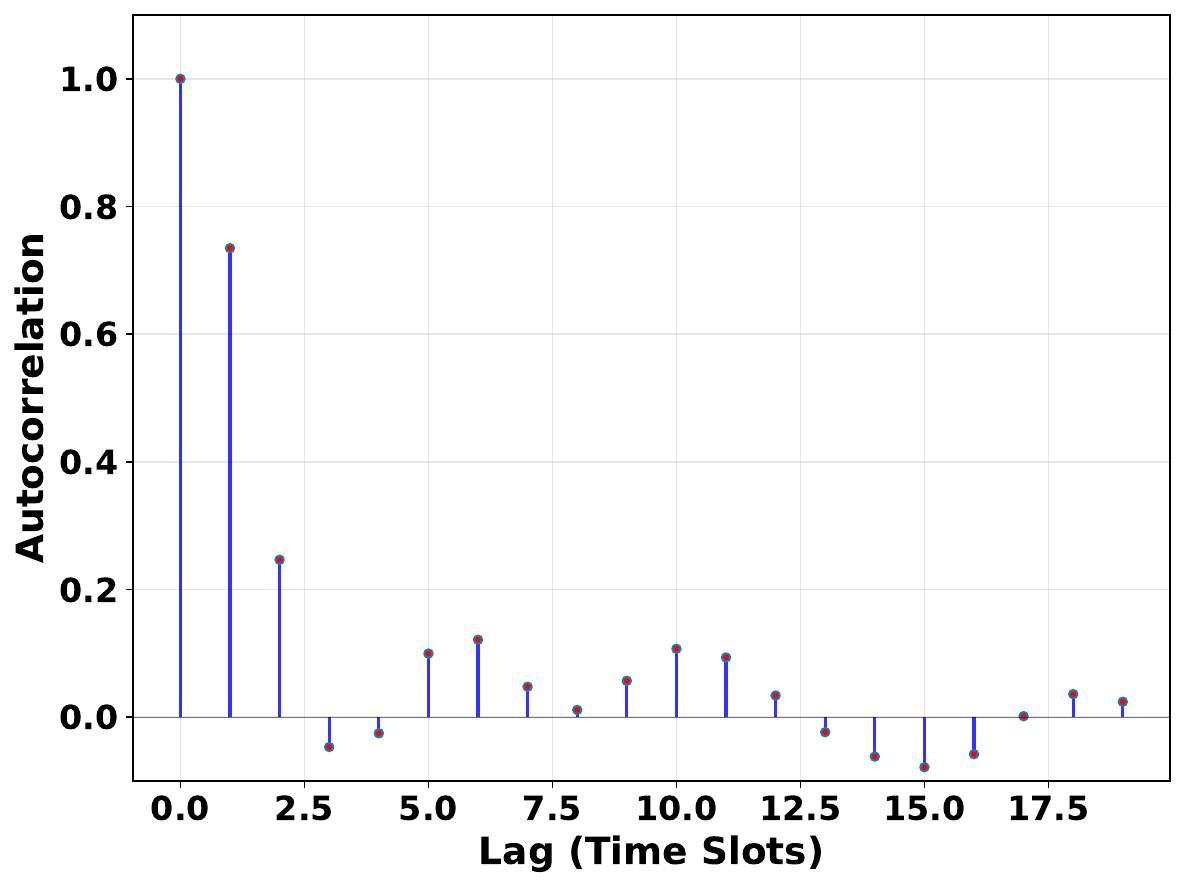}
    \label{fig:fdd_acf_cm_a_ds_30ns_ms_12mps_gen}
  }\
  \subfloat[30\,m/s]{
    \includegraphics[width=0.45\linewidth]{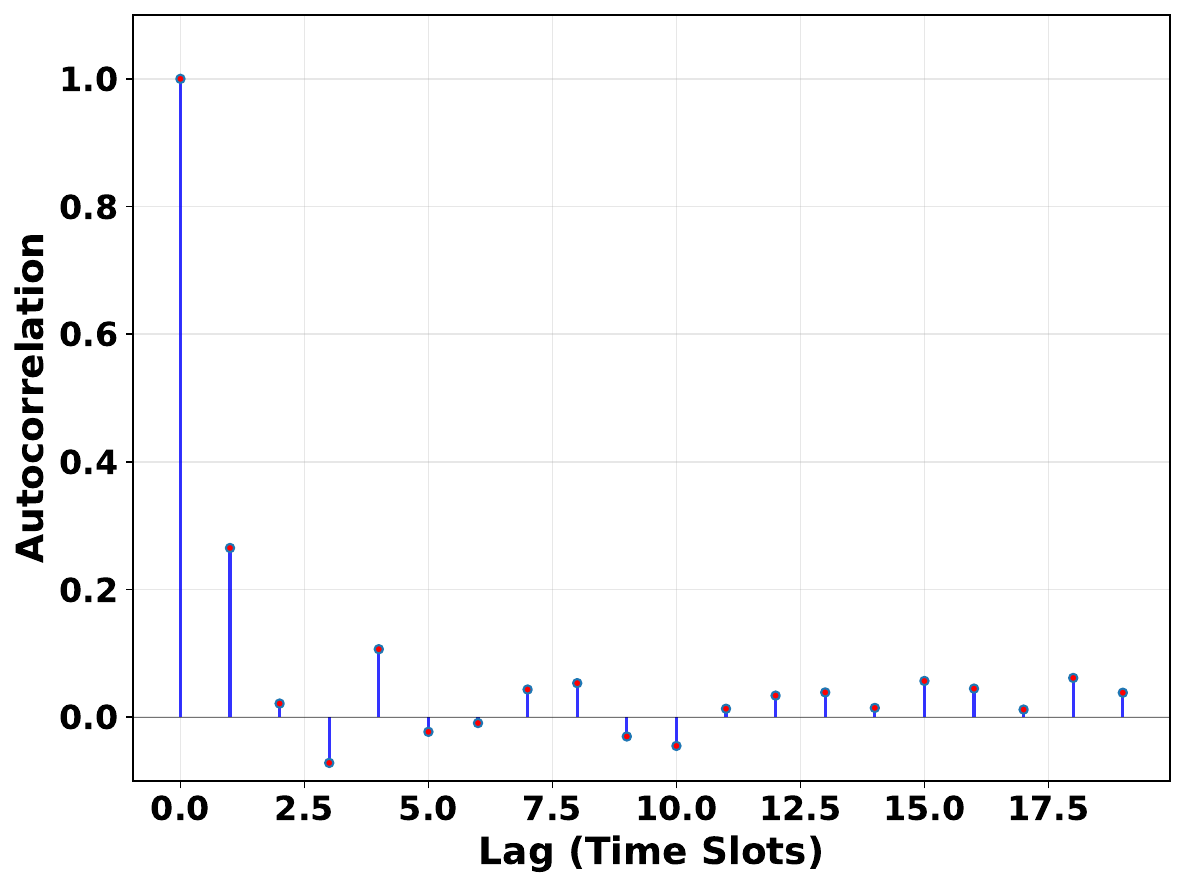}
    \label{fig:fdd_acf_cm_a_ds_30ns_ms_30mps_gen}
  }
  \subfloat[45\,m/s]{
    \includegraphics[width=0.45\linewidth]{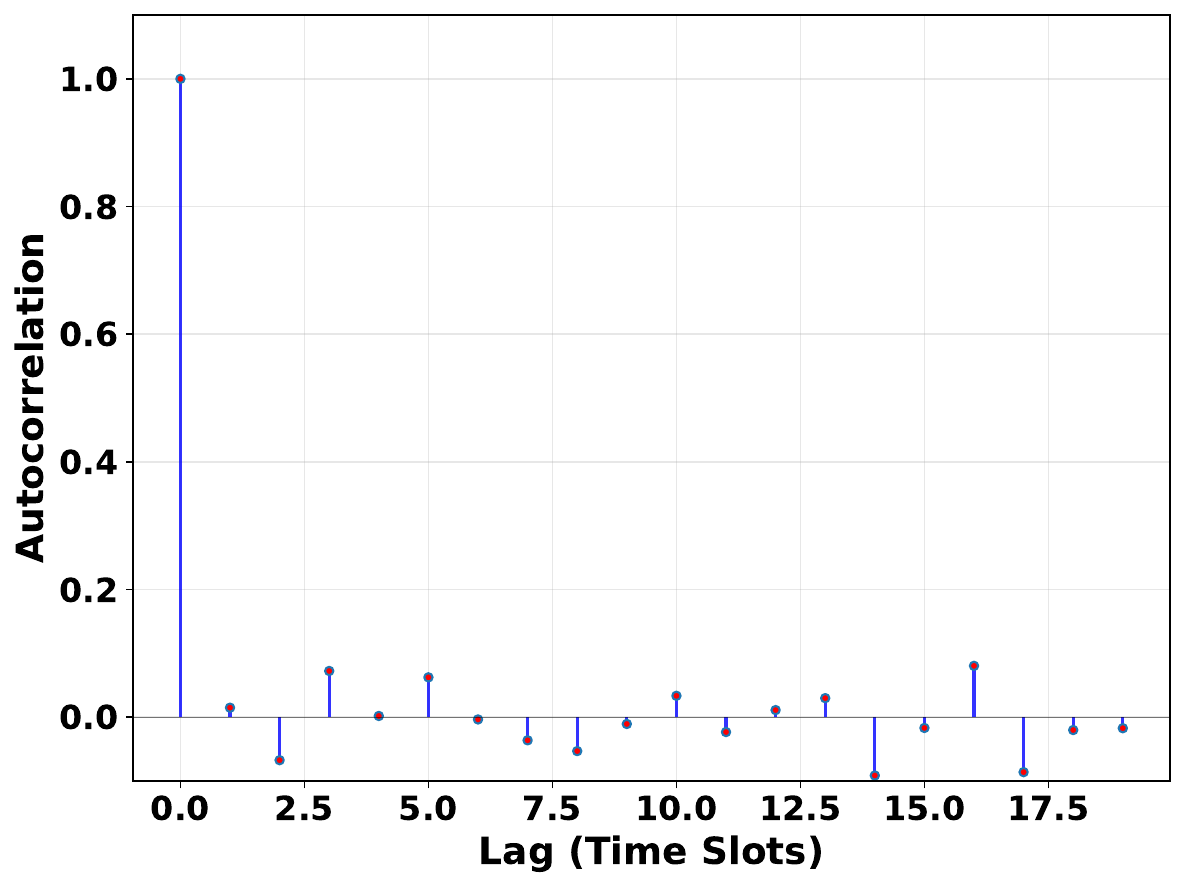}
    \label{fig:fdd_acf_cm_a_ds_30ns_ms_45mps_gen}
  }
  \caption{\textbf{ACF across different user velocities (FDD | CDL-A | 30ns)}}
  \label{fig:acf-across-user-velocities-fdd}
\end{figure}

\begin{figure}[h]
  \centering
  \subfloat[1\,m/s]{
    \includegraphics[width=0.45\linewidth]{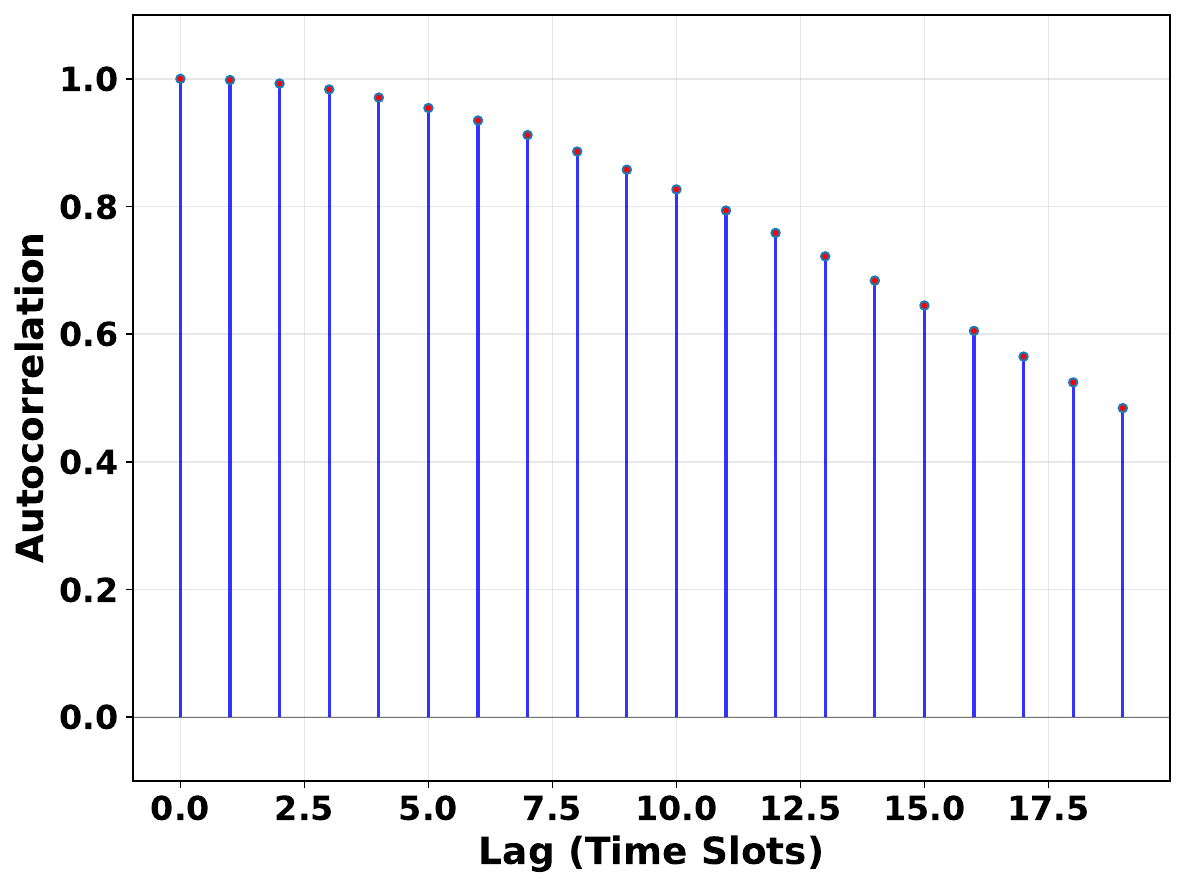}
    \label{fig:tdd_acf_cm_a_ds_30ns_ms_1mps_gen}
  }
  \subfloat[3\,m/s]{
    \includegraphics[width=0.45\linewidth]{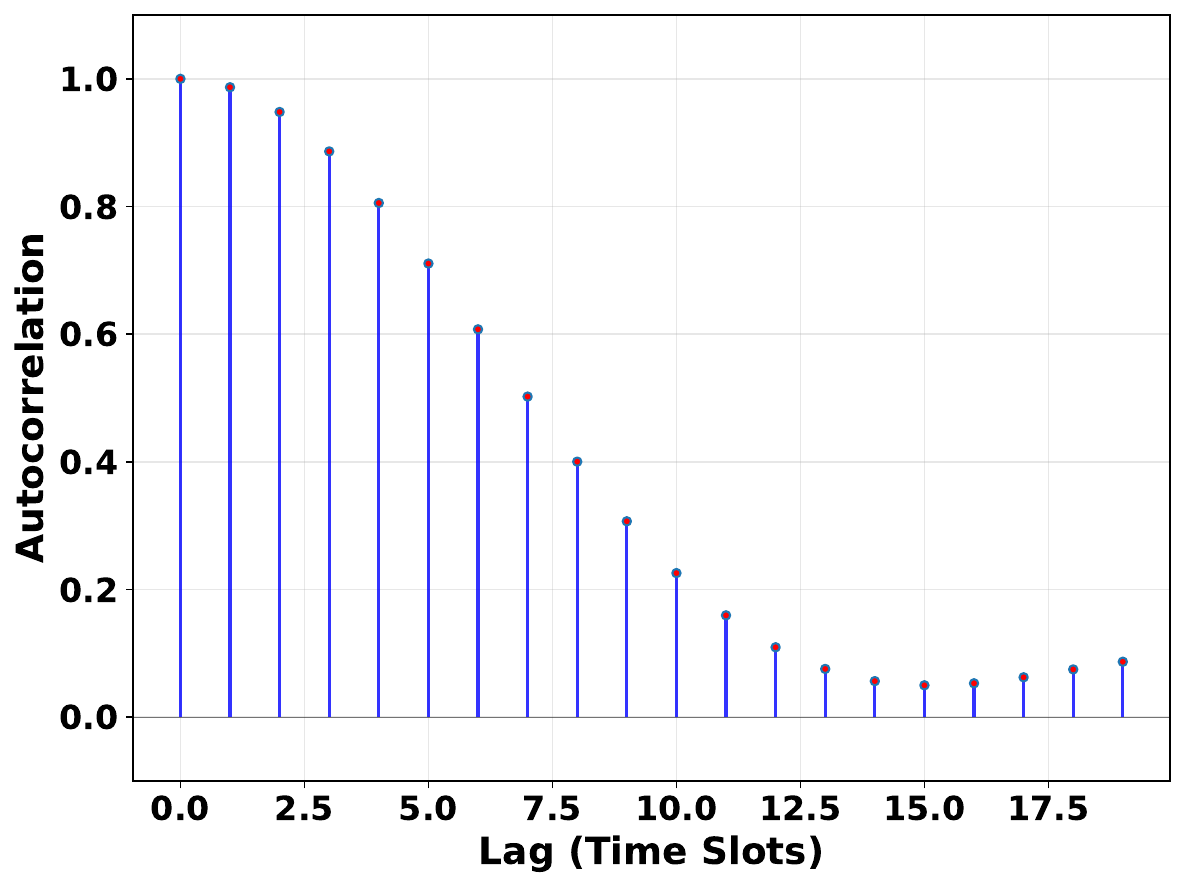}
    \label{fig:tdd_acf_cm_a_ds_30ns_ms_3mps_gen}
  }\
  \subfloat[6\,m/s]{
    \includegraphics[width=0.45\linewidth]{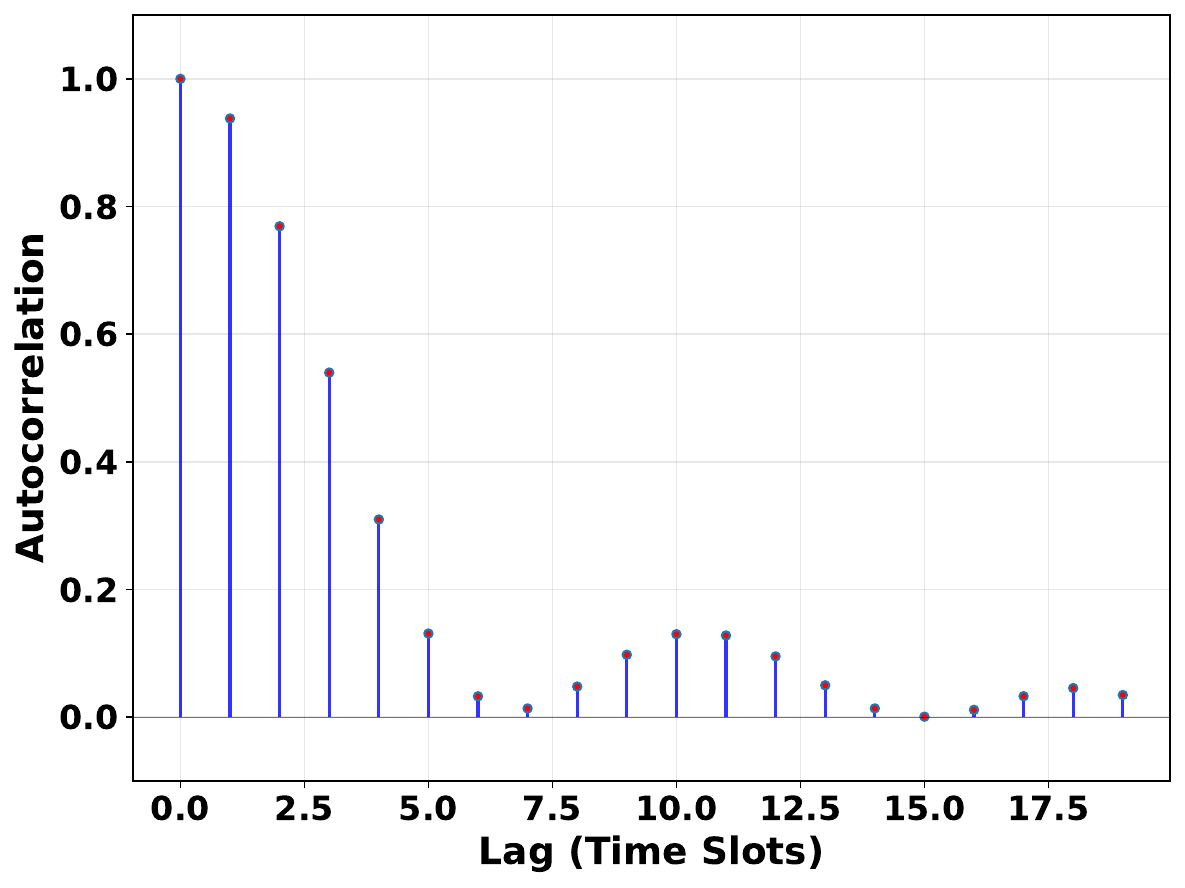}
    \label{fig:tdd_acf_cm_a_ds_30ns_ms_6mps_gen}
  }
  \subfloat[12\,m/s]{
    \includegraphics[width=0.45\linewidth]{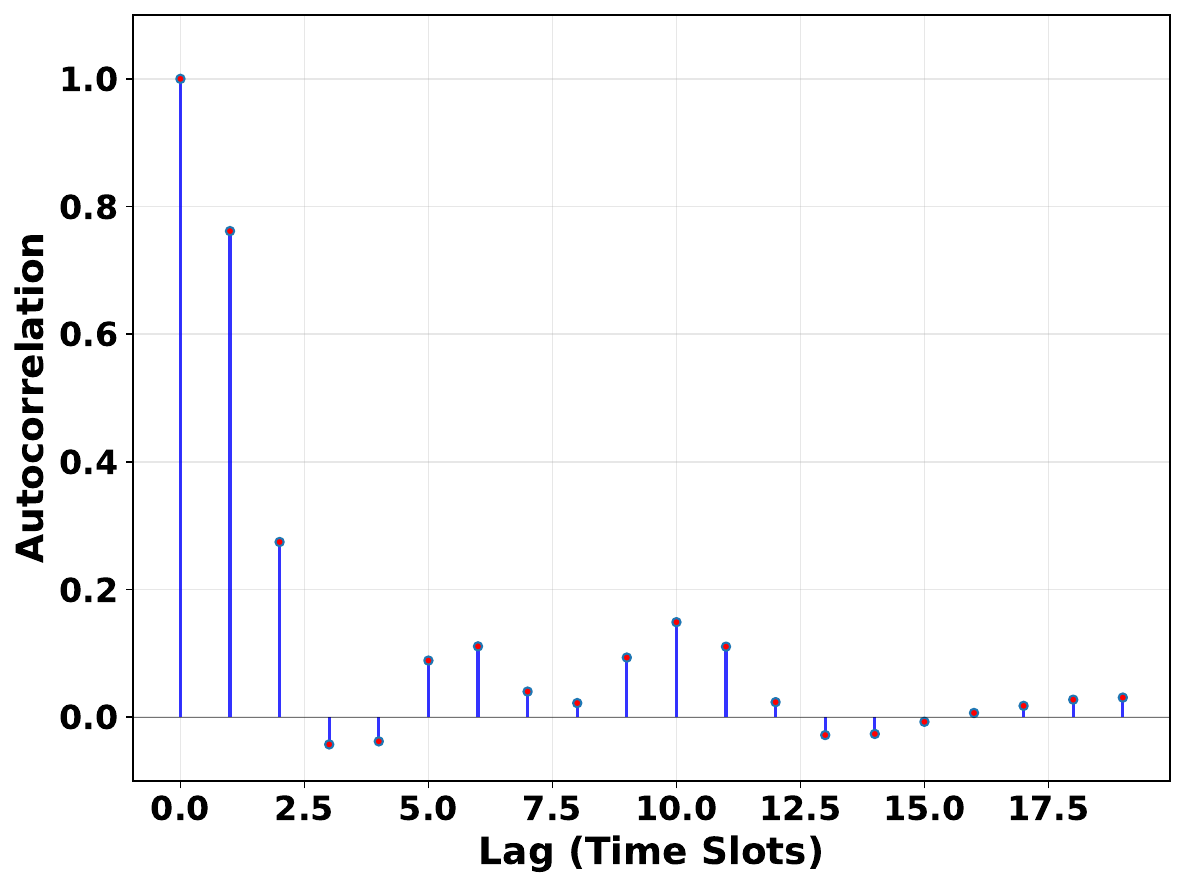}
    \label{fig:tdd_acf_cm_a_ds_30ns_ms_12mps_gen}
  }\
  \subfloat[30\,m/s]{
    \includegraphics[width=0.45\linewidth]{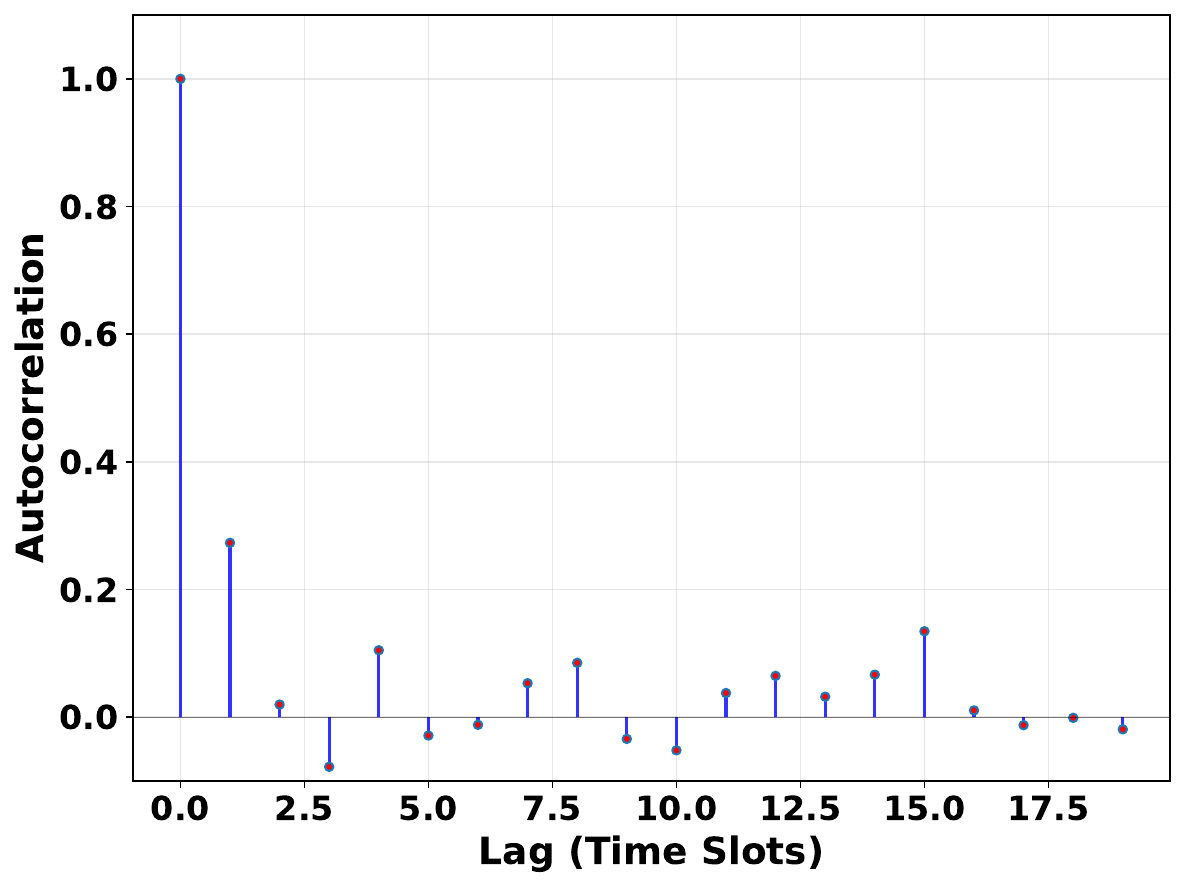}
    \label{fig:tdd_acf_cm_a_ds_30ns_ms_30mps_gen}
  }
  \subfloat[45\,m/s]{
    \includegraphics[width=0.45\linewidth]{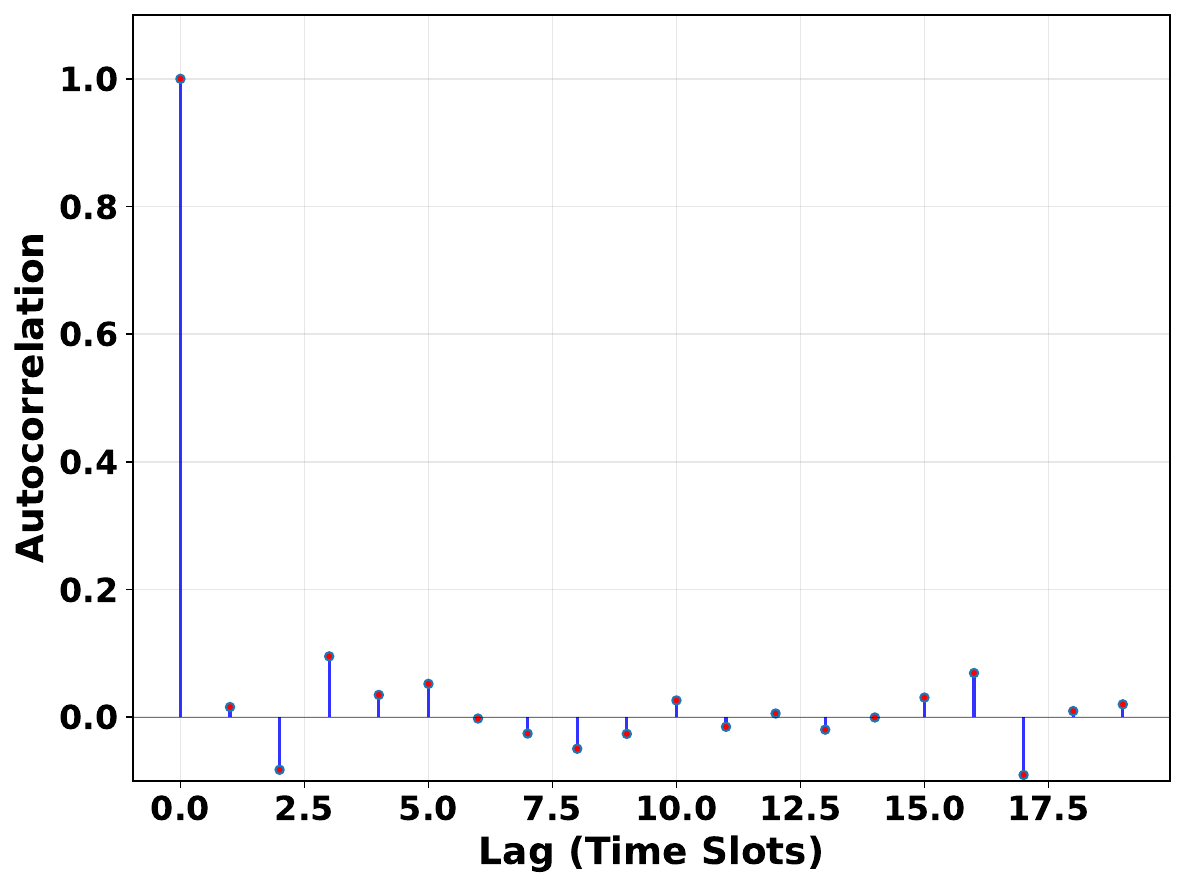}
    \label{fig:tdd_acf_cm_a_ds_30ns_ms_45mps_gen}
  }
  \caption{\textbf{ACF across different user velocities (TDD | CDL-A | 30ns)}}
  \label{fig:acf-across-user-velocities-tdd}
\end{figure}

This section presents the autocorrelation function (ACF) of CSI across user velocities. Figs.~\ref{fig:acf-across-user-velocities-fdd}–\ref{fig:acf-across-user-velocities-tdd} report results for FDD and TDD under CDL-A with a 30\,ns delay spread. For each duplexing mode and velocity, the test tensor has shape $100\times 32\times 20\times 300$ (samples $\times$ antennas $\times$ timestamps $\times$ subcarriers). The data are decomposed into per-(antenna, subcarrier) time series of length 20; the sample ACF is computed for each series and then averaged across all samples, antennas, and subcarriers. The resulting mean ACF is shown as a stem plot.

Two patterns emerge. (i) At low speed, both TDD and FDD exhibit pronounced temporal correlation (slow ACF decay), with TDD showing stronger correlation than FDD. (ii) At high speed, temporal correlation diminishes rapidly in both duplexing modes.

\section{Autocorrelation Function (ACF) across different delay spreads}
\label{sec:appendix-acf-delay-spreads}

\begin{figure}[!h]
  \centering
  \subfloat[30ns]{
    \includegraphics[width=0.45\linewidth]{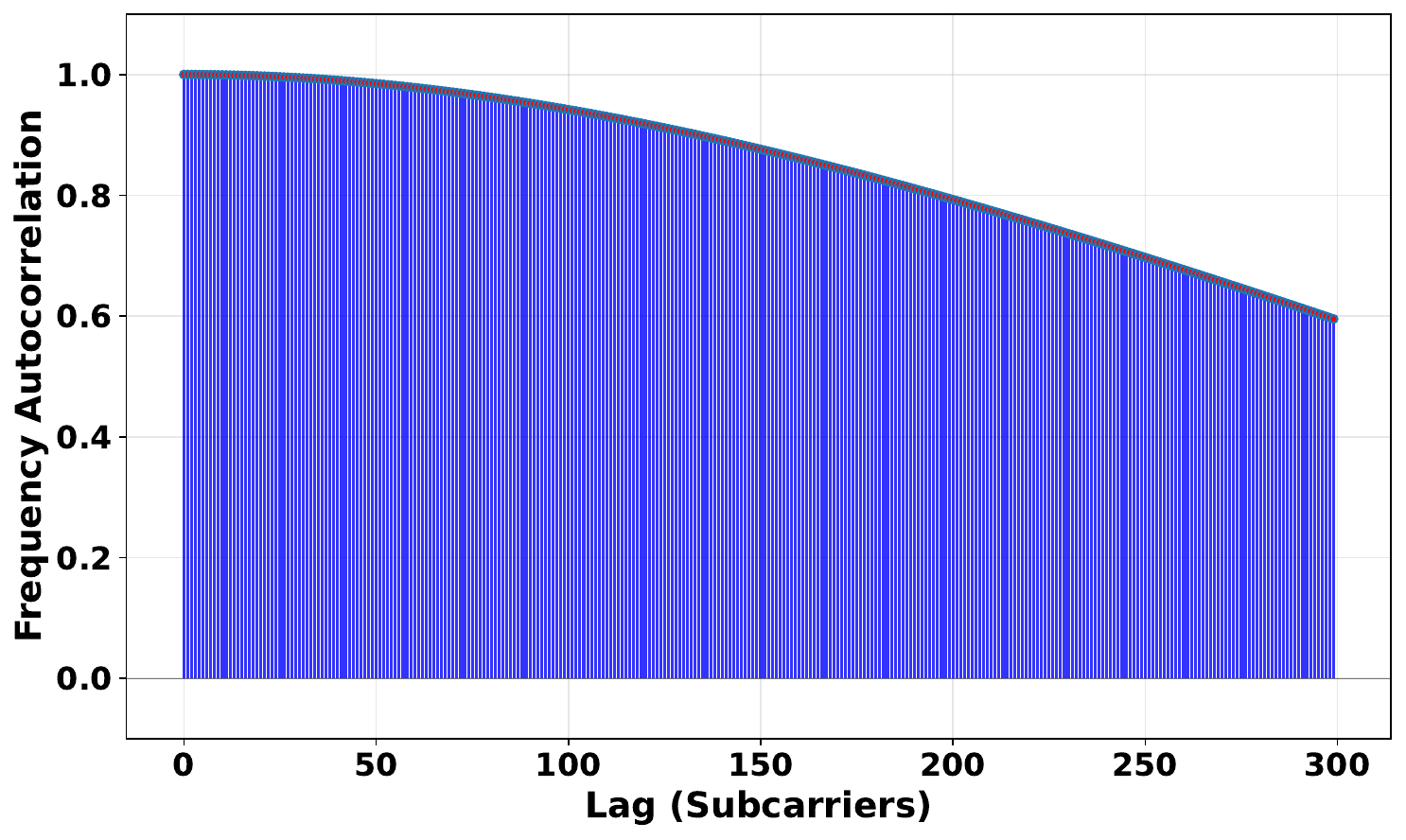}
    \label{fig:fdd_freq_acf_cm_c_ds_30ns_ms_1mps_gen_ds}
  }
  \subfloat[50ns]{
    \includegraphics[width=0.45\linewidth]{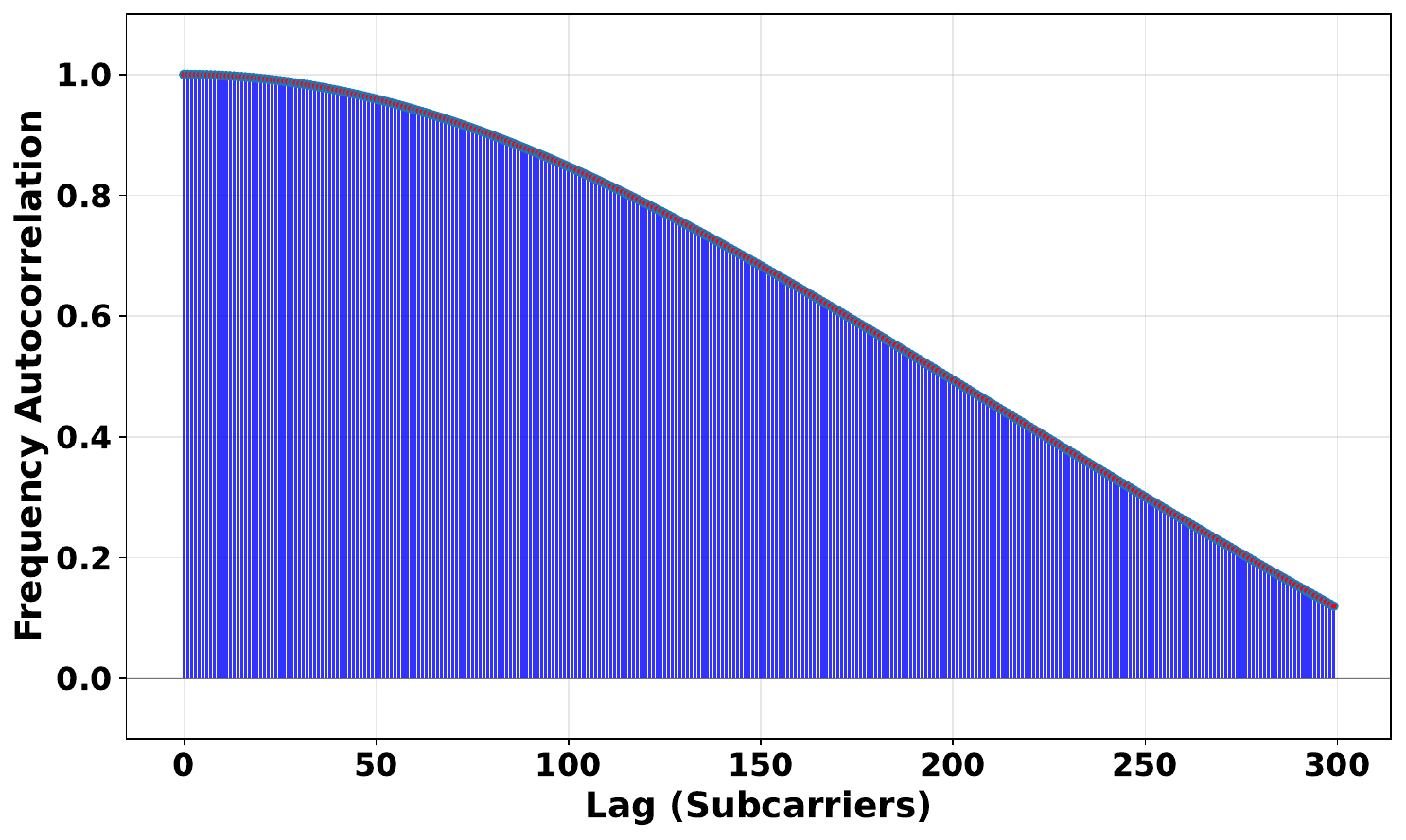}
    \label{fig:fdd_freq_acf_cm_c_ds_50ns_ms_1mps_gen_ds}
  }\
  \subfloat[100ns]{
    \includegraphics[width=0.45\linewidth]{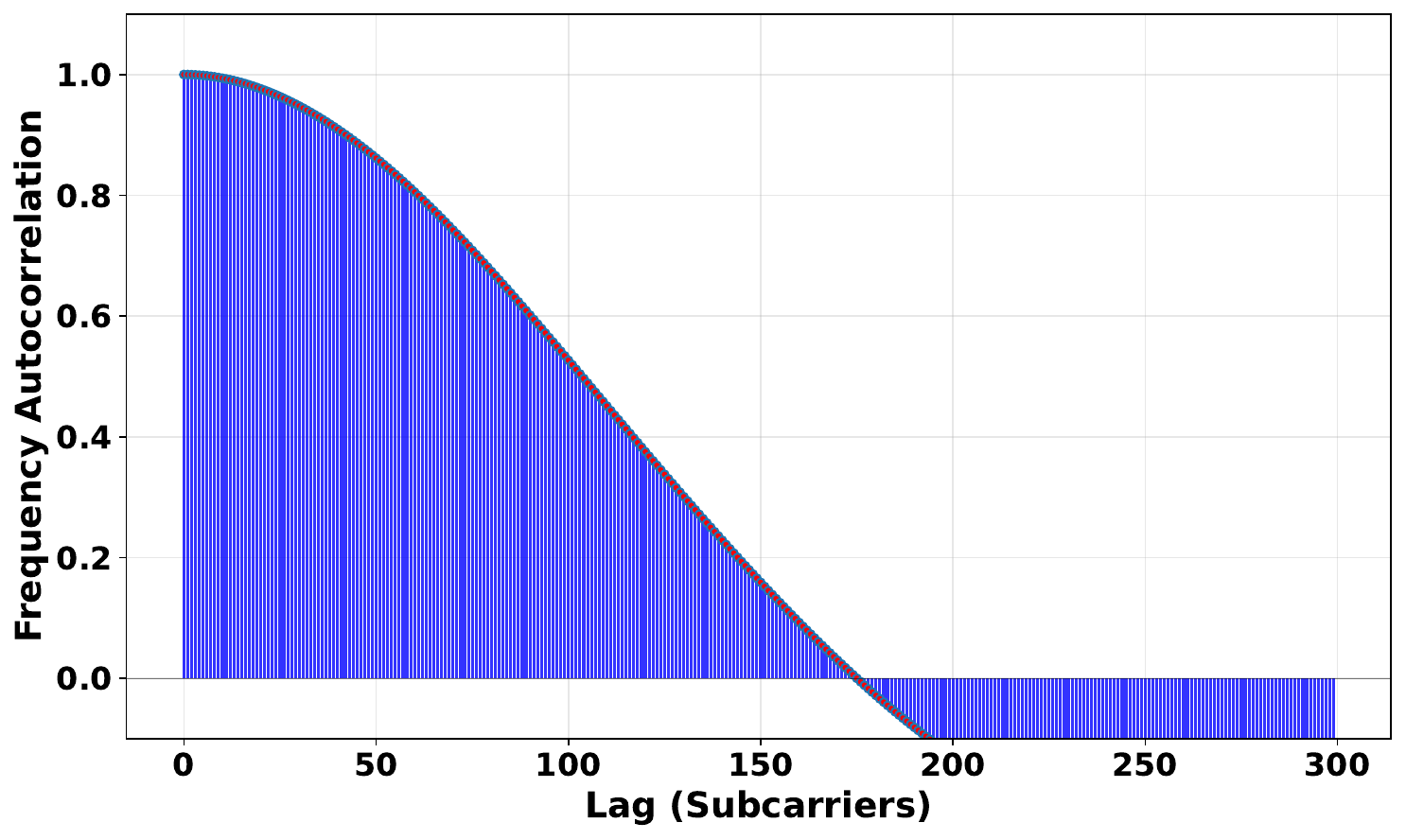}
    \label{fig:fdd_freq_acf_cm_c_ds_100ns_ms_1mps_gen_ds}
  }
  \subfloat[200ns]{
    \includegraphics[width=0.45\linewidth]{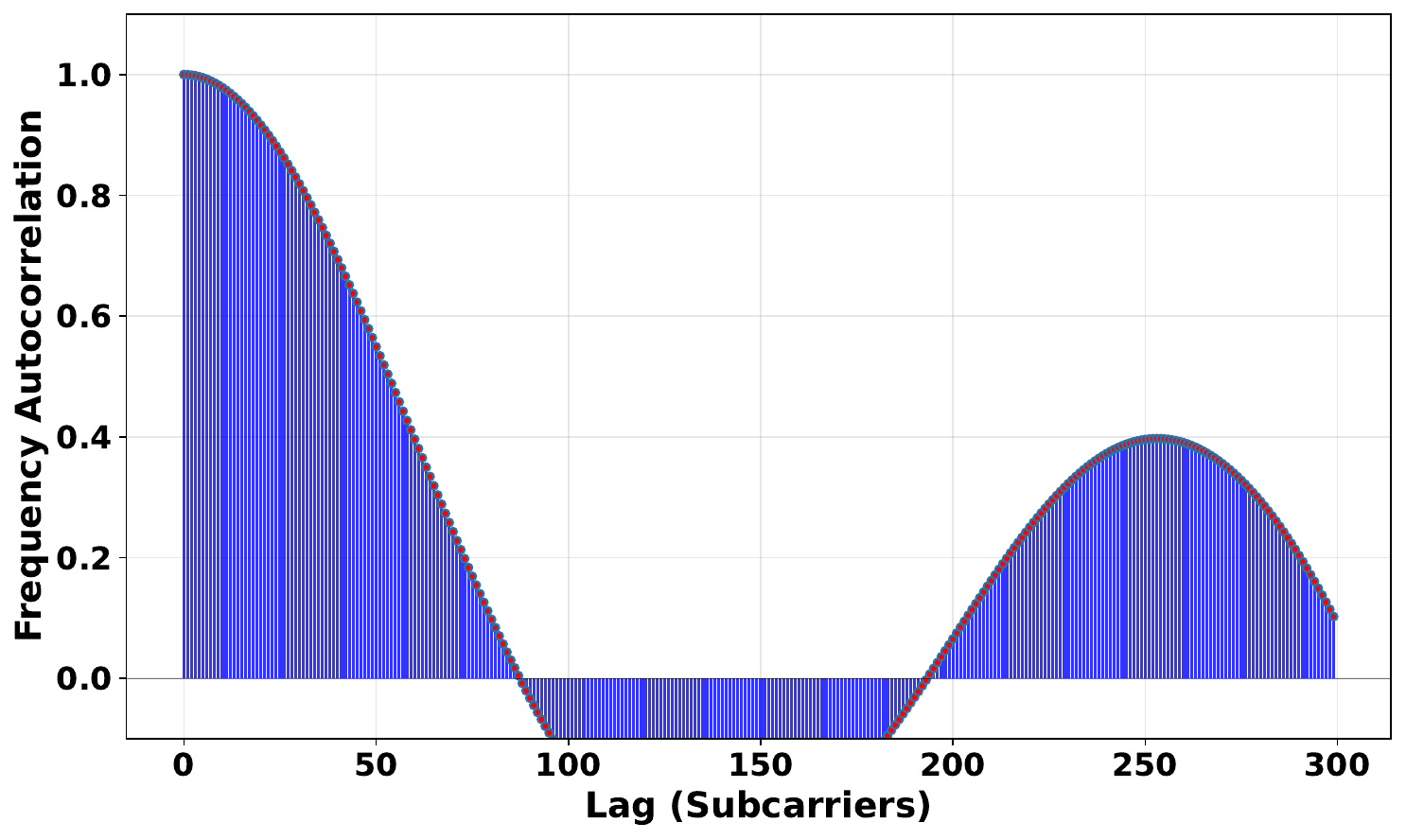}
    \label{fig:fdd_freq_acf_cm_c_ds_200ns_ms_1mps_gen_ds}
  }\
  \subfloat[300ns]{
    \includegraphics[width=0.45\linewidth]{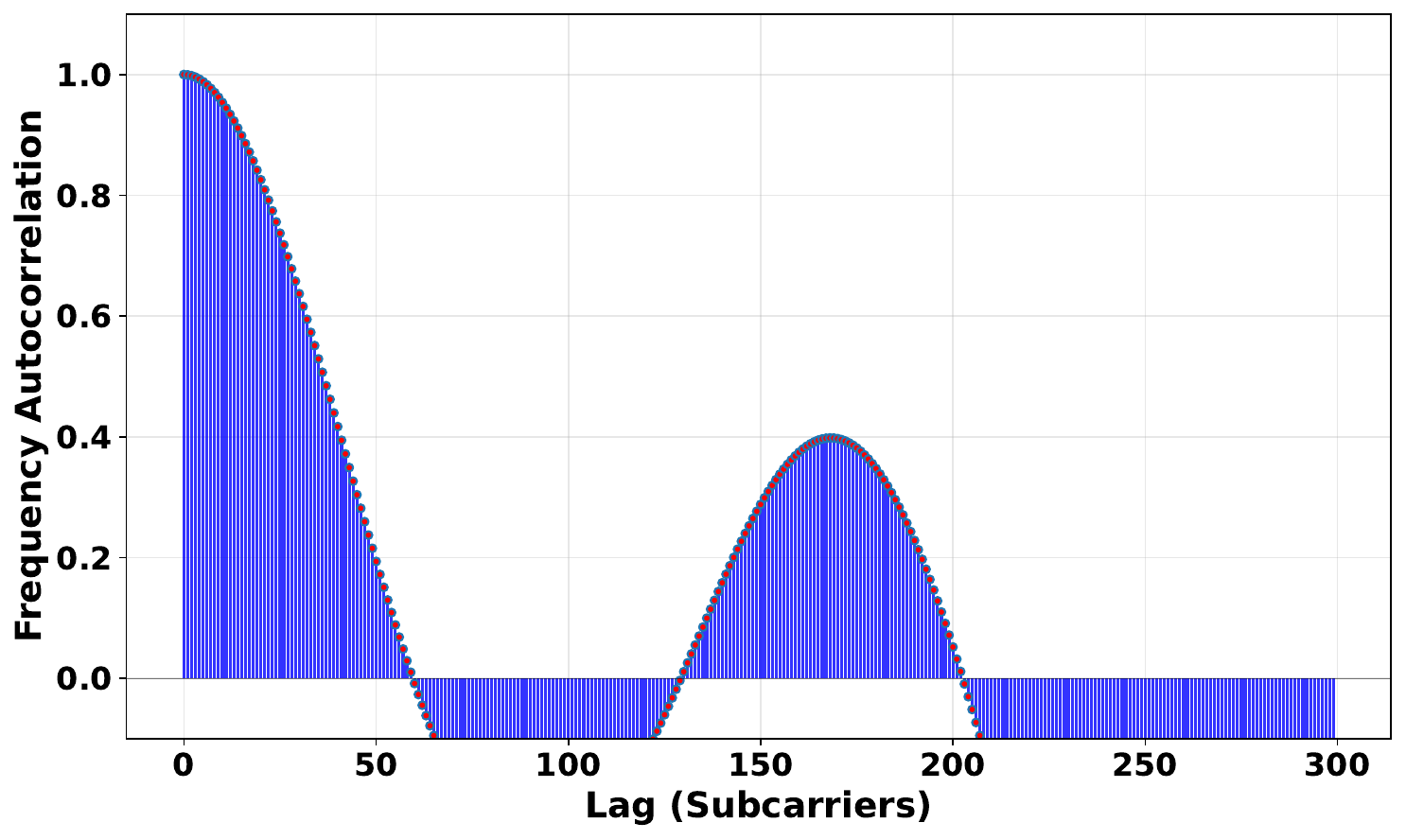}
    \label{fig:fdd_freq_acf_cm_c_ds_300ns_ms_1mps_gen_ds}
  }
  \subfloat[400ns]{
    \includegraphics[width=0.45\linewidth]{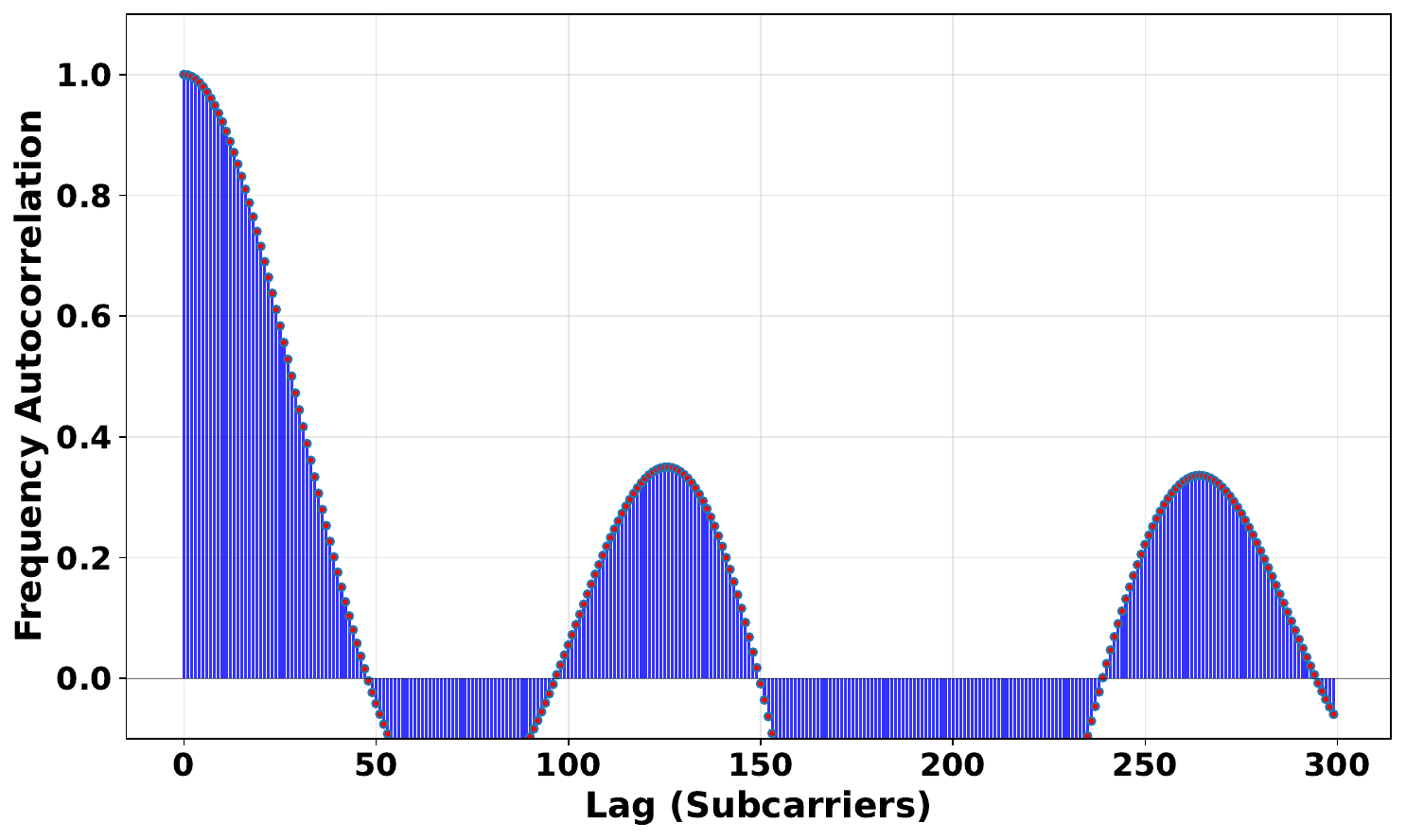}
    \label{fig:fdd_freq_acf_cm_c_ds_400ns_ms_1mps_gen_ds}
  }
  \caption{\textbf{Frequency ACF across different delay spreads. (FDD | CDL-C | 1m/s)}}
  \label{fig:acf-across-delay-spreads-fdd-freq}
\end{figure}

\begin{figure}[!ht]
  \centering
  \subfloat[30ns]{
    \includegraphics[width=0.45\linewidth]{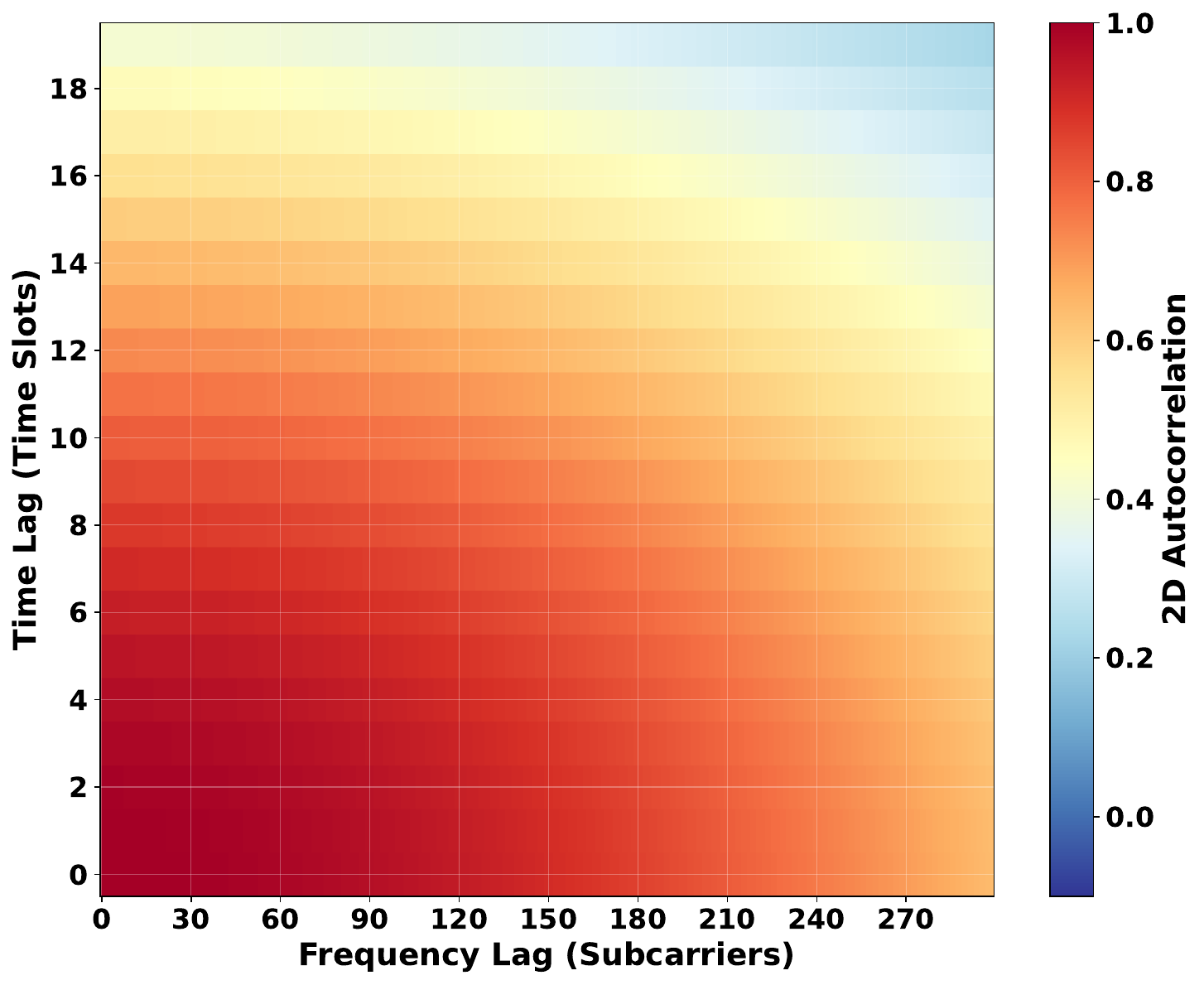}
    \label{fig:fdd_2d_acf_cm_c_ds_30ns_ms_1mps_gen_ds}
  }
  \subfloat[50ns]{
    \includegraphics[width=0.45\linewidth]{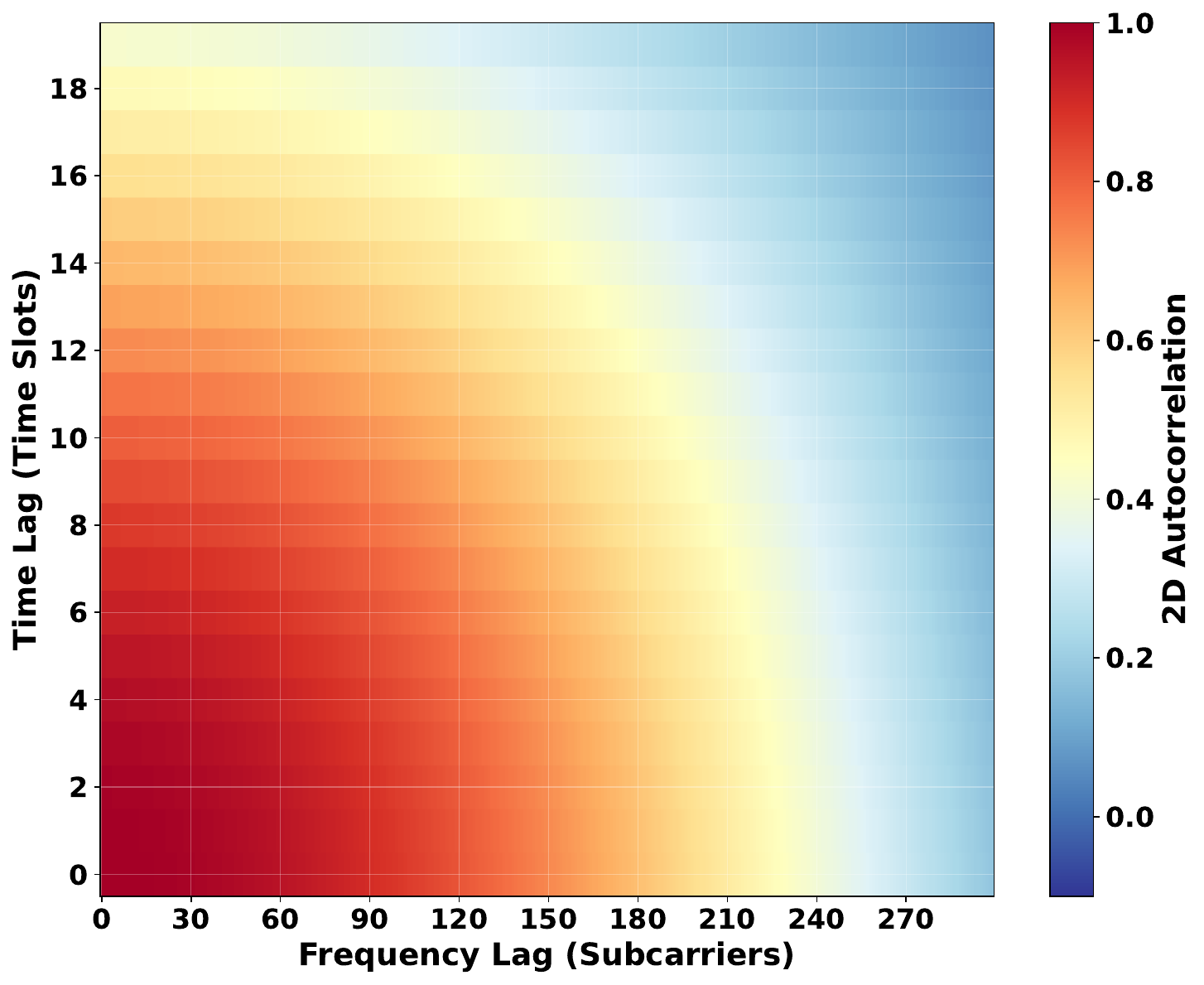}
    \label{fig:fdd_2d_acf_cm_c_ds_50ns_ms_1mps_gen_ds}
  }\
  \subfloat[100ns]{
    \includegraphics[width=0.45\linewidth]{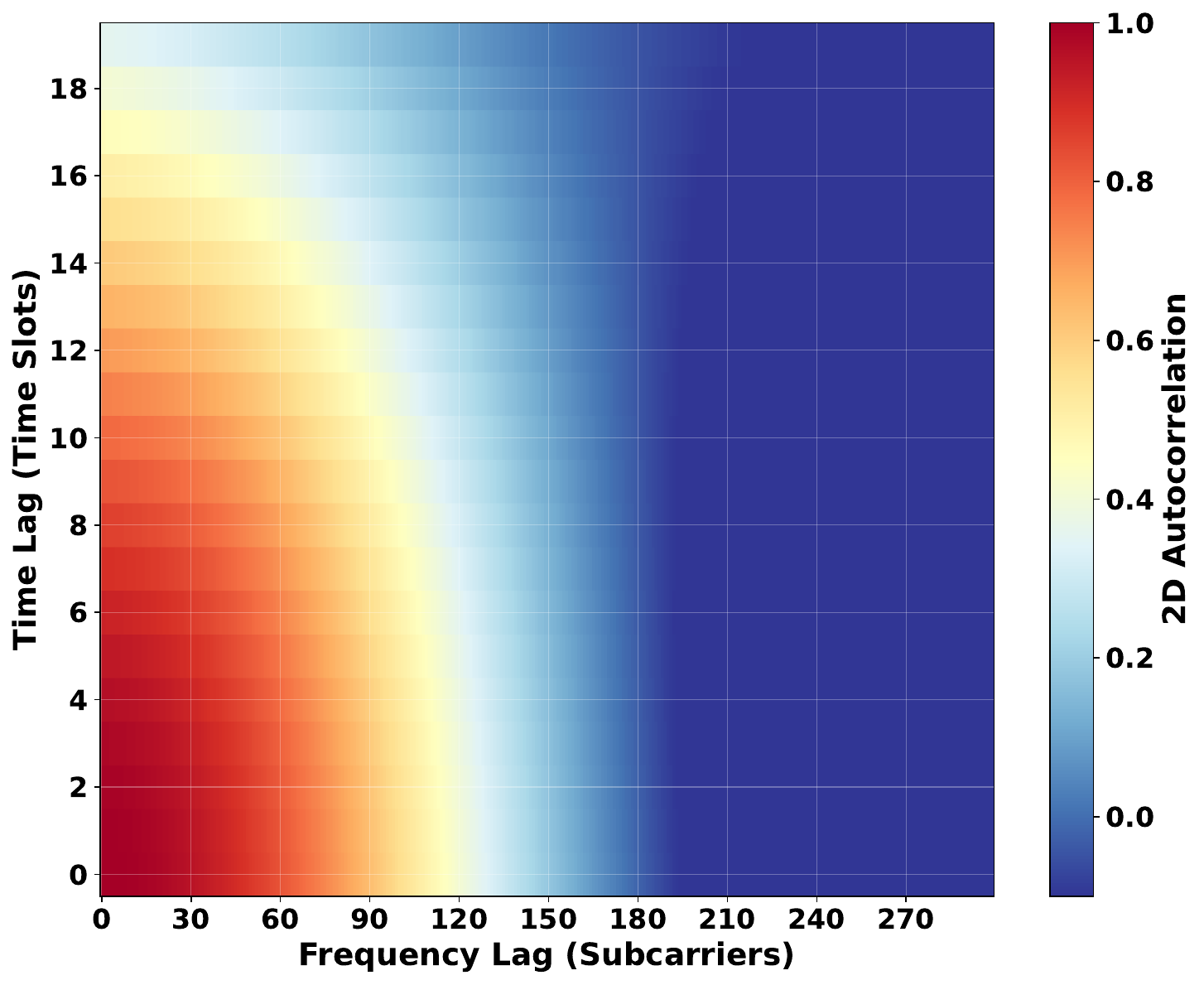}
    \label{fig:fdd_2d_acf_cm_c_ds_100ns_ms_1mps_gen_ds}
  }
  \subfloat[200ns]{
    \includegraphics[width=0.45\linewidth]{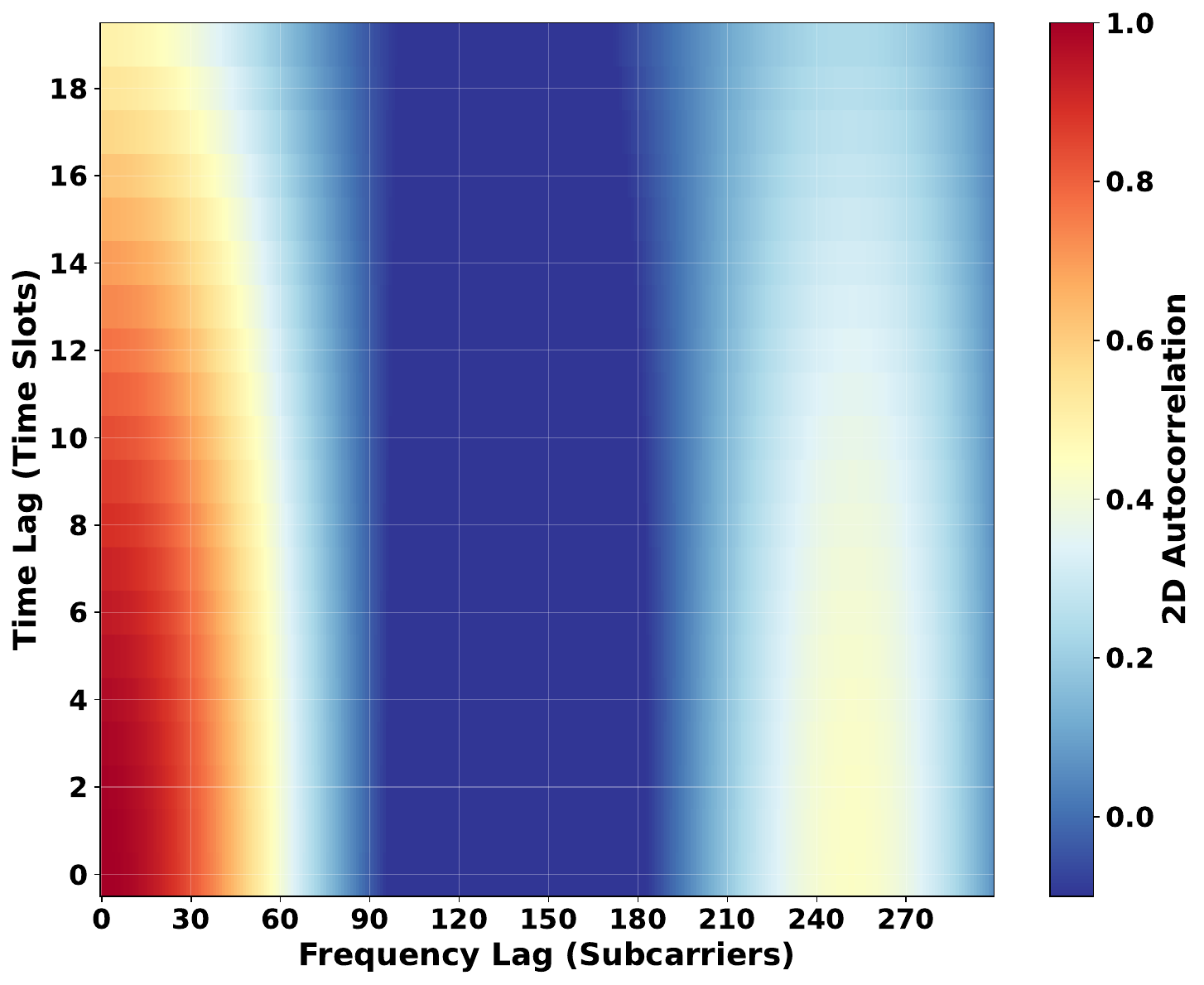}
    \label{fig:fdd_2d_acf_cm_c_ds_200ns_ms_1mps_gen_ds}
  }\
  \subfloat[300ns]{
    \includegraphics[width=0.45\linewidth]{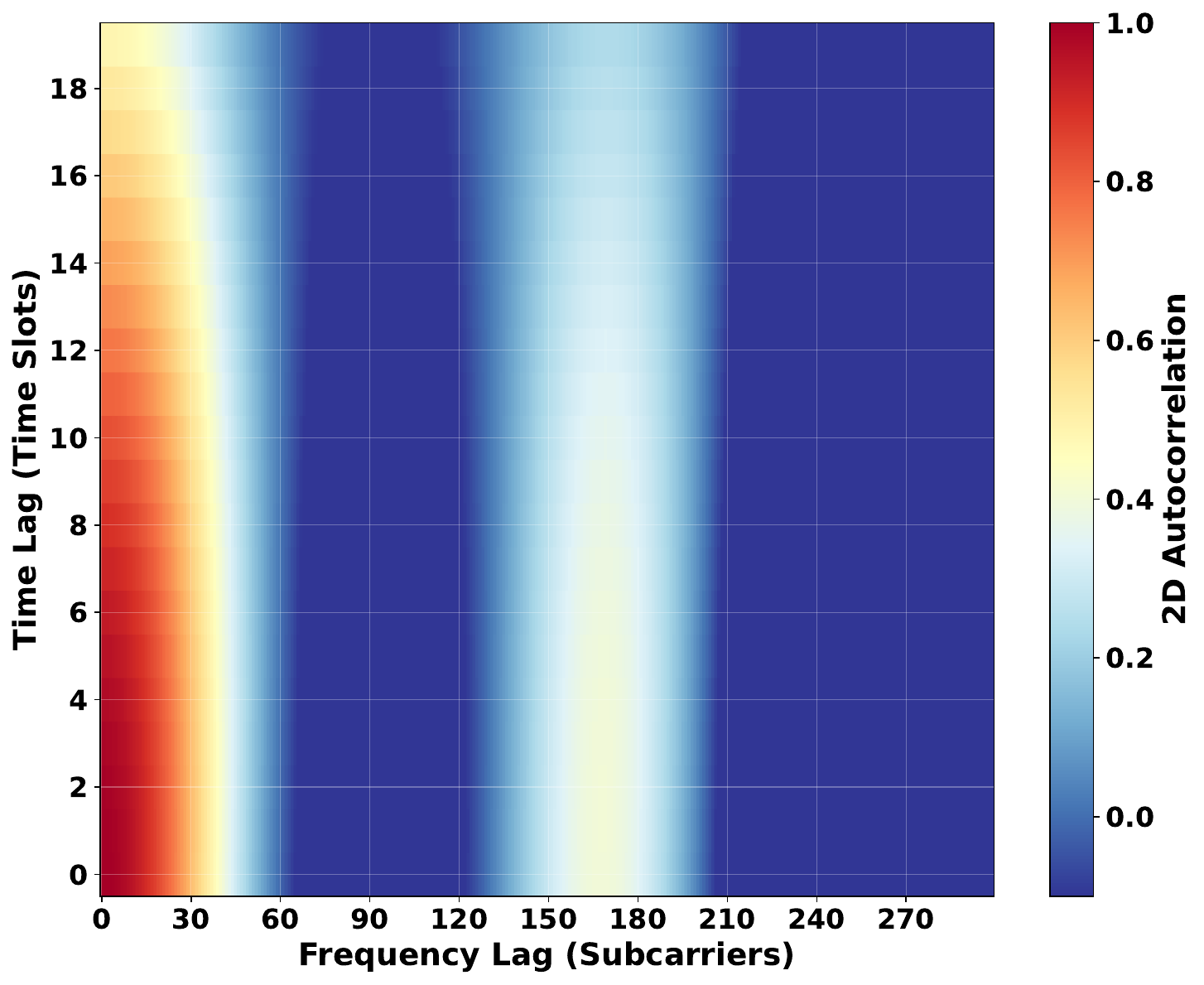}
    \label{fig:fdd_2d_acf_cm_c_ds_300ns_ms_1mps_gen_ds}
  }
  \subfloat[400ns]{
    \includegraphics[width=0.45\linewidth]{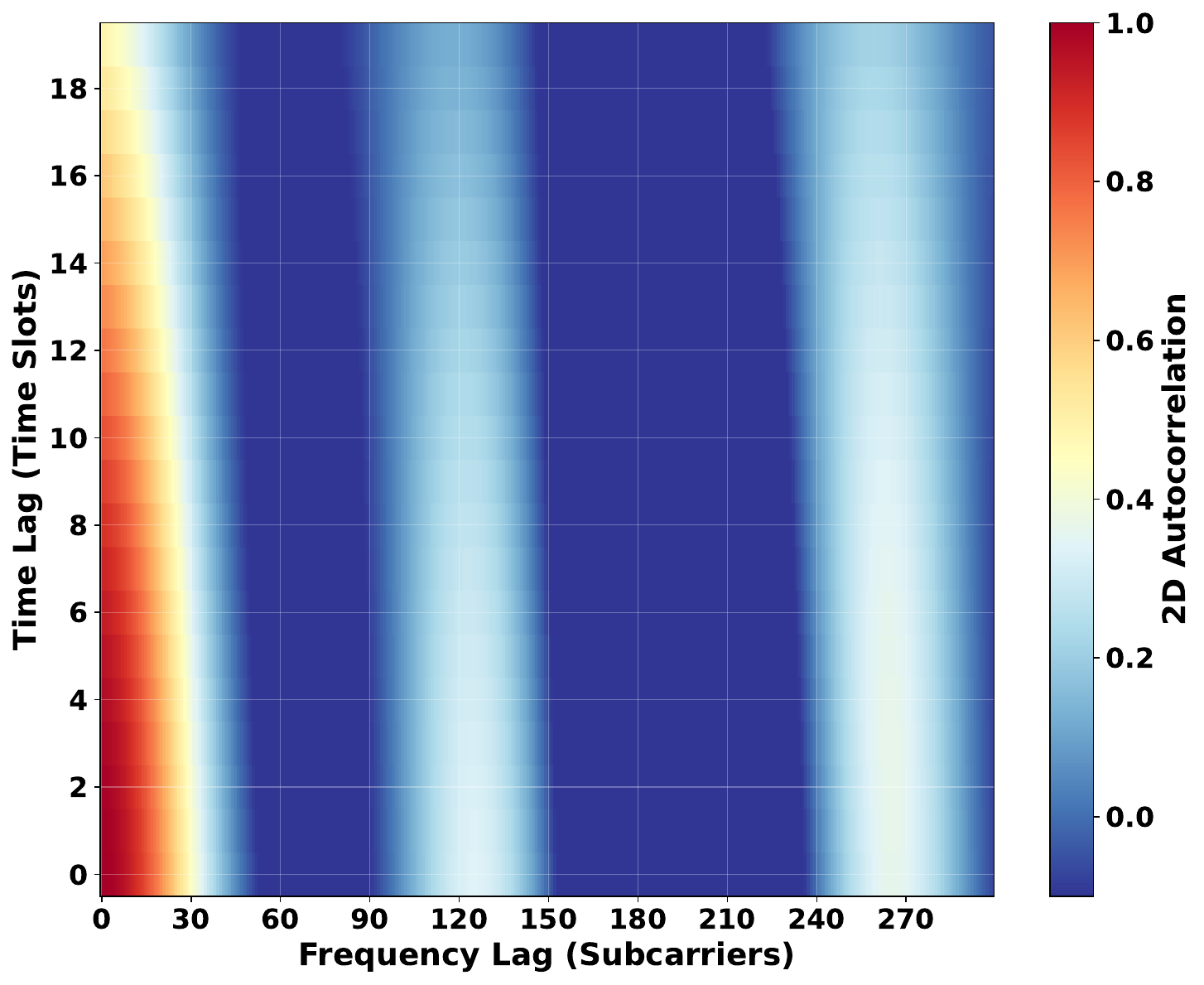}
    \label{fig:fdd_2d_acf_cm_c_ds_400ns_ms_1mps_gen_ds}
  }
  \caption{\textbf{2D Frequency ACF across different delay spreads. (TDD | CDL-C | 1m/s)}}
  \label{fig:acf-across-delay-spreads-tdd-freq-2d-cdl-c}
\end{figure}

\begin{figure}[!ht]
  \centering
  \subfloat[30ns]{
    \includegraphics[width=0.45\linewidth]{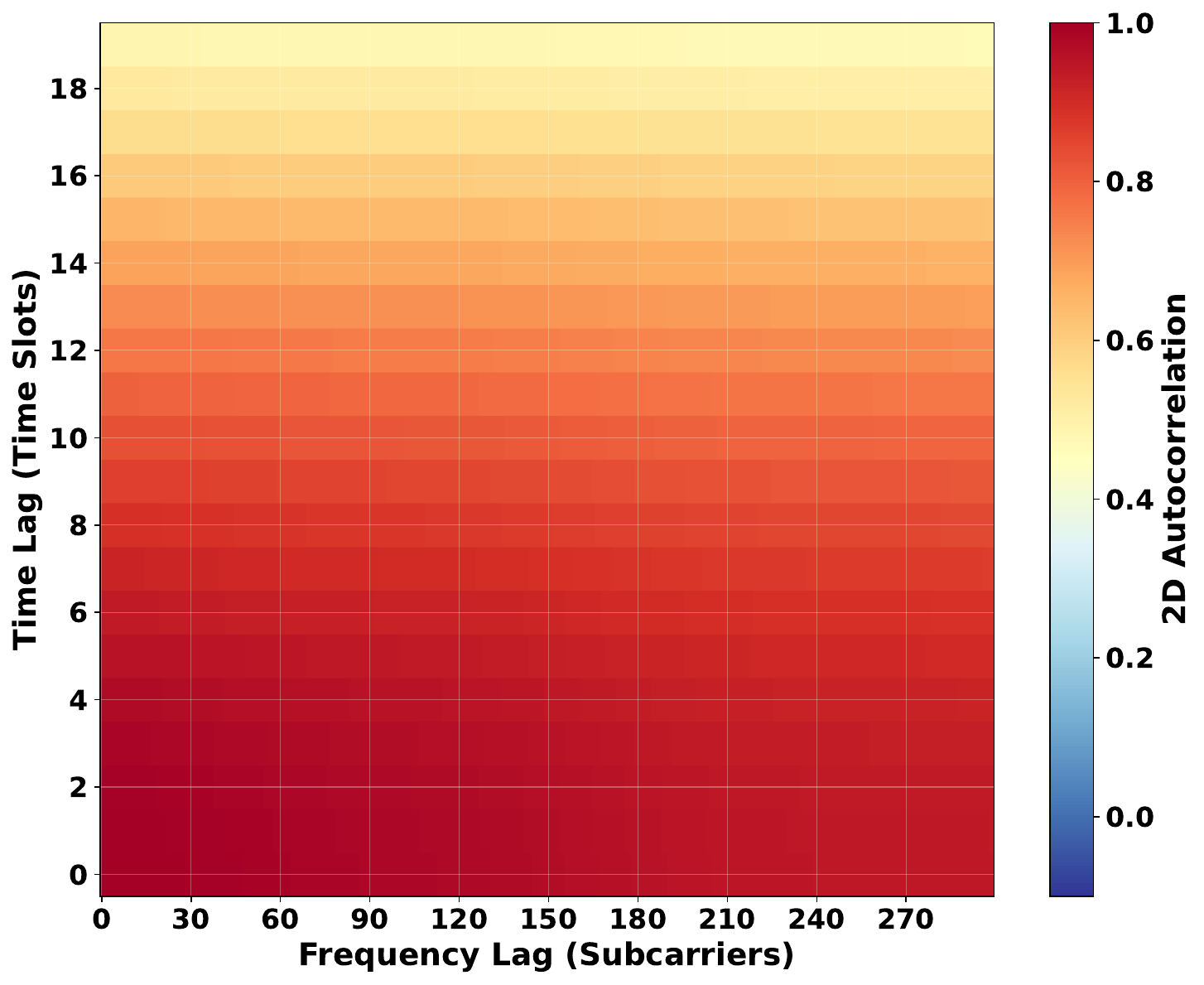}
    \label{fig:fdd_2d_acf_cm_d_ds_30ns_ms_1mps_gen_ds}
  }
  \subfloat[50ns]{
    \includegraphics[width=0.45\linewidth]{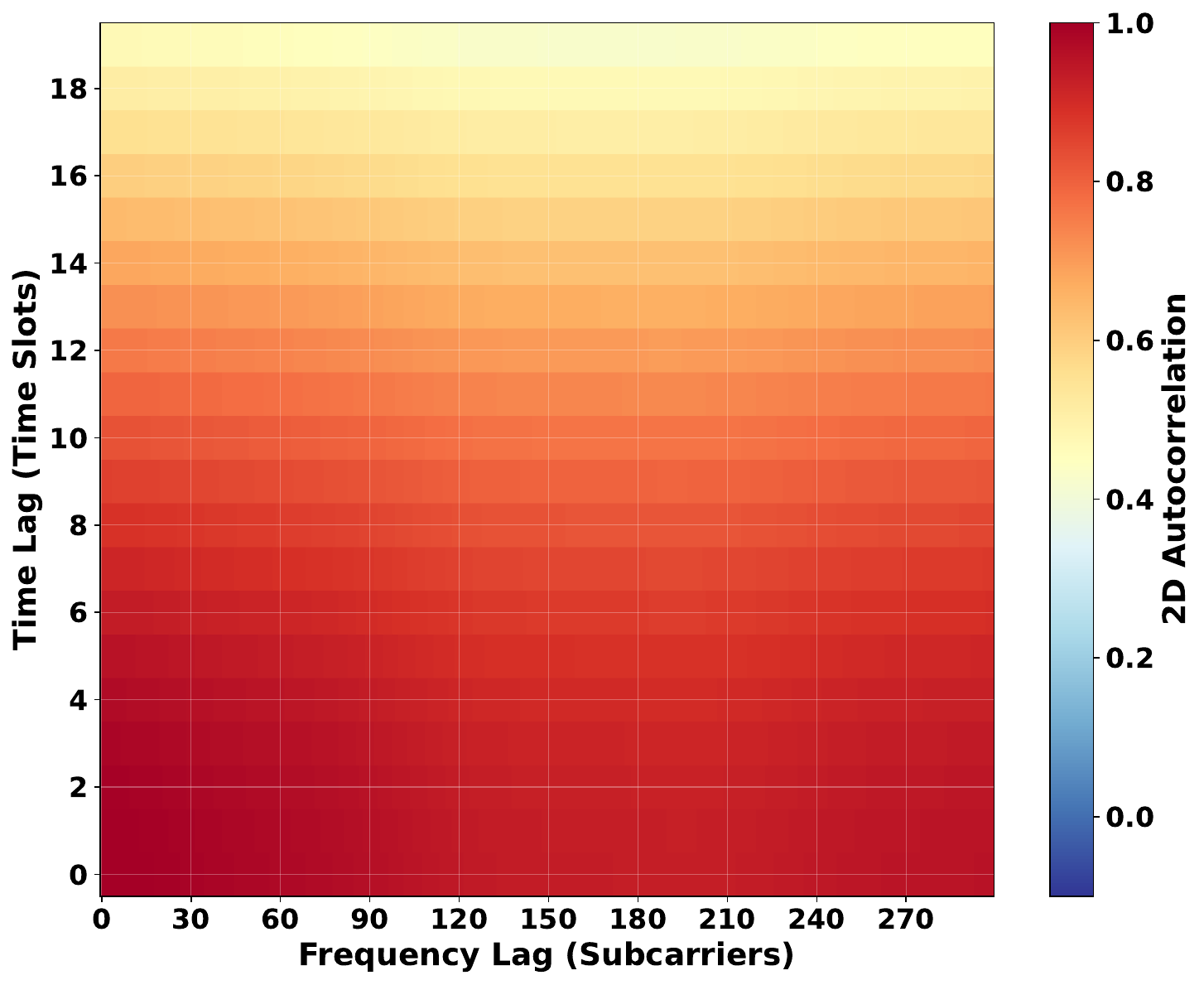}
    \label{fig:fdd_2d_acf_cm_d_ds_50ns_ms_1mps_gen_ds}
  }\
  \subfloat[100ns]{
    \includegraphics[width=0.45\linewidth]{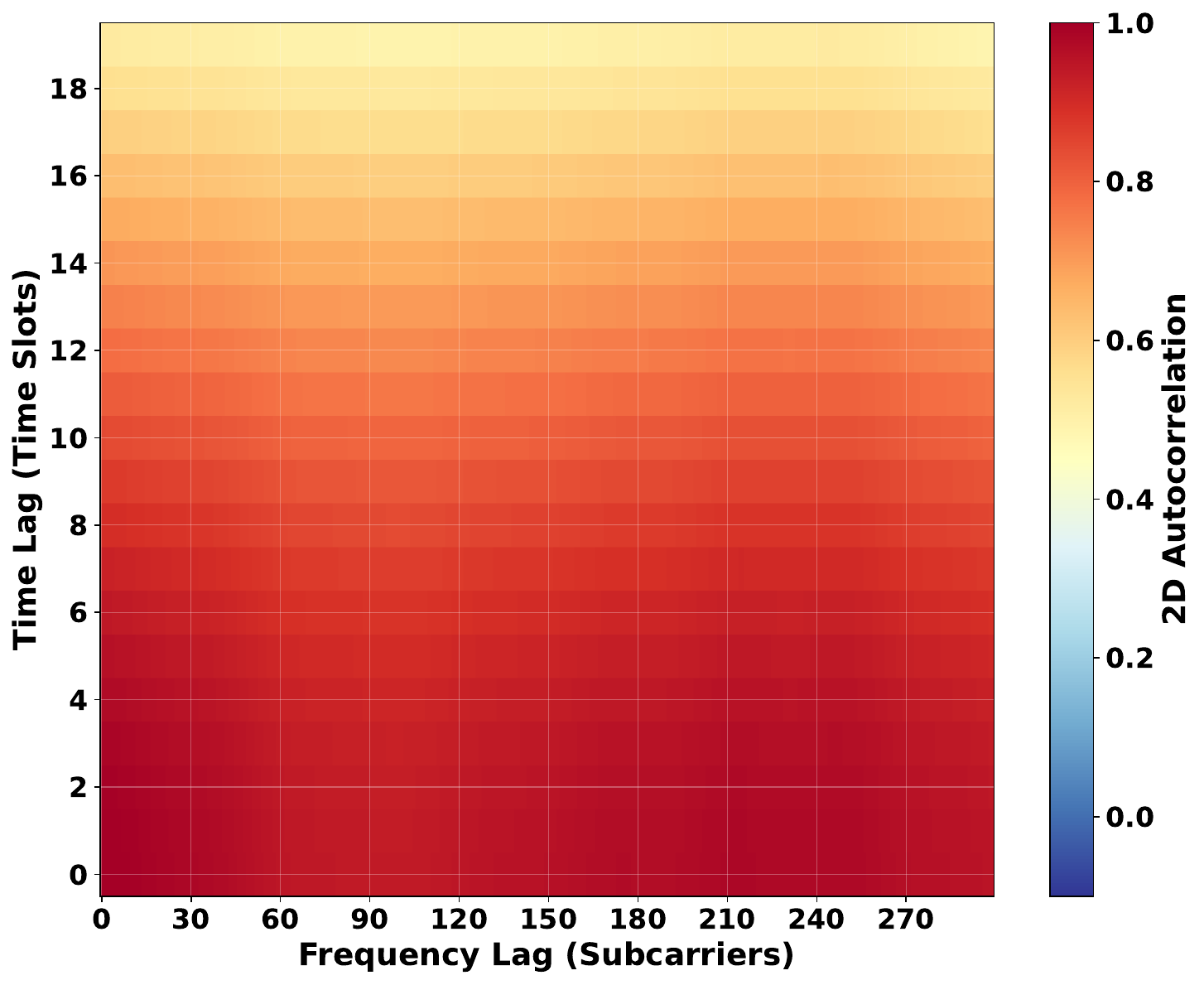}
    \label{fig:fdd_2d_acf_cm_d_ds_100ns_ms_1mps_gen_ds}
  }
  \subfloat[200ns]{
    \includegraphics[width=0.45\linewidth]{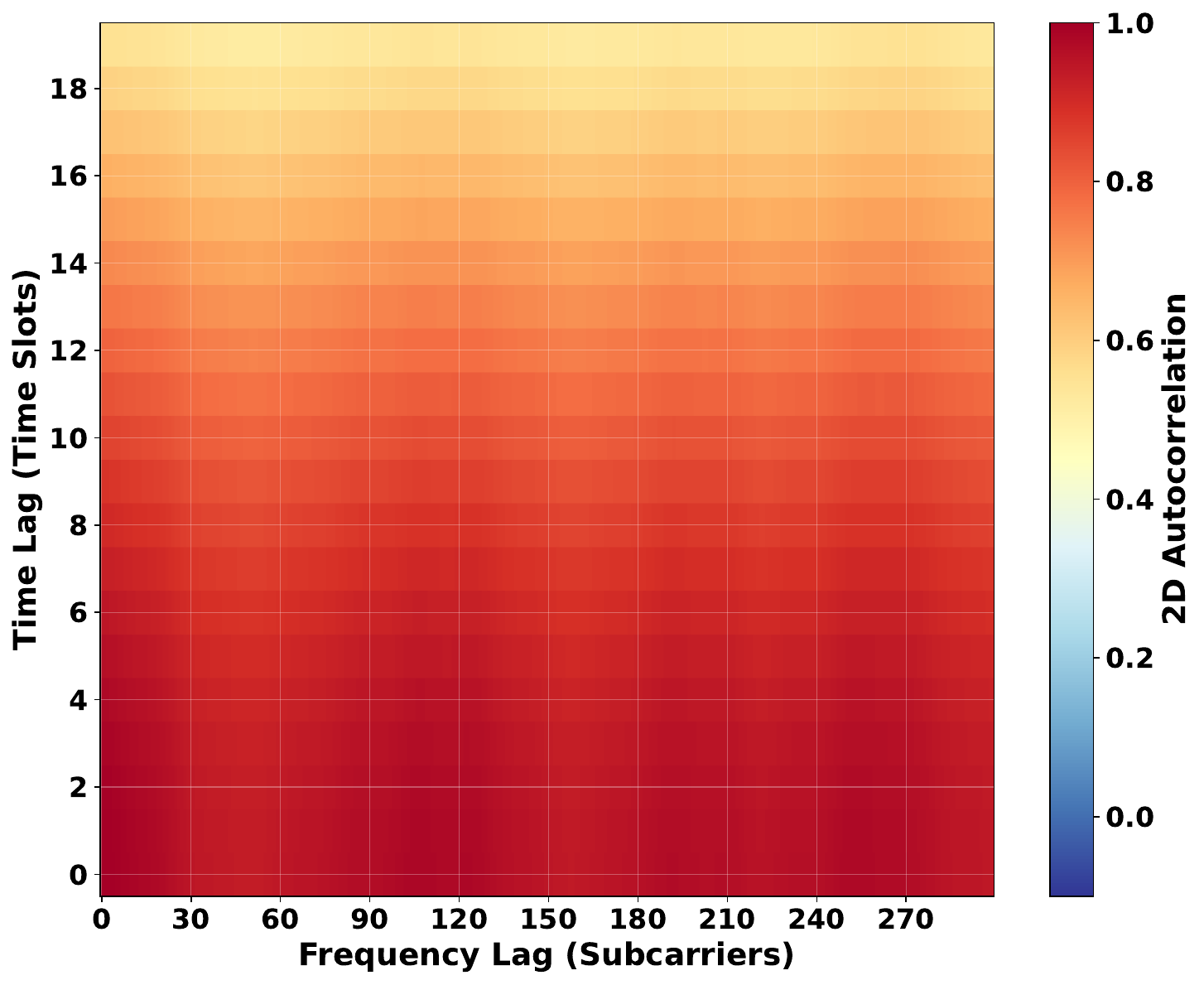}
    \label{fig:fdd_2d_acf_cm_d_ds_200ns_ms_1mps_gen_ds}
  }\
  \subfloat[300ns]{
    \includegraphics[width=0.45\linewidth]{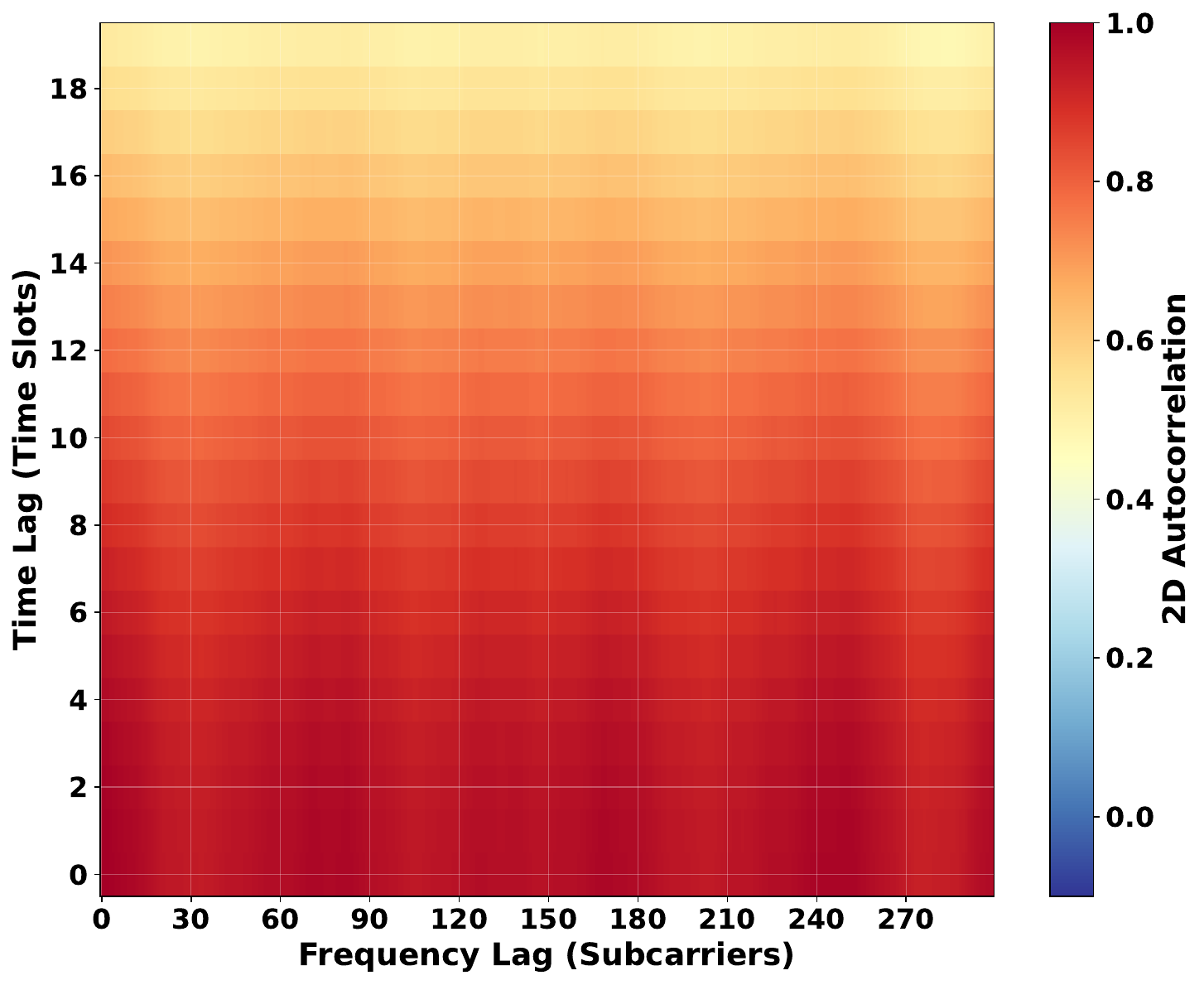}
    \label{fig:fdd_2d_acf_cm_d_ds_300ns_ms_1mps_gen_ds}
  }
  \subfloat[400ns]{
    \includegraphics[width=0.45\linewidth]{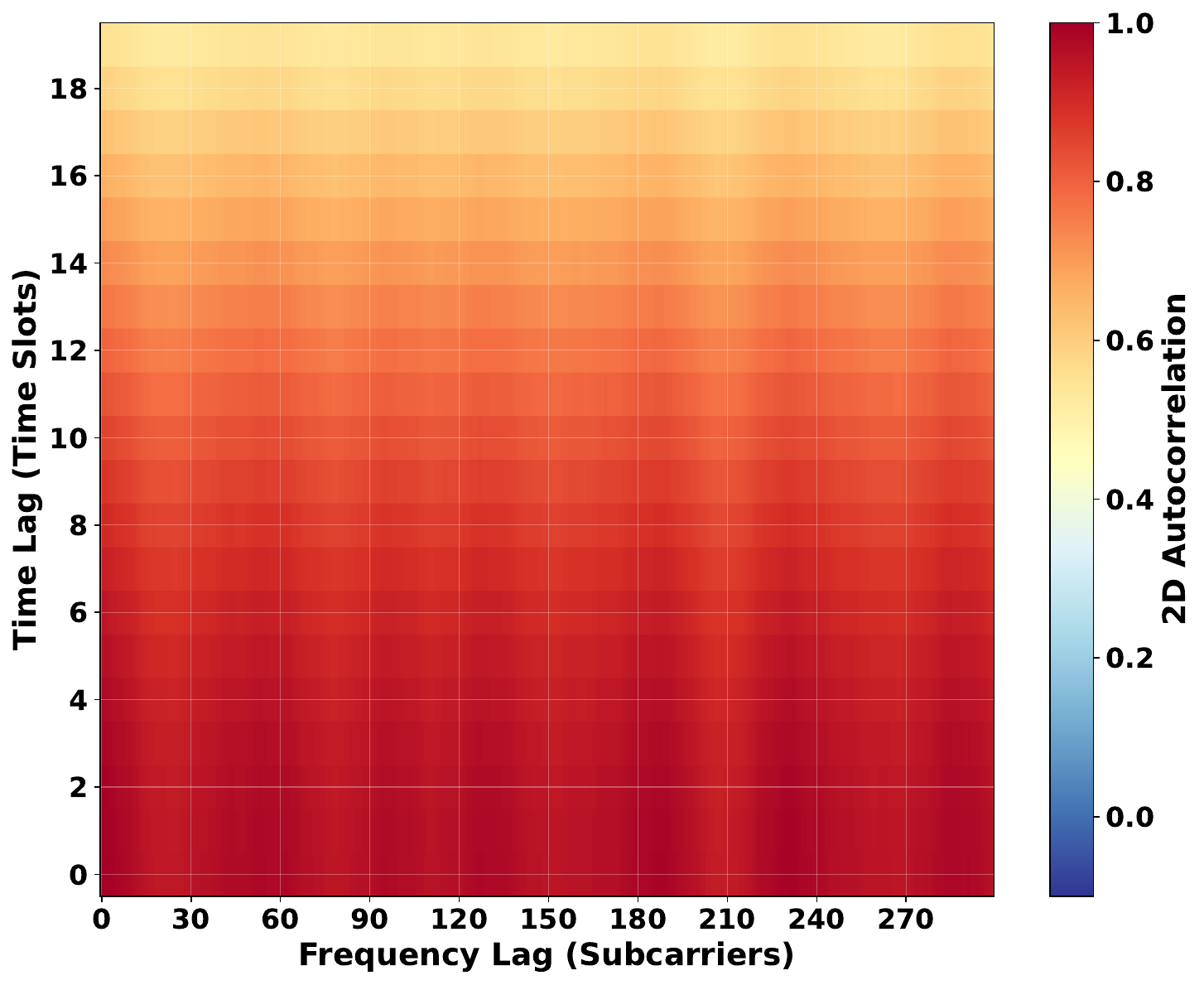}
    \label{fig:fdd_2d_acf_cm_d_ds_400ns_ms_1mps_gen_ds}
  }
  \caption{\textbf{2D Frequency ACF across different delay spreads. (TDD | CDL-D | 1m/s)}}
  \label{fig:acf-across-delay-spreads-tdd-freq-2d-cdl-d}
\end{figure}

Similar to the previous section, which reported temporal ACF across different user velocities, here the temporal-frequency ACF across different delay spreads is presented. Compared with the temporal ACF in Fig.~\ref{fig:acf-across-user-velocities-fdd}, the frequency ACF in Fig.~\ref{fig:acf-across-delay-spreads-fdd-freq} exhibits more significant variation across delay spreads. Furthermore, as illustrated in Fig.~\ref{fig:acf-across-delay-spreads-tdd-freq-2d-cdl-c} and Fig.~\ref{fig:acf-across-delay-spreads-tdd-freq-2d-cdl-d}, different channel models lead to distinct temporal-frequency ACF patterns across delay spreads. These results highlight that temporal-frequency ACF variation follows a more complex mechanism, jointly influenced by both the channel model and the delay spread.

\section{Additional Noise}
\label{sec:additive-noise}

In the following, the different types of additional noise are defined. For the accuracy of the definition, the element-wise noisy CSI is defined as follows:
\begin{equation}
  \begin{aligned}
  \ecsin & = \ecsi + \enoise \\ 
  \text{ where } & \ecsin = \noisyCSI^t[m, 1, k] \\
   \text{ and } & \response{\enoise = \noise^t[m, 1, k]},
  \label{eq:elementwise_noisy_csi}
\end{aligned}
\end{equation} 
with $m$ and $k$ denoting the BS antenna index and subcarrier index, $1$ denotes the receiver antenna index (the single omnidirectional receive antenna is considered in this work), respectively. The variable $t$ indicates the time index, and $\enoise$ represents the additional noise.

\subsection{Additive White Gaussian Noise (AWGN)}
\label{sec:additive-white-gaussian-noise}
AWGN is a common noise model in wireless communication systems. It describes the noise as a Gaussian random variable with zero mean and variance $\sigma^2$, yielding,
\begin{equation}
  \enoise \sim \response{\mathcal{CN} \left( 0, \sigma^2 \right)}.
  \label{eq:awgn}
\end{equation}
where $\sigma^2$ is the variance of the noise. The relationship between the SNR and the $\sigma^2$ is given by:
\begin{equation}
  \begin{aligned}
    \text{SNR} & = 10 \log_{10} \left( \frac{\| \CSI \|_F^2}{\sigma^2} \right) \\
    \sigma^2 & = \| \CSI \|_F^2 \cdot 10^{-\text{SNR} / 10}
  \end{aligned}
  \label{eq:snr-variance-of-awgn}
\end{equation}
The above explicit expression of the $\sigma^2$ is used to generate the AWGN noise with the target SNRs.

\subsection{Definition of Realistic Additional Noise}

\subsubsection{Phase Noise}
\label{sec:phase-noise}
Phase noise (Fig.~\ref{fig:phase-noise}) is pervasive in practical communication systems and is a crucial factor limiting the performance of high-speed communications, thus requiring model robustness to phase fluctuations \cite{rasekh-2021, mehrpouyan-2012}. The complex element CSI can be represented by gain and phase, namely,
\begin{equation}
    \ecsi = \vert \ecsi \vert \exp{(j\ethet)}.
    \label{eq:gain-and-phase-representation-of-the-element-csi}
\end{equation}
Accordingly, the Gaussian-like perturbation $\ethetnoise$ is introduced to the phase part of the element CSI $\ethet$, the element-wise phase noise is formulated as follows:
\begin{equation}
  \begin{aligned}
    \enoise & = \vert \ecsi \vert \left( \exp{(j \ethetn)} - \exp{(j \ethet)} \right) \\
    & = \vert \ecsi \vert \left( \exp{(j (\ethet + \ethetnoise))} - \exp{(j \ethet)} \right), \\
    & \text{where } \ethetnoise \sim \mathcal{N} \left( 0, \sigma^2 \right)
  \end{aligned}
  \label{eq:phase-noise}
\end{equation}
The resulting noisy CSI is given by:
\begin{equation}
  \ecsin = \vert \ecsi \vert \exp{(j \ethetn)}
  \label{eq:phase-noise-csi}
\end{equation}

\subsubsection{Burst Noise}
\label{sec:burst-noise}
To better mimic practical channel conditions, burst noise (Fig.~\ref{fig:burst-noise}) is introduced to simulate sudden spike-like disturbances that may result from abrupt environmental changes or unexpected obstacles between the BS and UE \cite{li-2004-performance-evaluation, zhou-2021-uplink-channel-estimation}. Currently, burst noise is modeled as a bell-shaped perturbation spanning $\lburst$ consecutive slots. The amplitude of this bell shape is $\aburst$ and the probability of burst noise occurring in any given slot is characterized by the Bernoulli trial with burst probability $\pburst$. One additional limitation is that in a single historical CSI input ($L$ packets), there is at most one burst noise.

Accordingly, the starting time index of the burst noise is formulated as a truncated geometric distribution, yielding,
\begin{equation}
  \mathbb{P} (\startburst = t) = (1 - \pburst)^{t-1} \pburst, \quad t = 1, 2, \ldots, L
  \label{eq:starting-time-index-of-burst-noise}
\end{equation}
The bell-shaped perturbation is formulated as follows:
\begin{equation}
    \begin{aligned}
      \ebell & = \aburst \, g(t - \startburst) \, o (t, \startburst) \\
      \text{where } \\
      o (t, \startburst) & = \mathbf{1} \left\{ 0 \leq t - \startburst \leq \min \{ L, \lburst -1 \} \right\} \\
      g(\tau) & = \exp \left( - \frac{(\tau-c)^2}{2} \right) \\
      c & = \frac{\lburst-1}{2}, \, \tau \in [0, \min \{ L, \lburst -1 \}]
    \end{aligned}
    \label{eq:burst-noise-bell}
\end{equation}  

\begin{equation}
  \enoise = \ebell \cdot \eepsilon, \quad \text{where } \eepsilon \sim \mathcal{N} \left( 0, 1 \right)
  \label{eq:burst-noise-csi}
\end{equation}

\subsubsection{Packet drop Noise} 
\label{sec:packet-drop-noise}
\begin{figure}[!ht]
  \centering
  \subfloat[Phase noise]{
    \includegraphics[width=0.7\linewidth]{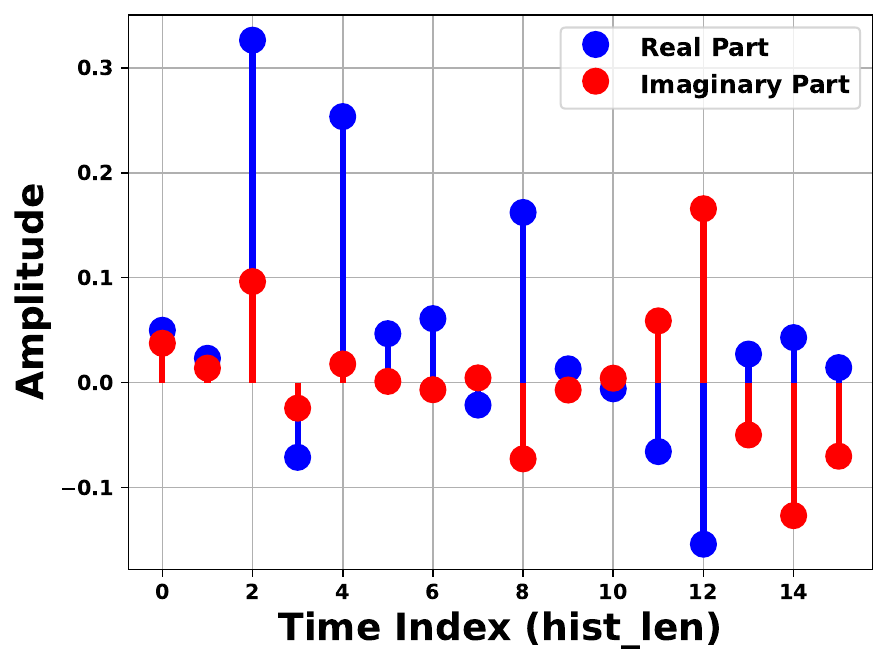}
    \label{fig:phase-noise}
  }\
  \subfloat[Burst noise]{
    \includegraphics[width=0.7\linewidth]{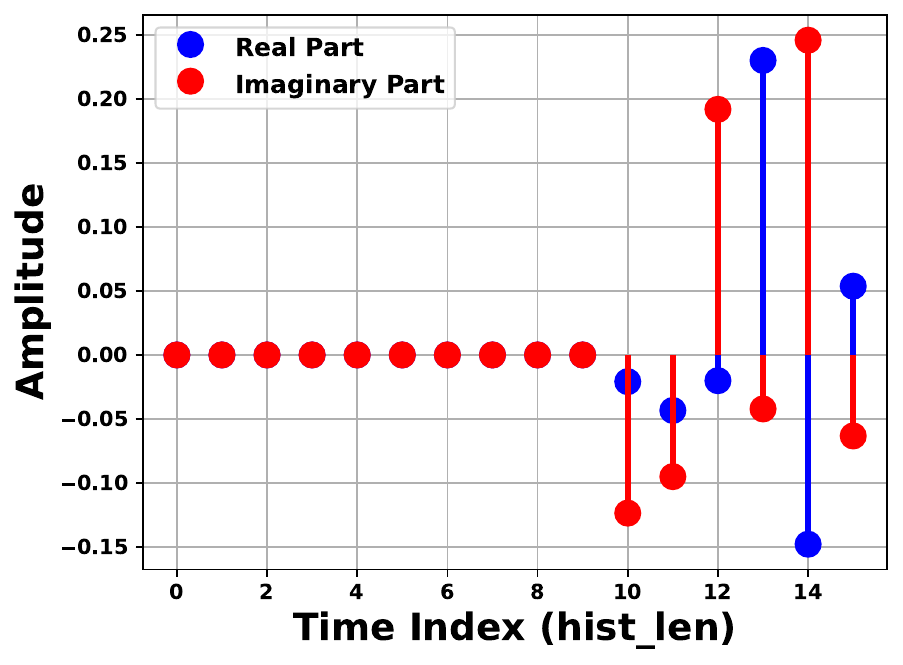}
    \label{fig:burst-noise}
  }\
  \subfloat[Packet drop noise]{
    \includegraphics[width=0.7\linewidth]{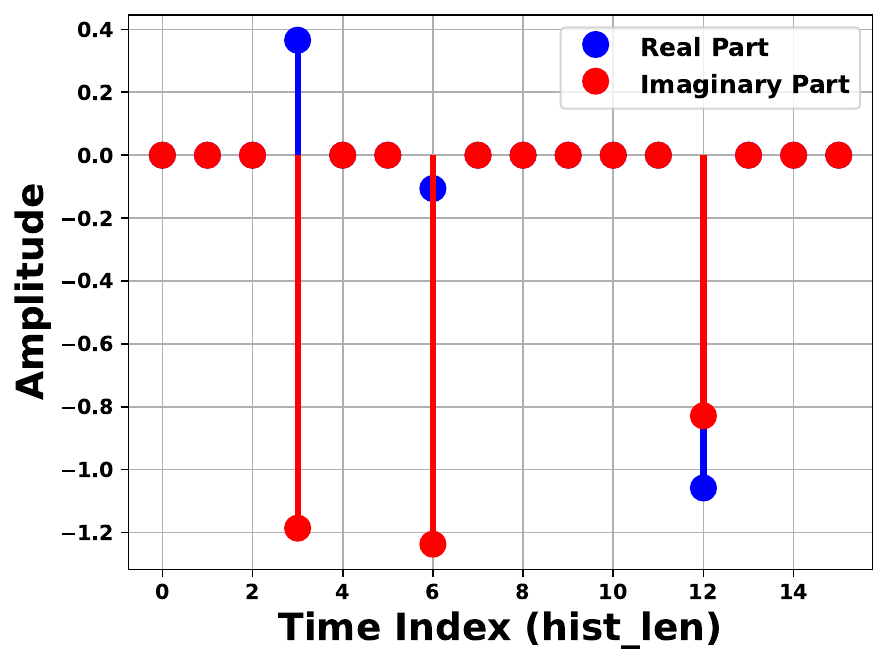}
    \label{fig:packet-drop-noise}
  }
  \caption{\textbf{Visualization of realistic additive noises.} \textnormal{For the first dataset sample, we plot the real and imaginary parts of the injected noise on antenna index $0$ and subcarrier index $1$ for each additional noise type (phase, burst, and packet drop).}}
  \label{fig:visualization-realistic-additional-noises}
\end{figure}

Packet drop noise (Fig.~\ref{fig:packet-drop-noise}) refers to the random omission of CSI packets \cite{morato2024packet}. For each time step $t$, whether a packet is dropped is modeled as a realization of a Bernoulli random variable with parameter $p_d$, i.e., $d^t \sim \text{Bernoulli}(p_d)$. The packet drop noise is then defined as:
\begin{equation}
  \enoise = 0 - \ecsi \cdot d^t.
  \label{eq:packet_drop_noise_elementwise}
\end{equation}
This implies that if $d^t = 1$, all CSI elements at time $t$ are dropped; otherwise, the CSI remains unchanged. Consequently, the resulting noisy CSI is given by:
\begin{equation}
  \ecsin = 
  \begin{cases}
    \ecsi, & \text{if } d^t = 0 \\
    0, & \text{if } d^t = 1
  \end{cases}.
  \label{eq:resulting_noisy_csi_packet_drop}
\end{equation}

\subsection{Experiment Details}

\paragraph{Noise calibration and SNR definition.}
For \emph{phase noise}, the standard deviation $\sigma$ of the Gaussian perturbation is controlled; larger $\sigma$ yields larger phase excursions. For \emph{burst noise}, both the pulse amplitude $\aburst$ and the occurrence probability $\pburst$ are set proportional to a controllable \emph{noise-degree} parameter $nd$ (i.e., $\aburst\propto nd$ and $\pburst\propto nd$), so higher $nd$ produces stronger and more frequent bursts.

The simulation experiments are conducted to empirically calibrate $(\sigma, nd)$ against the resulting SNR. Throughout, SNR is defined as the signal-to-noise power ratio as follows:
\begin{equation}
  \text{SNR} = 10 \log_{10} \left( \frac{\| \CSI \|_F^2}{\sum_t \sum_m \sum_k \vert \ecsi \vert^2} \right).
  \label{eq:snr-of-phase-noise}
\end{equation}
Then the $(\sigma, nd)$ are selected to realize matched SNR targets $\{10,15,20,25\}$\,dB for fair comparison. Very low SNRs (e.g., $0$–$5$\,dB) would require extreme parameter values (e.g., unusually large phase excursions) and are uncommon in practice (e.g., 0.59 radians on the phase perturbation corresponds to the SNR equals to 5\,dB). For \emph{packet-drop} noise, we use per-step Bernoulli erasures with drop probabilities $\{0.01,0.02,\ldots,0.10\}$ to reflect realistic operating conditions.

\paragraph{Imputation for the packet drop noise.}

Following \cite{zhao2024mining}, packet drops can be detected by monitoring inter-packet intervals of consecutive CSI frames; accordingly, indices of dropped packets are treated as known. In this work, missing CSI samples are handled via simple \emph{imputation}: each missing CSI sample is replaced by the last available(/observed) sample. To maintain a simple protocol and isolate the intrinsic robustness of the prediction models, more sophisticated imputers are intentionally omitted.

\section{Model-based Baselines}
\label{sec:appendix-model-based-baselines}

\subsection{Overview}
\response{To clarify the experimental setting, we highlight the following points regarding the model-based baselines:
\begin{itemize}
  \item \textbf{TDD/FDD tasks:} AR models temporal evolution within the same band, and PAD was originally developed to address TDD channel aging via forward extrapolation from delayed same-band CSI. Accordingly, following their original formulations and assumptions, AR and PAD are evaluated only on the TDD task. In contrast, Wiener does not assume a specific duplexing structure and is evaluated on both TDD and FDD tasks.
  \item \textbf{Subcarrier-wise fitting:} For AR and Wiener, the linear model is fitted independently for each subcarrier using the channel vector across antennas. This design preserves spatial dependence across the antenna array while keeping the required matrix computations tractable. A full CSI-level formulation would lead to extremely large coefficient or covariance matrices, making both storage and matrix inversion prohibitively expensive in practice. For example, for Wiener with a 16-step history, the input dimension is $32\times16\times300=153{,}600$, which would require forming and solving a $153{,}600\times153{,}600$ covariance system.
  \item \textbf{PAD safe fallback mechanism:} The PAD implementation is sensitive to structured, non-Gaussian noise and may occasionally exhibit catastrophic numerical failures (NMSE $>10^{30}$). To ensure a meaningful comparison, we incorporate a simple safe fallback mechanism into the PAD implementation; details are provided in Appendix~\ref{sec:appendix-pad-safe-fallback}.
\end{itemize}}

\subsection{Modified PAD Baseline with Regularization and Safe Fallback}
\label{sec:appendix-pad-safe-fallback}

\response{The PAD baseline in this paper is implemented with a safe fallback mechanism to improve numerical robustness under burst noise and packet-drop noise. The following notation is from \cite{yin-2020}:}

\response{Let $g_{u,n}(t_\ell)$ denote a non-negligible angular-delay coefficient at receive antenna index $u$ and angular-delay position $n$. The Prony coefficients are estimated as:}
\begin{equation}
  \response{\hat{\mathbf p}_{u,n} = -\mathbf G_{u,n}^{\dagger}\mathbf b_{u,n},}
  \label{eq:prony-coefficients}
\end{equation}
\response{where $\mathbf G_{u,n}$ is the Hankel matrix and $\mathbf b_{u,n}$ is the shifted target vector. The future coefficient is then generated recursively as:}
\begin{equation}
  \response{\hat g_{u,n}(t_{L+1}) = -\mathbf b^{(L)}_{u,n}\hat{\mathbf p}_{u,n}.}
  \label{eq:future-coefficients}
\end{equation}
\response{The fragile part of PAD lies in the pseudoinverse $\mathbf G_{u,n}^{\dagger}$. In particular,}
\begin{equation}
  \response{\|\hat{\mathbf p}_{u,n}\|_2
\le
\|\mathbf G_{u,n}^{\dagger}\|_2\,\|\mathbf b_{u,n}\|_2
=
\frac{\|\mathbf b_{u,n}\|_2}{\sigma_{\min}^+(\mathbf G_{u,n})},}
  \label{eq:fragile-part-of-pad}
\end{equation}
\response{where $\sigma_{\min}(\mathbf G_{u,n})$ is the smallest singular value of $\mathbf G_{u,n}$. Therefore, when $\mathbf G_{u,n}$ is ill-conditioned, small perturbations in the observed CSI may produce very large Prony coefficients, which can lead to an unstable recursive rollout.}

\response{This issue is particularly severe under burst noise and packet-drop noise, as defined in Appendix~\ref{sec:additive-noise}. Burst noise introduces short, high-energy corruptions over consecutive slots. These corruptions act as strong outliers in the least-squares fit: a few corrupted slots may dominate the objective and contaminate multiple rows of $\mathbf G_{u,n}$. A true packet drop erases an entire CSI packet, which directly breaks the smooth exponential evolution assumed by the Prony model. Even after the simple last-observation imputation used in the current benchmark, the resulting historical CSI contains repeated or piecewise-constant segments. Such segments do not follow the exponential dynamics assumed by PAD. In both cases, the estimated Prony coefficients may become abnormally large, and the recursive prediction may diverge. In preliminary experiments, standard PAD occasionally produced catastrophic failures under these two noise types, with NMSE exceeding $10^{30}$.}

\response{To avoid such numerical blow-ups, the current implementation first replaces the raw pseudoinverse with a ridge-regularized solve:}
\begin{equation}
  \response{\hat{\mathbf p}^{\mathrm{reg}}_{u,n}
  =
  -\left(\mathbf G_{u,n}^{\mathrm H}\mathbf G_{u,n} + \lambda_{u,n}\mathbf I\right)^{-1}
  \mathbf G_{u,n}^{\mathrm H}\mathbf b_{u,n},}
  \label{eq:ridge-prony-coefficients}
\end{equation}
\response{where $\mathbf G_{u,n}^{\mathrm H}$ denotes the Hermitian transpose and $\lambda_{u,n}>0$ is an adaptive damping factor proportional to the average diagonal energy of $\mathbf G_{u,n}^{\mathrm H}\mathbf G_{u,n}$. This damping reduces sensitivity to near-singular Hankel matrices and suppresses excessively large Prony coefficients.}

\response{After the recursive rollout, the PAD prediction is accepted only if it satisfies a simple numerical safety check. Denote the PAD output by $\hat{\mathbf H}^{P}_{\mathrm{PAD}}$ and the observed history by $\mathbf H^{L}_{\mathrm{hist}}$. The prediction is declared safe when:}
\begin{equation}
  \begin{aligned}
    & \response{ \max_{u,n} |\hat{\mathbf H}^{P}_{\mathrm{PAD}}(u,n)| \text{ is finite}} \\
    \response{ \text{and }} & \response{\max_{u,n} |\hat{\mathbf H}^{P}_{\mathrm{PAD}}(u,n)|
    \le \tau \max_{u,n} |\mathbf H^{L}_{\mathrm{hist}}(u,n)|.}
  \end{aligned}
  \label{eq:pad-safe-condition}
\end{equation}
\response{where $\tau$ is a fixed threshold. Otherwise, the method falls back to the NP predictor, which repeats the last observed CSI slot:}
\begin{equation}
  \response{\hat{\mathbf H}_{\mathrm{NP}}(t_{L+r}) = \mathbf H(t_L),}
  \qquad r=1,\ldots,P.
  \label{eq:np-repeat-last}
\end{equation}
\response{Accordingly, the final prediction is:}
\begin{equation}
  \response{\hat{\mathbf H}^{P}_{\mathrm{PAD\text{-}safe}}}
\response{=\begin{cases}
\hat{\mathbf H}^{P}_{\mathrm{PAD}}, & \text{if safe},\\
\hat{\mathbf H}^{P}_{\mathrm{NP}}, & \text{otherwise}.
\end{cases}}
  \label{eq:safe-fallback-pad}
\end{equation}
\response{This safeguard is introduced only for the burst-noise and packet-drop settings. For AWGN and phase noise, PAD remains numerically stable because these perturbations better preserve the smooth exponential dynamics assumed by the Prony model; thus, the reported result is the standard PAD prediction without fallback. The purpose of this modification is not to claim an optimal redesign of PAD, but to make the comparison under burst and packet-drop corruption more meaningful. The evaluation of PAD under such structured, non-Gaussian disturbances remains largely underexplored, and this simple safeguard is adopted to prevent numerical instability from dominating the baseline comparison.}

\section{Training Configurations}
\label{sec:appendix-training-configuration}

Table \ref{tab:hparam-space} outlines the defined hyperparameter search space and the trainer settings used with the Optuna framework for automated tuning. The optimizer, scheduler, and training settings are shared across models, while architecture-specific hyperparameters are listed separately.

Each model undergoes one round of hyperparameter tuning for both the TDD and FDD datasets, as the distinct characteristics of the two duplexing modes demand separate configurations. The subcarrier-wise ACL layer in \Model{} is enabled only under the FDD setting.

Baseline models follow the official implementations provided by their authors \cite{liu-2024, mourya-2024}. Although a flexible search space is applied, some constraints remain due to the limited configurability of the original codebases.

\begin{table*}[!ht]
  \caption{\textbf{Hyperparameter search space (domains are inclusive).}}
  \label{tab:hparam-space}
  \centering
  \scriptsize
  \setlength{\tabcolsep}{3pt}
  \renewcommand{\arraystretch}{1.1}
  \newcolumntype{Z}{>{\centering\arraybackslash}X}
  \begin{tabularx}{0.96\textwidth}{@{} l Z Z Z Z @{}}
    \toprule
    \textbf{Module} & \textbf{Hyperparameter} & \textbf{Domain} & \textbf{Type} & \textbf{Notes} \\
    
    \midrule
    \multicolumn{5}{@{}l}{\textbf{Optimizer}}\\
    \cmidrule(lr){1-5}
     & name & \{Adam, AdamW\} & categorical & -- \\
     & lr & $[10^{-5},\,5\!\times\!10^{-3}]$ & log-uniform & -- \\
     & weight\_decay & $[10^{-6},\,10^{-2}]$ & log-uniform & -- \\
     & beta\_1 & $[0.85,\,0.95]$ & uniform & step $= 0.005$ \\
     & beta\_2 & $[0.98,\,0.999]$ & uniform & step $= 0.001$ \\
    
     \midrule
    \multicolumn{5}{@{}l}{\textbf{Scheduler}}\\
    \cmidrule(lr){1-5}
     & \textit{ReduceLROnPlateau}.factor & $[0.1,\,0.7]$ & uniform & mode $= min$ \\
     & \textit{ReduceLROnPlateau}.patience & \{5,10,20\} & categorical & threshold $= 10^{-4}$ \\
     & \textit{ReduceLROnPlateau}.cooldown & $[0]$ & integer & -- \\
     & \textit{ReduceLROnPlateau}.min\_lr & $[10^{-8},\,10^{-5}]$ & log-uniform & -- \\
    
     \midrule
    \multicolumn{5}{@{}l}{\textbf{Training} (ensure all models have sufficiently large effective batch size)}\\
    \cmidrule(lr){1-5}
     & batch\_size & \{4, 8, 16\} & categorical & -- \\
     & accumulate\_grad\_batches & \{1, 2, 4\} & categorical & -- \\
    
     \midrule
    \multicolumn{5}{@{}l}{\textbf{\Model{}}}\\
    \cmidrule(lr){1-5}
    \multicolumn{5}{@{}l}{\emph{CNN-based Residual Representation}}\\
     & num\_filters\_2d & $[1,5]$ & integer & step $= 1$ \\
     & filter\_size\_2d & \{3, 5\} & categorical & -- \\
     & filter\_size\_1d & \{3, 5\} & categorical & -- \\
     & is\_residual & \{True, False\} & categorical & -- \\
     & activation & \{tanh, relu, gelu\} & categorical & -- \\
    \multicolumn{5}{@{}l}{\emph{Adaptive Correction Layers (time)}}\\
     & layers & $[2,4]$ & integer & step $= 1$ \\
     & hidden\_dim & \{128, 256, 512\} & categorical & -- \\
     & out\_act & \{sigmoid, tanh, relu, none\} & categorical & -- \\
     & arl\_op & \{add, multiply\} & categorical & -- \\
    \multicolumn{5}{@{}l}{\emph{Adaptive Correction Layers (subcarrier)}}\\
     & layers & $[2,4]$ & integer & step $= 1$ \\
     & hidden\_dim & \{128, 256, 512, 1024, 2048\} & categorical & -- \\
     & out\_act & \{sigmoid, tanh, relu, none\} & categorical & -- \\
     & arl\_op & \{add, multiply\} & categorical & -- \\
    \multicolumn{5}{@{}l}{\emph{Shuffle Blocks}}\\
     & res\_layers & $[4,6]$ & integer & step $= 1$ \\
     & res\_dim & \{64, 128, 256\} & categorical & -- \\
     & groups & \{4, 8\} & categorical & -- \\
     & dropout & \{0.1, 0.2, 0.3\} & categorical & -- \\
    \multicolumn{5}{@{}l}{\emph{Transformer Encoder}}\\
     & $d_{\text{model}}$ & \{512, 768, 1024, 2048\} & categorical & -- \\
     & num\_layers & $[4,6]$ & integer & step $= 1$ \\
     & num\_heads & $[4,8]$ & integer & step $= 1$ \\
     & hidden\_dim & \{512, 1024, 2048\} & categorical & -- \\
     & dropout\_prob & \{0.1, 0.2, 0.3\} & categorical & -- \\
    
    \midrule
    \multicolumn{5}{@{}l}{\textbf{STEGMNN} (follows \cite{mourya-2024}'s implementation)}\\
    \cmidrule(lr){1-5}
     & n\_stacks & \{2\} & integer & fixed by authors \\
     & multi\_layer & \{2, 4, 8, 16\} & categorical & -- \\

    \midrule
    \multicolumn{5}{@{}l}{\textbf{LLM4CP} (follows \cite{liu-2024}'s implementation)}\\
    \cmidrule(lr){1-5}
    & res\_layers & $[2,8]$ & integer & step $= 1$ \\
    & res\_dim & \{64, 128, 256, 512, 1024, 2048, 4096\} & categorical & -- \\
    & gpt\_type & \{gpt2, gpt2-medium, gpt2-large\} & categorical & -- \\
    & gpt\_layers & $[2,8]$ & integer & step $= 1$ \\
    
    \midrule
    \multicolumn{5}{@{}l}{\textbf{RNN} (follows \cite{liu-2024}'s implementation)}\\
    \cmidrule(lr){1-5}
    & rnn\_hidden\_dim & \{128, 256, 512, 1024, 2048, 4096\} & -- \\
    & rnn\_num\_layers & $[1,8]$ & integer & step $= 1$ \\
    
    \midrule
    \multicolumn{5}{@{}l}{\textbf{CNN} (follows \cite{liu-2024}'s implementation)}\\
    \cmidrule(lr){1-5}
     & num\_filters & $[3,10]$ & integer & step $= 1$ \\
    \bottomrule
  \end{tabularx}
\end{table*}
\section{Spectral Efficiency (SE)}\label{sec:appendix-se}

While NMSE quantifies the accuracy of CSI prediction by measuring the element-wise deviation between the predicted and ground truth CSI, it does not directly reflect the impact of prediction quality on overall system performance. To ensure practical relevance, this study also evaluates the predicted DL CSI using end-to-end performance metrics, specifically spectral efficiency (SE).

SE is defined as the maximum achievable data rate per unit bandwidth and is a key performance metric in wireless communication systems. Maximizing SE is a central objective in system design. Its derivation is based on Shannon's capacity formula~\cite{shannon-1949}, applied to the signal model at subcarrier $k$:
\begin{equation}
  y_k = \csi_k^\dagger \mathbf{\omega}_k x_k + n_k,
  \label{eq:signal_model}
\end{equation}
where $x_k$ and $y_k$ are the transmitted and received signals, respectively, $\csi_k = \CSI^t [k]$ is the channel state information (CSI), $\mathbf{\omega}_k$ is the precoding vector, and $n_k$ represents additive white Gaussian noise (AWGN) with zero mean and variance $\sigma_n^2$.

\begin{figure*}[!ht]
  \centering
  \subfloat[\Regular{}]{
      \includegraphics[width=0.32\textwidth]{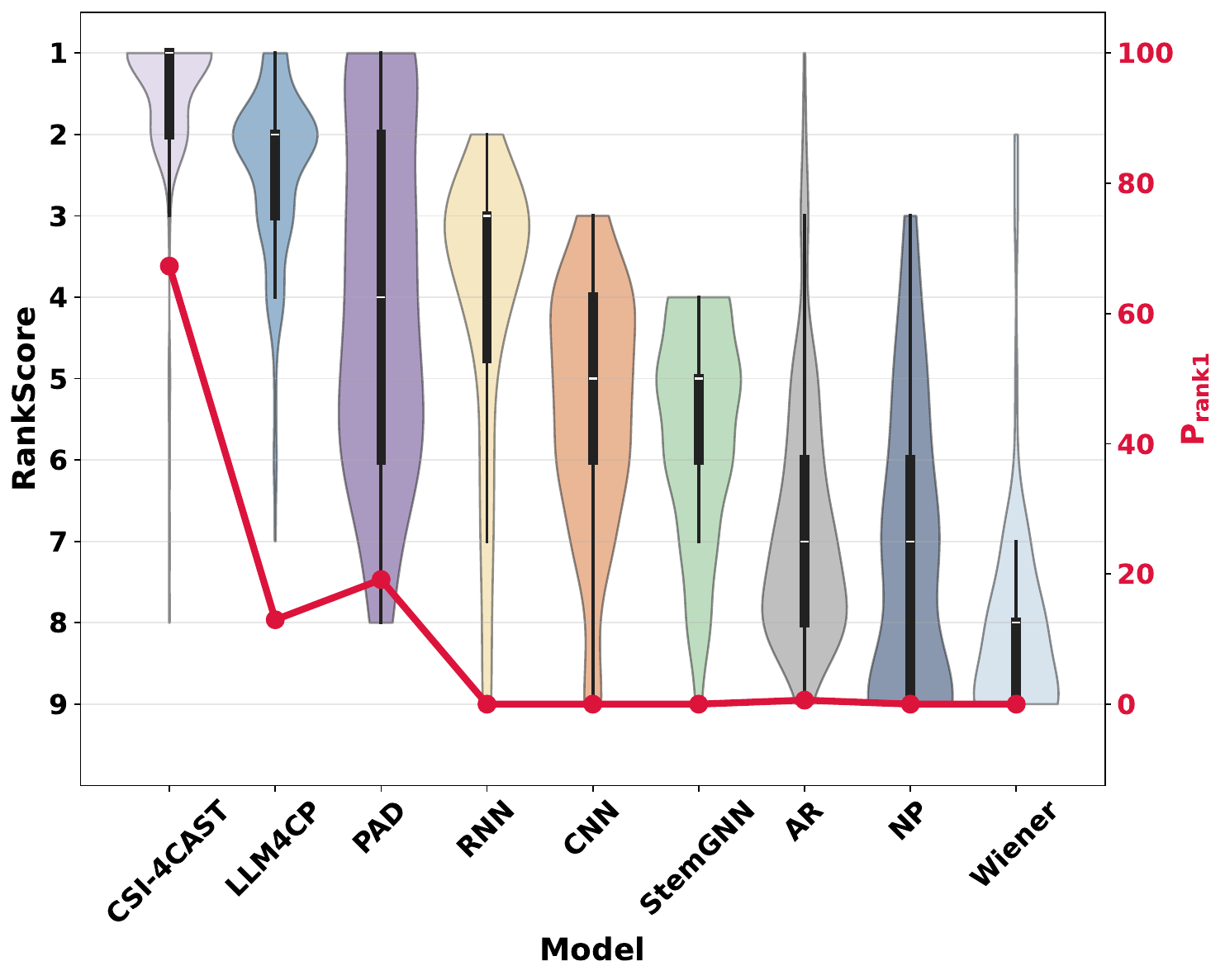}
      \label{fig:violin_regular_tdd_se}
  }
  \subfloat[\Robustness{}]{
      \includegraphics[width=0.32\textwidth]{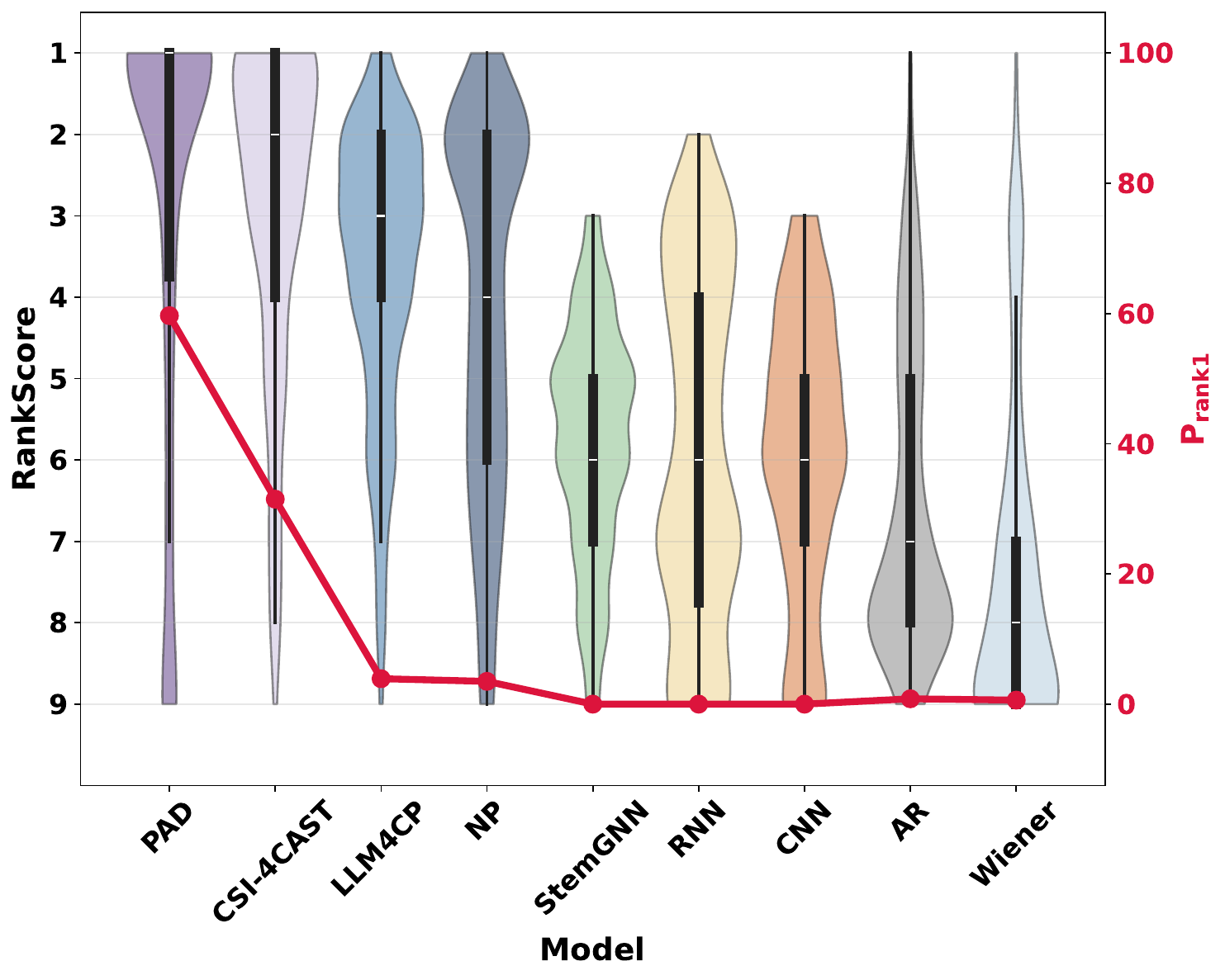}
      \label{fig:violin_robustness_tdd_se}
   }
  \subfloat[\Generalization{}]{
      \includegraphics[width=0.32\textwidth]{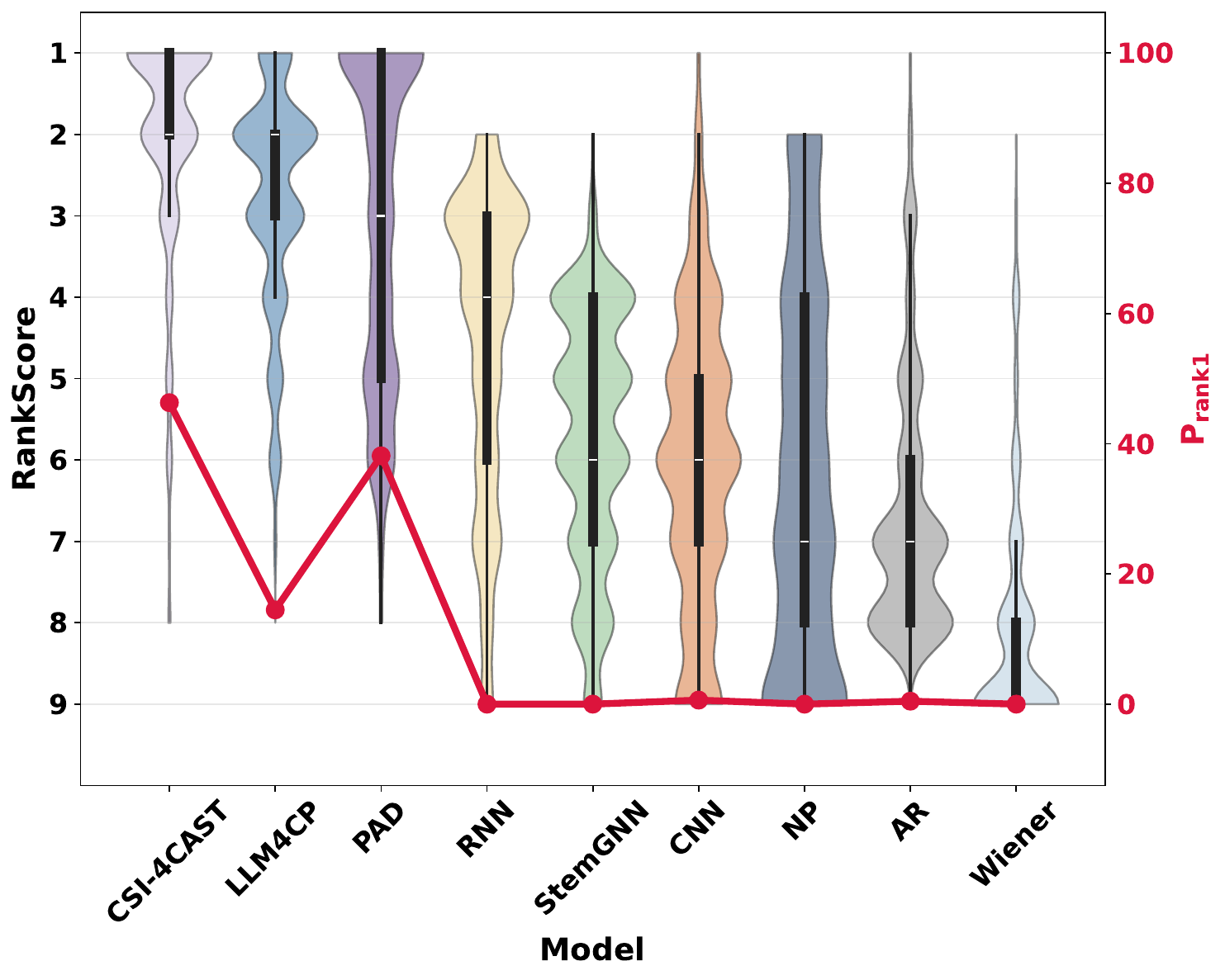}
      \label{fig:violin_generalization_tdd_se}
  }
  \caption{\textbf{\response{TDD: SE rank distribution of \Regular{}, \Robustness{}, and \Generalization{}.}} \textnormal{Within each panel, models are ordered left to right by their mean rank, $\mathrm{MeanRank}$ in \eqref{eq:mean-rank} (lower is better). Rank distributions are shown as violin plots, while top-1 percentages, $\mathbf{P}_{\mathrm{rank1}}$ in \eqref{eq:rank-1-percentage}, are plotted as a line graph.}}
  \label{fig:performance-regular-robustness-generalization-tdd-se}
\end{figure*}

\begin{figure*}[!ht]
  \centering
  \subfloat[\Regular{}]{
      \includegraphics[width=0.32\textwidth]{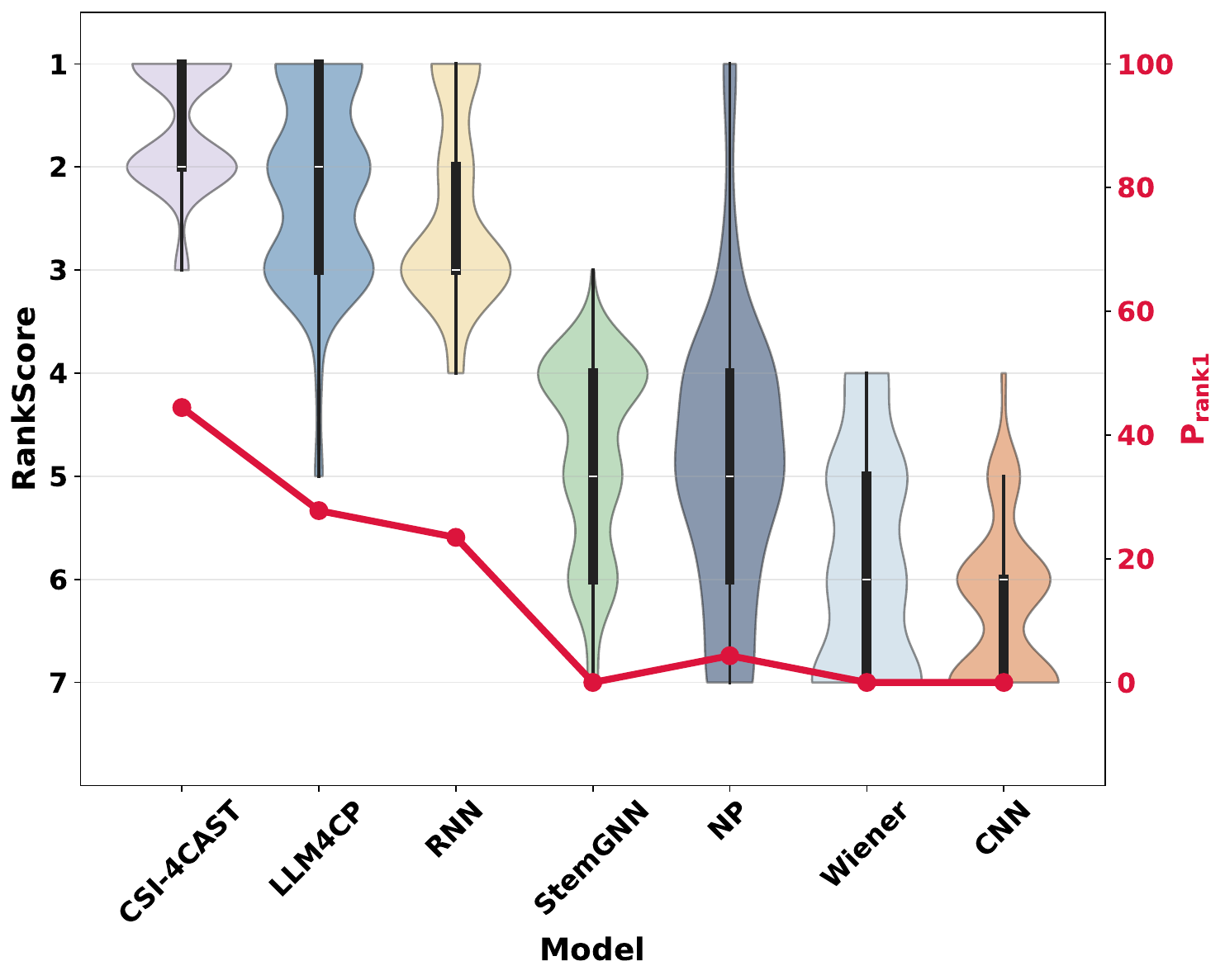}
      \label{fig:violin_regular_fdd_se}
  }
  \subfloat[\Robustness{}]{
      \includegraphics[width=0.32\textwidth]{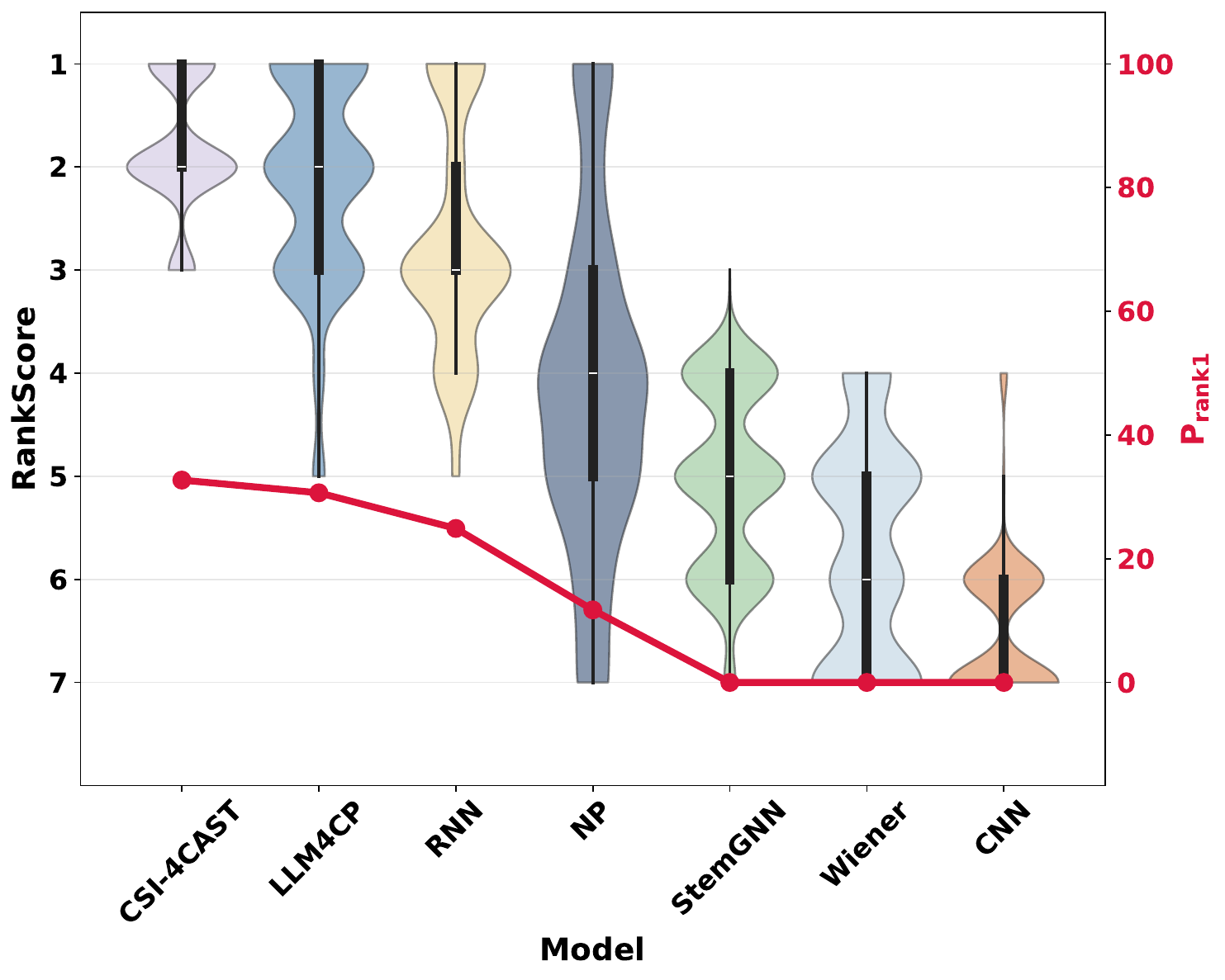}
      \label{fig:violin_robustness_fdd_se}
  }
  \subfloat[\Generalization{}]{
      \includegraphics[width=0.32\textwidth]{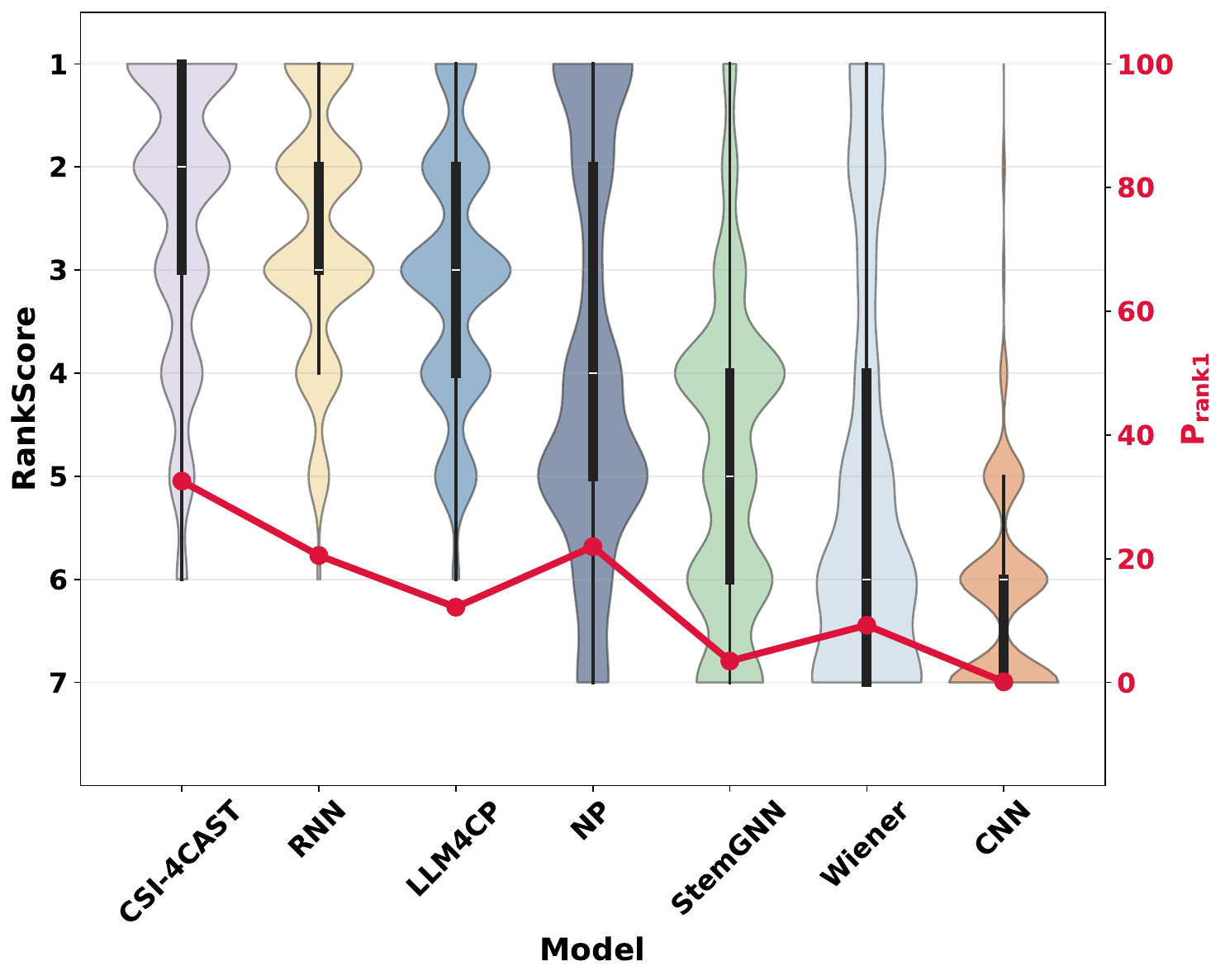}
      \label{fig:violin_generalization_fdd_se}
  }
  \caption{\textbf{\response{FDD: SE rank distribution of \Regular{}, \Robustness{}, and \Generalization{}.}}}
  \label{fig:performance-regular-robustness-generalization-fdd-se}
\end{figure*}

The theoretical SE is computed as:
\begin{equation}
  \text{SE} = \frac{1}{N_{\mathrm{sc}}} \sum_{k=1}^{N_{\mathrm{sc}}} \log_2 \left( 1 + \frac{\csi_k^\dagger \csi_k}{\sigma_n^2} \right),
  \label{eq:definition_of_SE}
\end{equation}
achieved when $\mathbf{\omega}_k = \frac{\csi_k}{\|\csi_k\|}$. 

For predicted DL CSI, the SE is estimated using:
\begin{equation}
   \widehat{\text{SE}} = \frac{1}{N_{\mathrm{sc}}} \sum_{k=1}^{N_{\mathrm{sc}}} \log_2 \left( 1 + \frac{
    \vert \predcsi_k^\dagger \csi_k \vert^2
   }{
    \sigma_n^2 \| \predcsi_k \|^2
   } \right),
  \label{eq:definition_of_predicted_SE}
\end{equation}
where $\predcsi_k = \predCSI[t, k]$ and $\csi_k$ denote the predicted and ground truth DL CSI, respectively (the subscript $f$ is omitted for simplicity). This formulation reflects that the BS configures the precoding vector $\mathbf{\omega}_k$ using the predicted CSI $\predcsi_k$, while the resulting SE is evaluated with respect to the actual channel $\csi_k$.

The SE rank distributions in the TDD and FDD systems are shown in Figs.~\ref{fig:performance-regular-robustness-generalization-tdd-se}-\ref{fig:performance-regular-robustness-generalization-fdd-se}, providing an overview of the SE performance of the models.

\section{Ablation Study}
\label{sec:appendix-ablation}

\response{Complementary to Section~\ref{sec:ablation-study}, which focuses on the overall performance and efficiency of each ablated model when specific components are removed or replaced, the following sections provide a more in-depth discussion of each module individually and offer further insights from an architectural design perspective. Specifically, Section~\ref{sec:appendix-ablation-experimental-setup} describes the experimental setup for the ablation study, and Section~\ref{sec:appendix-ablation-acl} presents the ablation of the ACL.}

\subsection{Experimental Setup for the Ablation Study}
\label{sec:appendix-ablation-experimental-setup}

\response{To better justify the architectural design choices, a comprehensive ablation study is conducted on \Model{}. Each ablation variant is defined by changing exactly one module of the full model while leaving all other modules unchanged. For the CNN, IDFT, ShuffleNet, and Transformer modules, which share the same structure in TDD and FDD, ablations are performed only in the TDD setting. For the ACL, whose structure differs between TDD and FDD, separate ablation studies are conducted with setting-specific configurations. The following variants are considered:}

\subsubsection{CNN ablation - TDD setting.}
\response{\begin{itemize}
  \item \textit{No CNN}: This ablation evaluates the performance of \Model{} with and without the initial CNN residual component.
\end{itemize}}

\subsubsection{IDFT ablation - TDD setting.}
\response{\begin{itemize}
  \item \textit{No IDFT}: The delay-domain branch is removed; the model uses only the frequency-domain path and does not perform the inverse DFT or the subsequent delay-branch processing. This ablation isolates whether the meaningful performance gain is due to delay-domain processing (IDFT).
\end{itemize}}

\subsubsection{ACL ablations}
\response{\begin{itemize}
    \item TDD setting:
    \begin{itemize}
      \item \textit{No ACL}: The ACL is replaced by a pass-through: all learnable temporal and subcarrier corrections are removed. This ablation isolates whether the meaningful contribution comes from the ACL. 
      \item \textit{Norm Replace ACL}: The ACL is replaced by layer normalization over the temporal dimension on each branch, with no MLP-based correction. The normalization applied on the temporal dimension is the same as the original ACL which is only applied on the temporal dimension. This ablation tests whether lightweight normalization alone is sufficient, or whether content-dependent adaptive correction is necessary.
      \item \textit{Add Subcarrier ACL}: The TDD model is given the same subcarrier-axis correction as in FDD, in addition to the standard temporal ACL. This ablation tests whether cross-subcarrier adaptation helps even when TDD reciprocity already reduces uplink-to-downlink mismatch.
    \end{itemize}
    \item FDD setting:
    \begin{itemize}
      \item \textit{No ACL}: The same pass-through replacement as above is applied in the FDD setting. The model discards all learnable ACL transformations. This ablation isolates whether the meaningful contribution comes from the ACL.
      \item \textit{No Subcarrier ACL}: The subcarrier correction in the ACL is disabled; only temporal correction remains. This ablation measures how much of the FDD improvement is due specifically to subcarrier-domain adaptation.
    \end{itemize}
\end{itemize}}

\subsubsection{ShuffleNet ablations - TDD setting.}
\response{\begin{itemize}
    \item \textit{MLP Replace ShuffleNet}: The ShuffleNet block is replaced by an MLP-based embedding that first flattens the time-subcarrier feature map into a one-dimensional vector and then processes it with fully connected layers. As a result, the model ignores the local 2D structure across time and subcarriers in this stage. This ablation tests whether the reported gain depends on the convolutional ShuffleNet inductive bias or can also be achieved with simpler dense feature extraction.
    \item \textit{MobileNet Replace ShuffleNet}: The ShuffleNet block is replaced by a MobileNet-style convolutional backbone (inverted residual blocks with optional squeeze-and-excitation). This ablation tests whether the benefit is specific to the ShuffleNet design or can be reproduced by another lightweight convolutional backbone.
\end{itemize}}

\subsubsection{Transformer ablations - TDD setting.}
\response{\begin{itemize}
    \item \textit{MLP Replace Transformer}: The Transformer encoder is replaced by a feed-forward MLP that preserves input and output dimensions but removes self-attention and sequence-wise token interaction. This is an aggressive ablation that tries to isolate the contribution of the attention mechanism and sequence modeling.
    \item \textit{LSTM Replace Transformer}: The Transformer predictor is replaced by an LSTM. This preserves recurrent temporal modeling over the history sequence while removing self-attention-based global token interaction. The ablation therefore tests whether the predictor gain is specific to attention-based sequence modeling or can also be achieved with a recurrent backbone.
\end{itemize}}

\response{Taken together, these ablations separate the contributions of the CNN, the delay-domain path (IDFT), the ACL (adaptive temporal and subcarrier correction), the ShuffleNet block (convolutional feature extraction), and the Transformer (attention-based sequence modeling).}

\subsubsection{Training Process for the Ablations}
\response{Furthermore, to make the comparison as fair as possible, the training process for the ablations follows the same procedure as that for the baseline models. Specifically, the hyperparameter tuning process is conducted with the same Optuna framework using 1 NVIDIA H200 GPU for 30 hours. Each trial is budgeted for 25 training epochs. With manual verification, the tuning process typically covers 35--50 trials with good coverage of the different hyperparameters. The hyperparameter search space for the ablations is shown in Table~\ref{tab:hparam-space-ablation}. Note that for the ablations, when a specific module is replaced, the hyperparameters of the remaining modules are fixed to the best values found for the full model. Therefore, for ablations that merely remove some modules, no model-specific hyperparameters need to be tuned; only the optimizer, scheduler, and training hyperparameters are tuned, and thus no tuning parameters are listed in the table. The hyperparameters for the optimizer, scheduler, and training are the same as those of the baseline models in Table~\ref{tab:hparam-space}.}

\begin{table*}[!t]
    \caption{\textbf{\response{Model-specific hyperparameter search space for the ablations (domains are inclusive).}}}
    \label{tab:hparam-space-ablation}
    \centering
    \scriptsize
    \setlength{\tabcolsep}{3pt}
    \renewcommand{\arraystretch}{1.1}
    \newcolumntype{Z}{>{\centering\arraybackslash}X}
    \begin{tabularx}{0.96\textwidth}{@{} l Z Z Z Z @{}}
      \toprule
      \multicolumn{5}{@{}l}{\textbf{Add Subcarrier ACL}}\\
      \cmidrule(lr){1-5}
      & layers & $[1,4]$ & integer & step $= 1$ \\
      & hidden\_dim & \{256, 512, 1024\} & categorical & -- \\
      & out\_act & \{tanh, relu\} & categorical & -- \\
      & arl\_op & \{add, multiply\} & categorical & -- \\

      \midrule
      \multicolumn{5}{@{}l}{\textbf{MLP Replace ShuffleNet}}\\
      \cmidrule(lr){1-5}
      & mlp\_num\_layers & $[2,6]$ & integer & step $= 1$ \\
      & mlp\_hidden\_dim & \{128, 256, 512, 1024, 2048\} & categorical & -- \\
      & mlp\_dropout & \{0.1, 0.2, 0.3\} & categorical & -- \\

      \midrule
      \multicolumn{5}{@{}l}{\textbf{MobileNet Replace ShuffleNet}}\\
      \cmidrule(lr){1-5}
      & mobilenet\_num\_blocks & $[2,8]$ & integer & step $= 1$ \\
      & mobilenet\_base\_channels & \{16, 24, 32, 48, 64, 96, 128\} & categorical & -- \\
      & mobilenet\_expand\_ratio & \{1, 2, 3\} & categorical & -- \\
      & mobilenet\_kernel\_size & \{3, 5\} & categorical & -- \\
      & mobilenet\_se\_ratio & \{2, 4, 8\} & categorical & -- \\

      \midrule
      \multicolumn{5}{@{}l}{\textbf{MLP Replace Transformer}}\\
      \cmidrule(lr){1-5}
      & mlp\_num\_layers & $[2,6]$ & integer & step $= 1$ \\
      & mlp\_hidden\_dim & \{128, 256, 512, 1024, 2048\} & categorical & -- \\
      & mlp\_dropout & \{0.1, 0.2, 0.3\} & categorical & -- \\

      \midrule
      \multicolumn{5}{@{}l}{\textbf{LSTM Replace Transformer}}\\
      \cmidrule(lr){1-5}
      & lstm\_num\_layers & $[1,4]$ & integer & step $= 1$ \\
      & lstm\_hidden\_dim & \{128, 256, 512, 1024\} & categorical & -- \\
      & lstm\_dropout & \{0.1, 0.2, 0.3\} & categorical & -- \\
      & lstm\_bidirectional & \{True, False\} & categorical & -- \\
      \bottomrule
    \end{tabularx}
  \end{table*}

\subsection{Further Discussion on the ACL Module}
\label{sec:appendix-ablation-acl}

\subsubsection{Visualization of the ACL's Mechanism}
\begin{figure*}[!ht]
  \centering
  \subfloat[Before ACL.]{
      \includegraphics[width=0.32\textwidth]{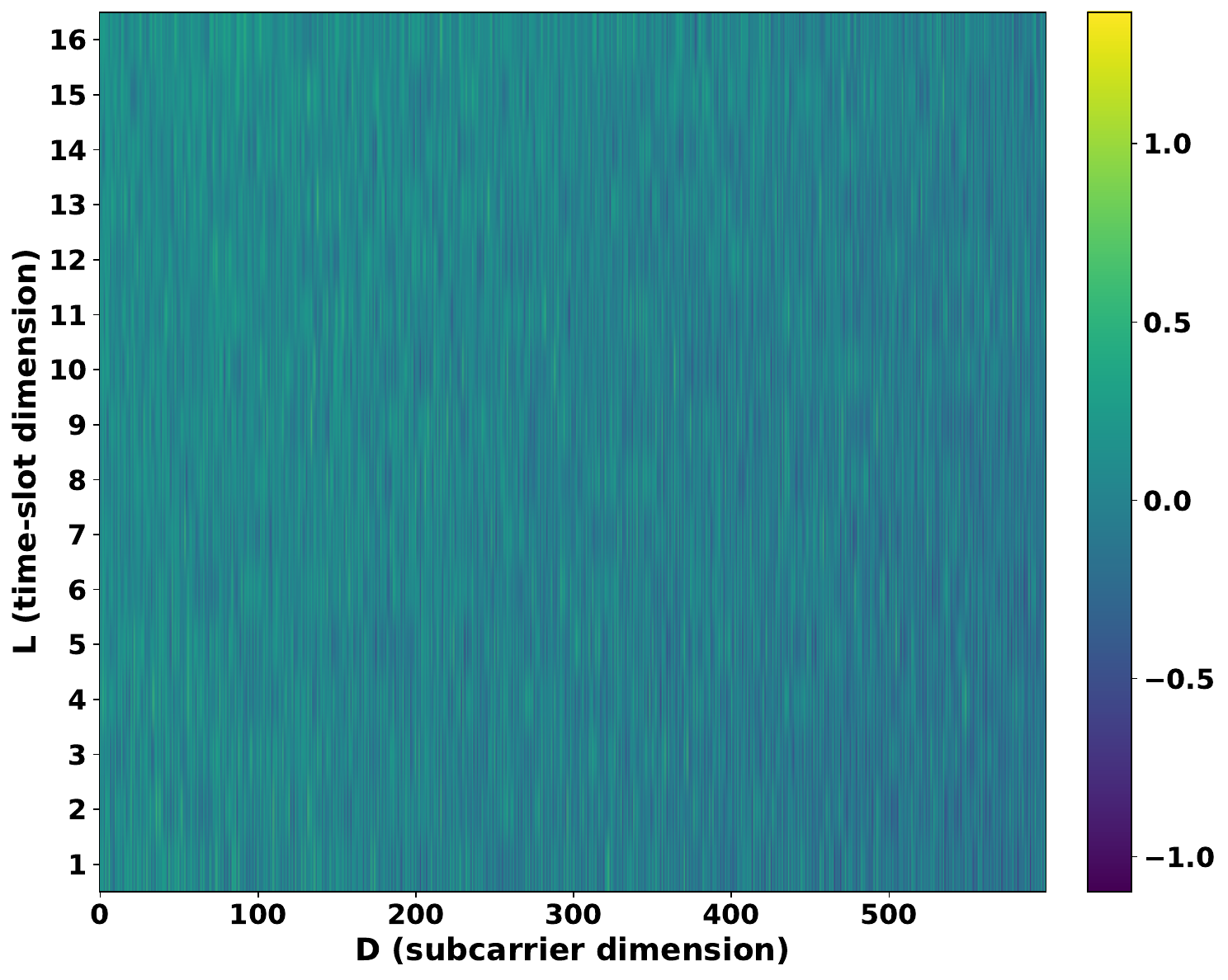}
      \label{fig:ABL_ACL_TDD_freq_origin}
  }
  \subfloat[After temporal ACL.]{
      \includegraphics[width=0.32\textwidth]{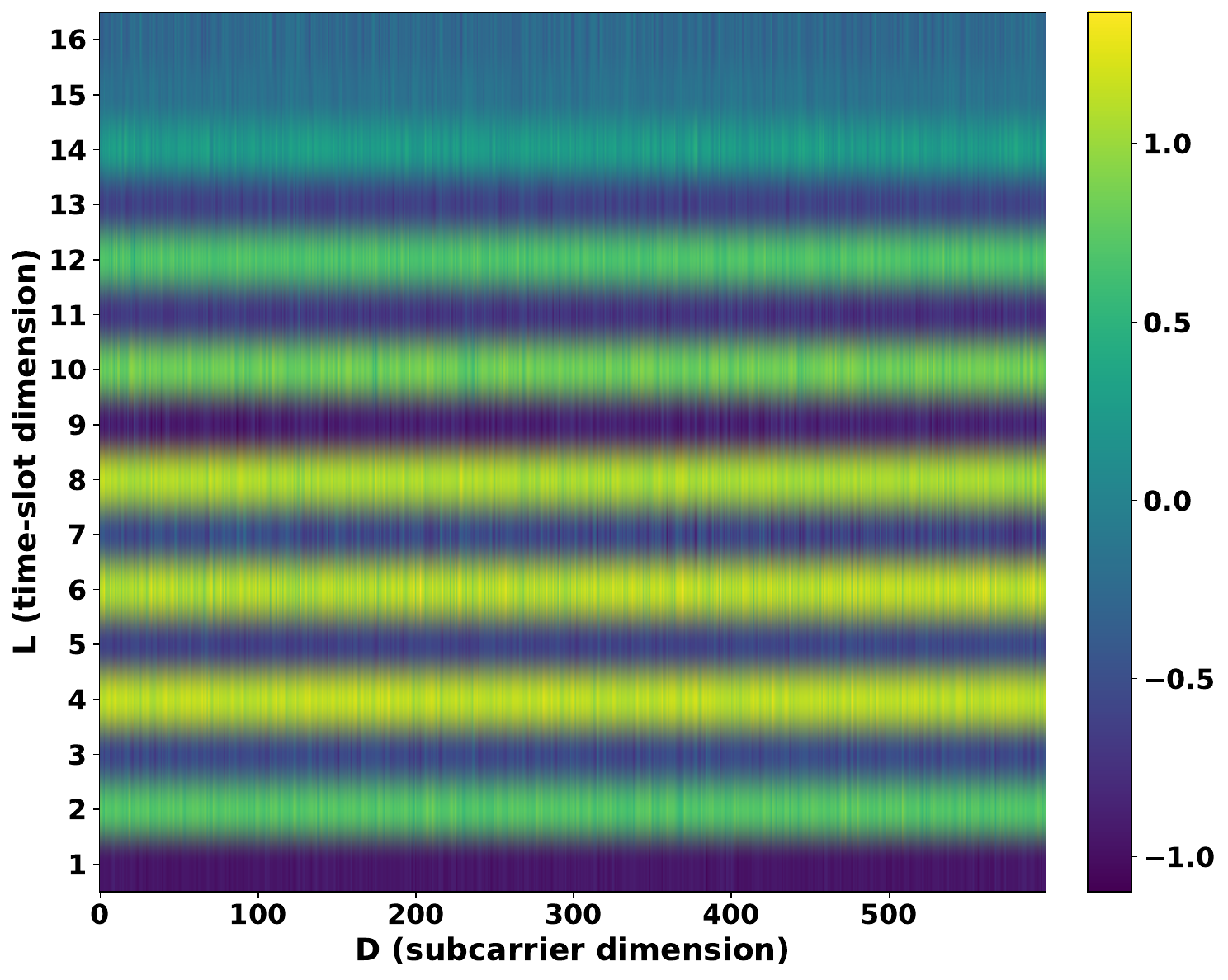}
      \label{fig:ABL_ACL_TDD_freq_after_temporal}
  }
  \caption{\textbf{Intermediate frequency-branch CSI representations for TDD.} \textnormal{Channel Model: CDL-A, Delay Spread: 30\,ns, Velocity: 1\,m/s. Visualization of CSI before the Adaptive Correction Layers (ACL), and after temporal ACL.}}
  \label{fig:ABL_ACL_TDD_freq}
\end{figure*}

\begin{figure*}[!ht]
  \centering
  \subfloat[Before ACL.]{
      \includegraphics[width=0.32\textwidth]{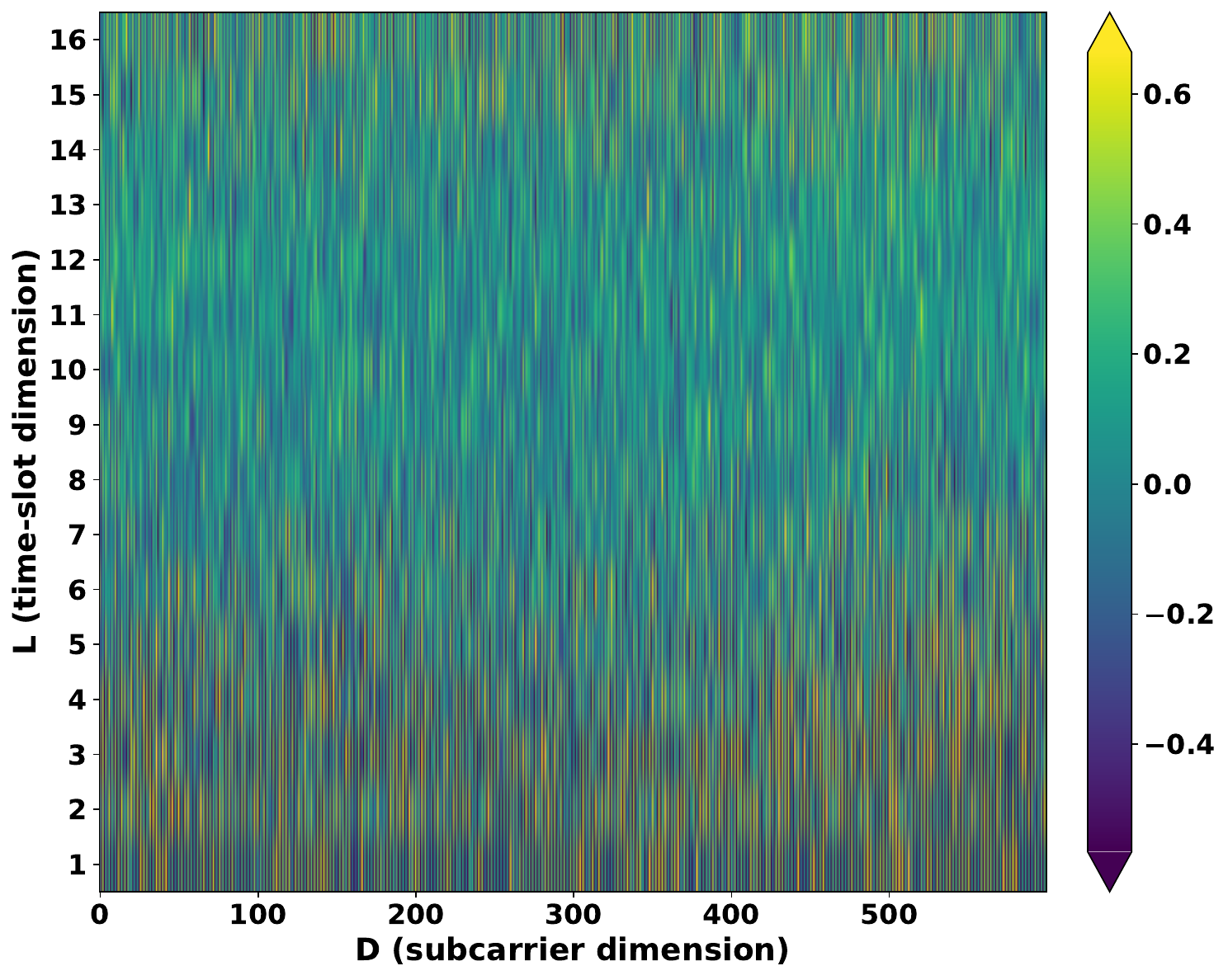}
      \label{fig:ABL_ACL_FDD_freq_origin}
  }
  \subfloat[After temporal ACL.]{
      \includegraphics[width=0.32\textwidth]{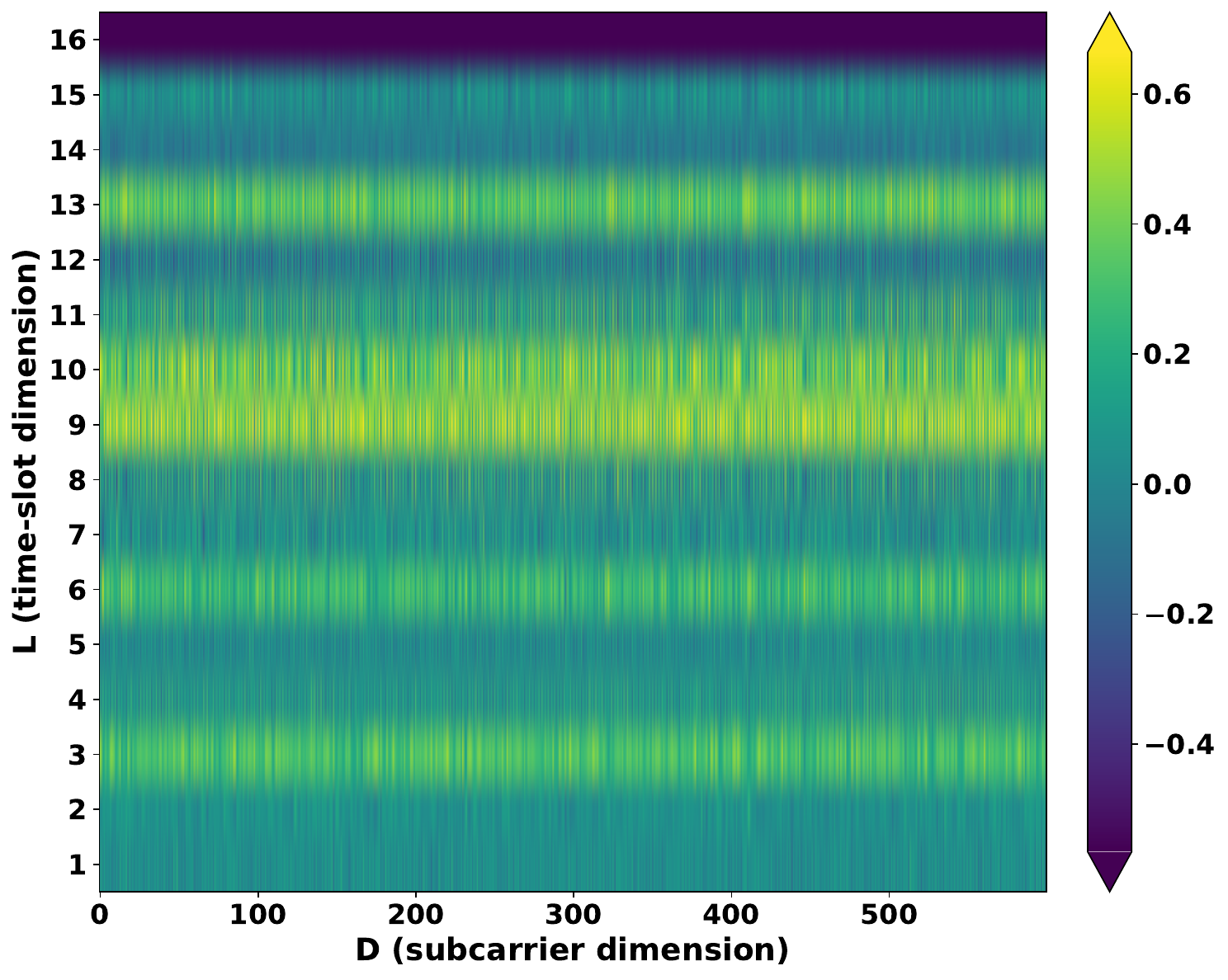}
      \label{fig:ABL_ACL_FDD_freq_after_temporal}
  }
  \subfloat[After subcarrier ACL.]{
      \includegraphics[width=0.32\textwidth]{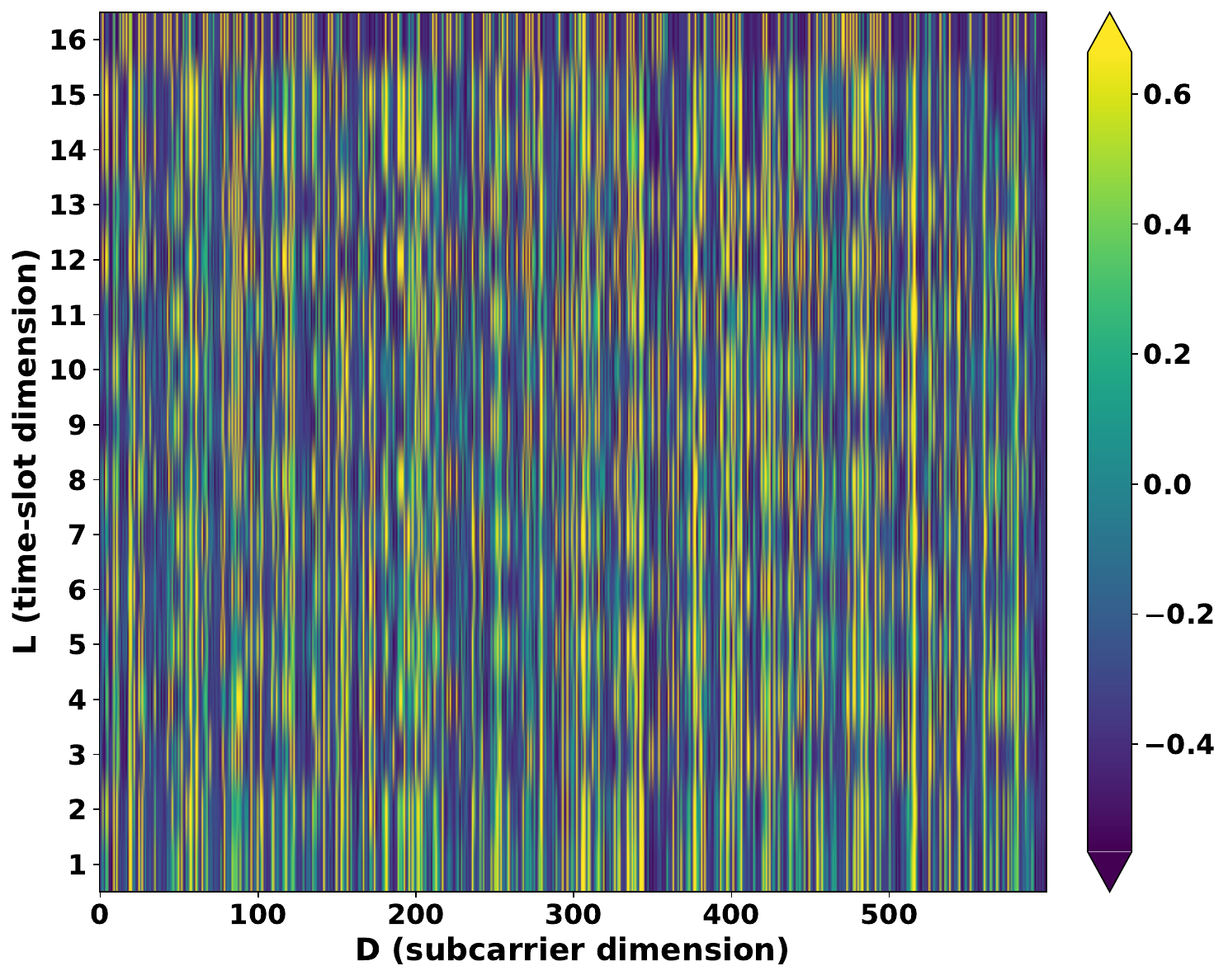}
      \label{fig:ABL_ACL_FDD_freq_after_subcarrier}
  }
  \caption{\textbf{Intermediate frequency-branch CSI representations for FDD.} \textnormal{Channel Model: CDL-A, Delay Spread: 30\,ns, Velocity: 1\,m/s. Visualization of CSI before the Adaptive Correction Layers (ACL), after temporal ACL, and after subcarrier ACL.}}
  \label{fig:ABL_ACL_FDD_freq}
\end{figure*}

\response{To make the action of ACL more interpretable, Fig.~\ref{fig:ABL_ACL_TDD_freq} and Fig.~\ref{fig:ABL_ACL_FDD_freq} visualize representative intermediate feature maps extracted from a trained CSI-4CAST model. A representative test sample from the scenario \textit{CDL-A, delay spread = 30 ns, velocity = 1 m/s} is selected and is passed through the trained model. The frequency-branch representation at three locations are presented: before ACL, after temporal ACL, and after subcarrier ACL (only for FDD). For visualization only, the intermediate representations corresponding to different antenna pairs are aggregated by element-wise averaging, yielding a single heatmap of size $|T|\times 2N_{sc}$. The vertical axis corresponds to the historical time index, and the horizontal axis corresponds to the real-valued frequency-feature dimension after concatenating the real and imaginary parts. The same procedure can be applied to the delay branch, where analogous behavior is observed. For brevity, only the frequency branch is shown.}

\response{For TDD, Fig.~\ref{fig:ABL_ACL_TDD_freq} shows that the temporal ACL transforms the comparatively unstructured feature map into clear horizontal bands, corresponding to different time steps of the historical CSI. These bands should be interpreted as lag-dependent correction patterns rather than as raw CSI patterns themselves. Because the same temporal MLP is applied to each frequency feature, the learned correction varies mainly with the time index and is broadly shared across the frequency axis. This behavior is consistent with the role of ACL1 as an adaptive temporal correction module: different historical slots contribute unequally to future prediction due to temporal coherence and lag-dependent reliability. Since TDD does not use subcarrier ACL, the after-temporal representation is the ACL module's output.}

\response{For FDD, Fig.~\ref{fig:ABL_ACL_FDD_freq} again shows horizontal banding after the temporal ACL, indicating that lag-dependent correction remains important. After the subcarrier ACL, additional vertical patterns appear, indicating frequency-dependent refinement across the feature axis. This is consistent with the FDD task, where inter-band prediction depends not only on temporal evolution but also on inter-subcarrier correlation and coherence-bandwidth structure. In the delay branch, the same mechanism appears as tap-dependent correction associated with multipath structure.}

\subsubsection{\textit{Norm Replace ACL} Ablation}
\label{sec:appendix-ablation-norm-replace-acl}

\response{It may seem counterintuitive that the \textit{Norm Replace ACL} variant, which applies layer normalization along the temporal dimension, produces worse performance than the \textit{No ACL} variant and even the worst performance among all ablations (Table~\ref{tab:ablation-study}). In fact, if the \textit{No ACL} variant is regarded as a passive ablation, the \textit{Norm Replace ACL} variant represents a more ``aggressive'' active distortion. When ACL is removed, the downstream encoder still receives the original branch representations, allowing the remaining network to learn directly from the raw temporal structure. In contrast, the \textit{Norm Replace ACL} variant does not preserve the original representation. It removes the temporal mean, absolute scale, and relative temporal variance of each subcarrier/tap channel, although these statistics themselves contain informative cues for TDD CSI prediction. In particular, for temporally coherent channels, slowly varying or nearly constant trajectories provide useful predictive signals; forcing them to zero mean and unit variance may suppress this information and even amplify small fluctuations or noise. Moreover, an intuitive explanation can be drawn from the visualization in Fig.~\ref{fig:ABL_ACL_TDD_freq}. ACL identifies specific lag-dependent patterns within the channel and leverages these patterns to refine the representation and improve performance. By contrast, the \textit{Norm Replace ACL} variant effectively reverses this process: normalization flattens the representation and removes much of the informative structure.}


\bibliographystyle{IEEEtran}
\bibliography{ref}

\end{document}